\documentclass[journal]{IEEEtran}
\usepackage{times}
\usepackage{epsfig}
\usepackage{graphicx}
\usepackage{amsmath}
\usepackage{amssymb}
\usepackage{bm}
\usepackage{amsthm}
\usepackage{pifont}
\usepackage{enumitem}
\newcommand{\cmark}{\ding{51}}%
\newcommand{\xmark}{\ding{56}}%
\usepackage{multicol}
\usepackage{multirow}
\usepackage{algorithm}
\usepackage{algorithmic}
\newtheorem{prop}{Proposition}
\usepackage{subcaption} % caption
\graphicspath{{image/}}
\usepackage{diagbox}
\usepackage[pagebackref=true, breaklinks=true, colorlinks, bookmarks=false]{hyperref}

\usepackage{cite}
\usepackage{colortbl,boldline,booktabs,makecell}
\usepackage{tabu}
\usepackage{balance}
\usepackage{cleveref}
\crefformat{section}{\S#2#1#3} % see manual of cleveref, section 8.2.1
\crefformat{subsection}{\S#2#1#3}
\crefformat{subsubsection}{\S#2#1#3}

\begin{document}

\title{Conditional Variational Image Deraining}

\author{
  Yingjun Du$^{1,\#}$,
  Jun Xu$^{2,\#}$,
  Xiantong Zhen$^3$,
  Ming-Ming Cheng$^2$,
  Ling Shao$^3$
\thanks{
$^\#$The first two authors contribute equally. $^1$Informatics Institute, University of Amsterdam, Amsterdam, Netherlands.
$^2$TKLNDST, College of Computer Science, Nankai University, Tianjin 300071, China.
$^3$Inception Institute of Artificial Intelligence, Abu Dhabi, UAE.
This work is an extension of our conference paper~\cite{vid_wacv}.
% Xiantong Zhen (zhenxt@gmail.com) is the corresponding author. 
%
%
}
}

% make the title area
\maketitle

\begin{abstract}
%Images captured in severe weather such as rain and snow will significantly degrade the accuracy of vision systems, e.g., for outdoor video surveillance or autonomous driving.\, 
Image deraining is an important yet challenging image processing task.
Though deterministic image deraining methods are developed with encouraging performance, they are infeasible to learn flexible representations for probabilistic inference and diverse predictions.
% they are limited by the deterministic deraining manner, since it is infeasible to define the optimal solution of a real-world rainy image.
%
Besides, rain intensity varies both in spatial locations and across color channels, making this task more difficult.
In this paper, we propose a Conditional Variational Image Deraining (CVID) network for better deraining performance, leveraging the exclusive generative ability of Conditional Variational Auto-Encoder (CVAE) on providing diverse predictions for the rainy image.
To perform spatially adaptive deraining, we propose a spatial density estimation (SDE) module to estimate a rain density map for each image.
Since rain density varies across different color channels, we also propose a channel-wise (CW) deraining scheme.
Experiments on synthesized and real-world datasets show that the proposed CVID network achieves much better performance than previous deterministic methods on image deraining.
Extensive ablation studies validate the effectiveness of the proposed SDE module and CW scheme in our CVID network.
The code is available at \url{https://github.com/Yingjun-Du/VID}.
\end{abstract}

% Note that keywords are not normally used for peer review papers.
\IEEEpeerreviewmaketitle

\section{Introduction}

\IEEEPARstart{T}{he} presence of rain undesirably degrades the visual authenticity of images for human perception, and drastically obstacle the performance of vision systems~\cite{kang2012automatic}.\,
Image deraining aims to remove the rain streaks from the degraded image, and recover its clean background.\, 
It has received increasing attention due to its prerequisite role in many practical applications, such as video surveillance~\cite{sultani2018real}, object detection~\cite{maskrcnn}, and object segmentation~\cite{RANet2019}, \textsl{etc}.

Previous image deraining methods can be roughly divided into three categories: the optimization based methods~\cite{luo2015removing,zhu2017joint,chang2017transformed}, the discriminative learning based methods~\cite{fu2017clearing,Ren_2019_CVPR,yang2019tip}, and the generative learning based methods~\cite{li2016rain,zhang2017image,qian2018attentive}.\, 
Among them, optimization based methods~\cite{chen2013generalized,chang2017transformed} employ proper regularizers to restore rainy image under the linear additive composite model~\cite{barnum2010analysis,kang2012automatic,STAR2020} or non-linear screen blend composite model~\cite{luo2015removing}.\, 
Discriminative learning based methods~\cite{fu2017removing,li2018recurrent,yang2019tip} directly learn deterministic (non-linear) mapping function from the rainy image to its clean background.\, 
Generative methods~\cite{li2016rain,zhang2017image,qian2018attentive} leverage the generative modeling capabilities of Gaussian Mixture Models~\cite{pgpd,gid2018} or conditional Generative Adversarial Networks (GANs)~\cite{gan,mirza2014conditional}, inspired by their success on synthesizing visually appealing images.

\begin{figure}[t]
\begin{center}
  \begin{subfigure}[t]{0.115\textwidth}
		\includegraphics[width=1\textwidth]{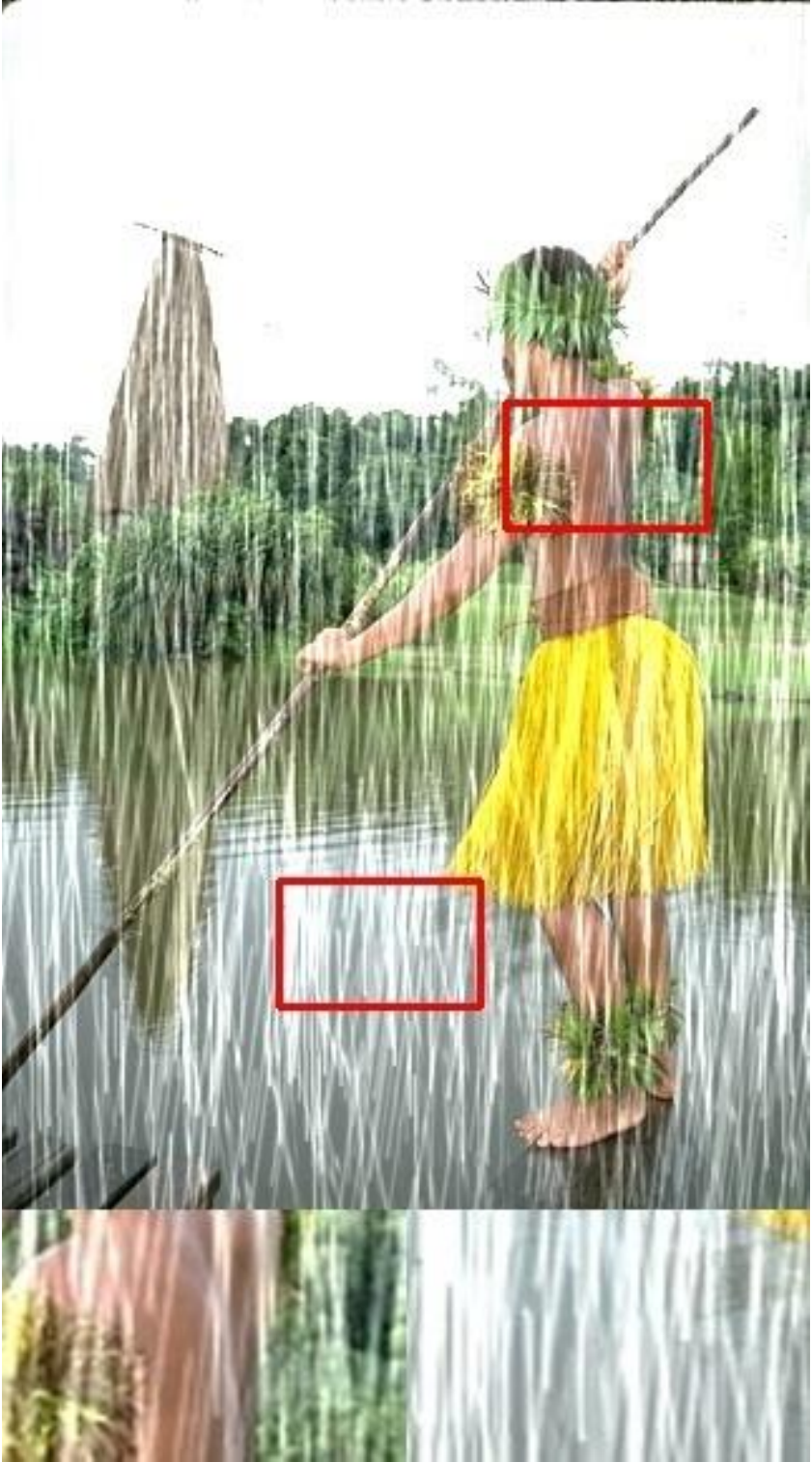}
		\subcaption{\scriptsize Rainy Image}
	\end{subfigure}
	\begin{subfigure}[t]{0.115\textwidth}
		\includegraphics[width=1\textwidth]{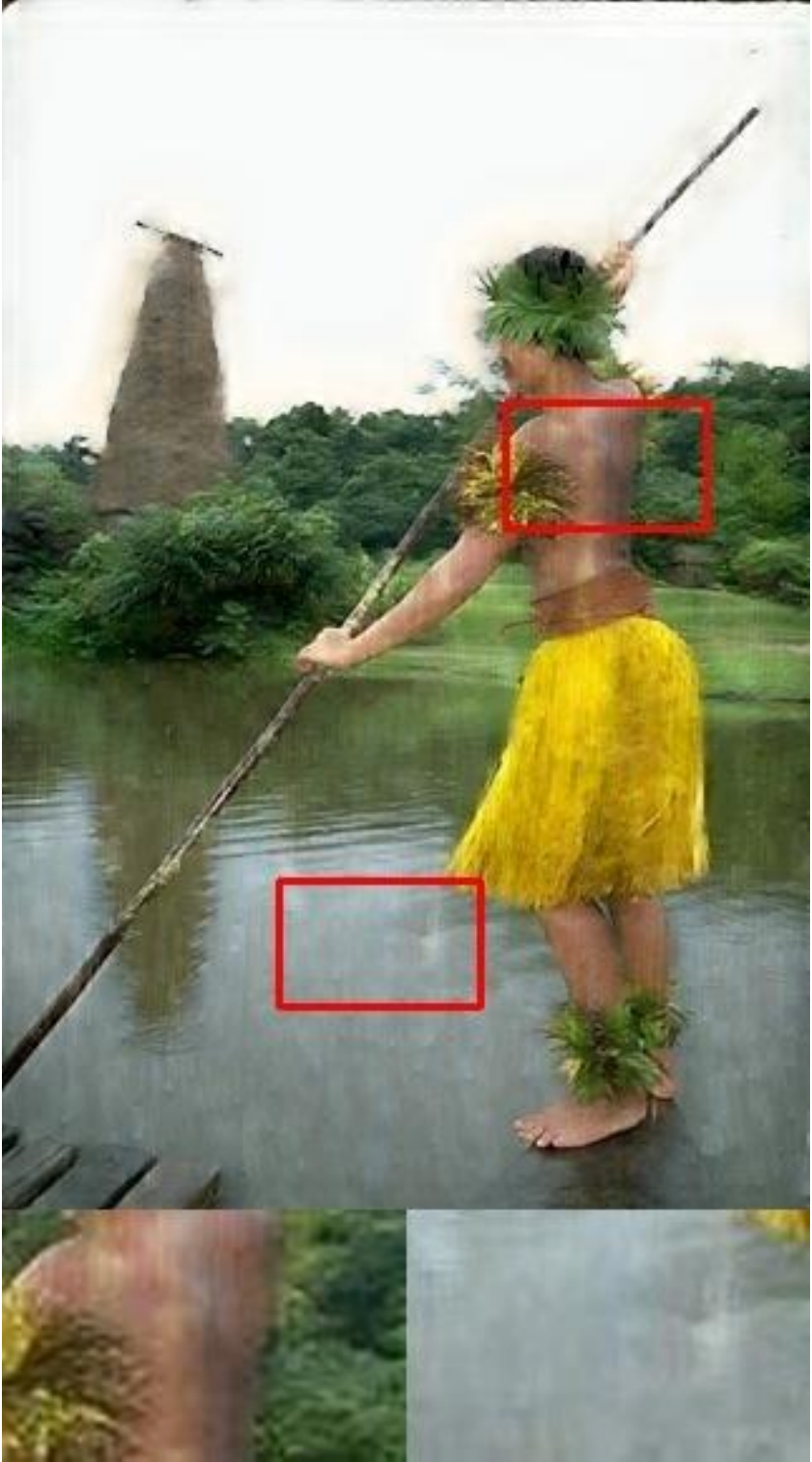}
		\subcaption{\scriptsize GMM~\cite{li2016rain}}
	\end{subfigure}
	\begin{subfigure}[t]{0.115\textwidth}
		\includegraphics[width=1\textwidth]{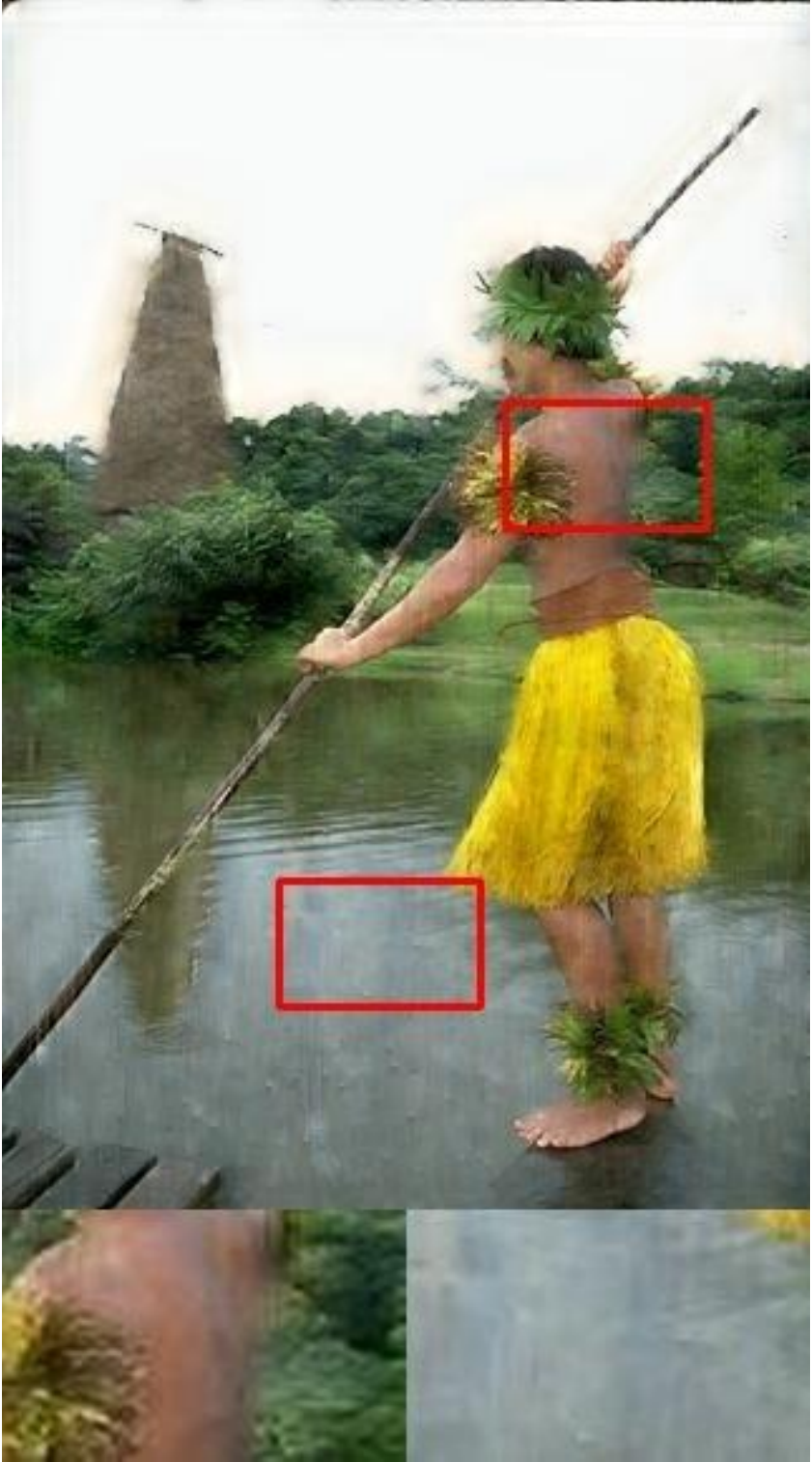}
		\subcaption{\scriptsize ID-GAN~\cite{zhang2017image}}
	\end{subfigure}
	\begin{subfigure}[t]{0.115\textwidth}
		\includegraphics[width=1\textwidth]{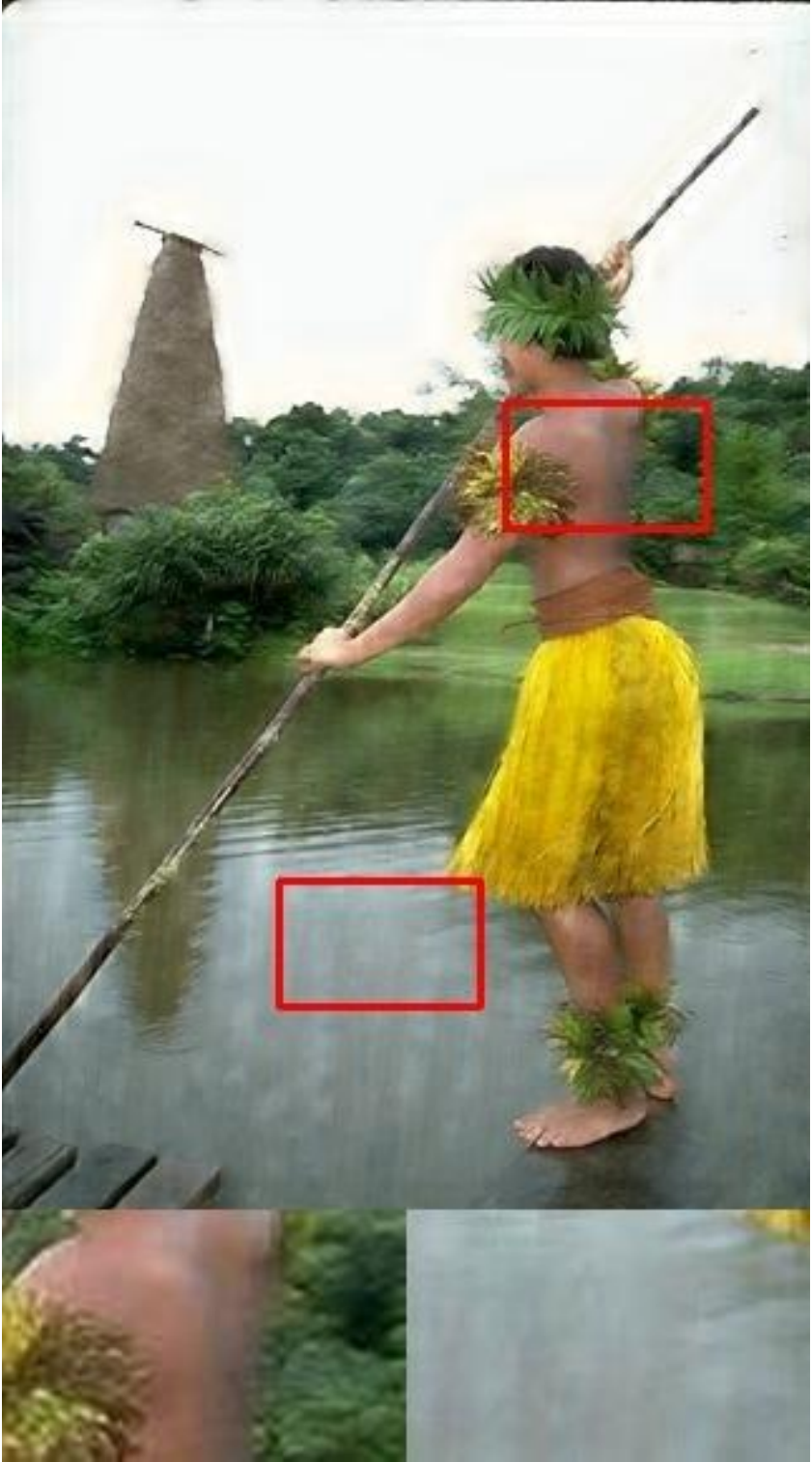}
		\subcaption{\scriptsize DDN~\cite{fu2017removing}}
	\end{subfigure}
	\begin{subfigure}[t]{0.115\textwidth}
		\includegraphics[width=1\textwidth]{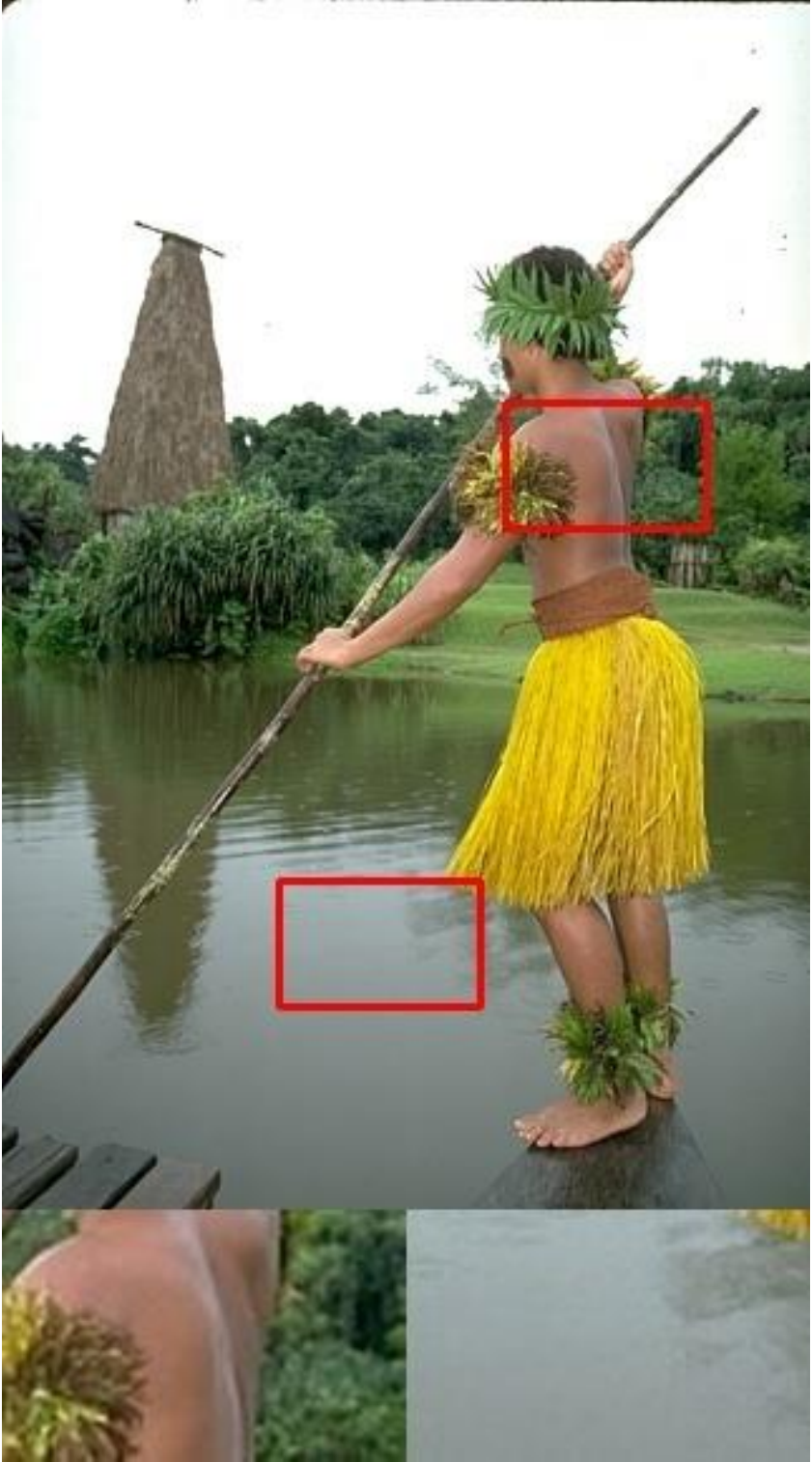}
		\subcaption{\scriptsize Ground Truth}
	\end{subfigure}
	\begin{subfigure}[t]{0.115\textwidth}
		\includegraphics[width=1\textwidth]{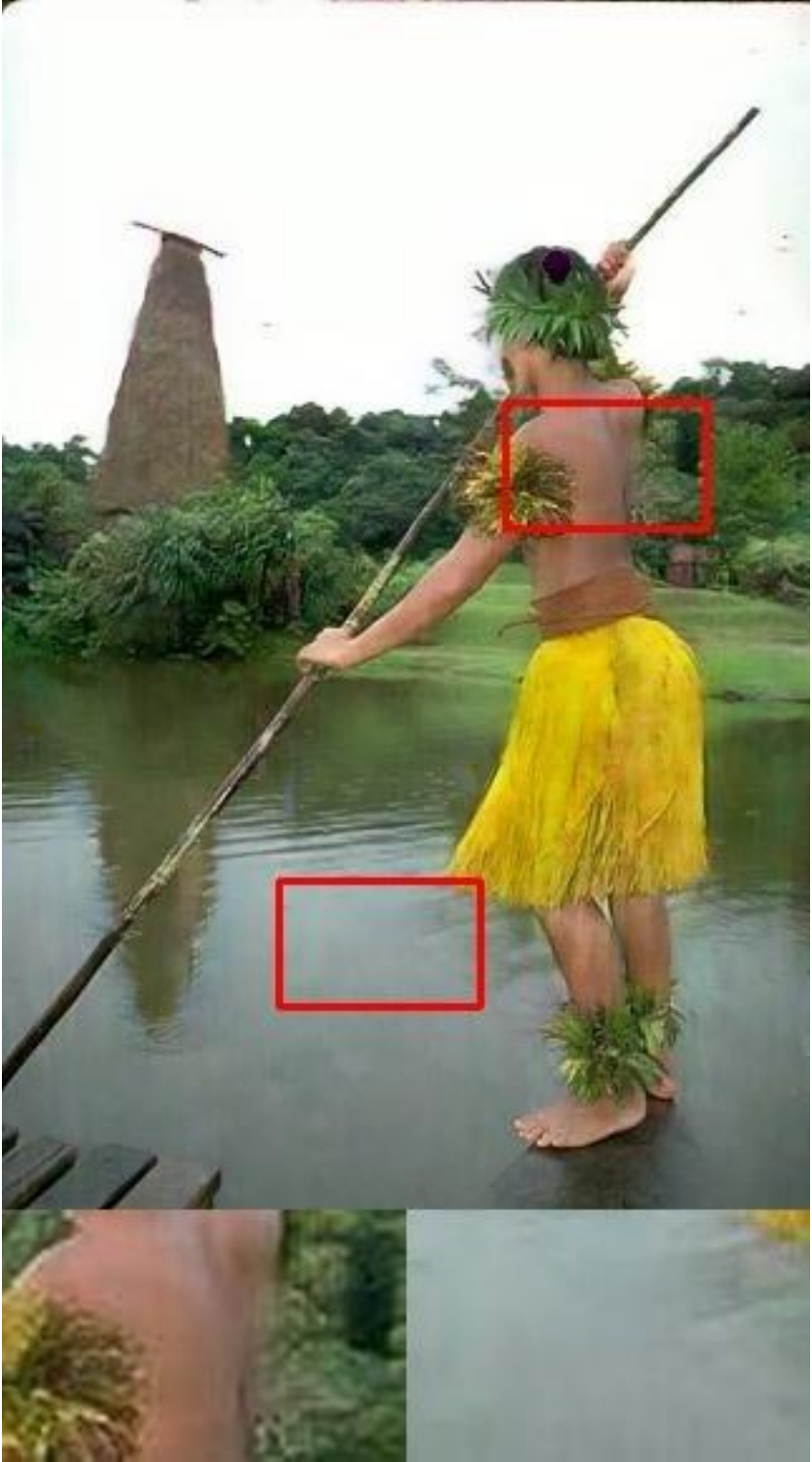}
		\subcaption{\scriptsize CVID$_1$}
	\end{subfigure}
	\begin{subfigure}[t]{0.115\textwidth}
		\includegraphics[width=1\textwidth]{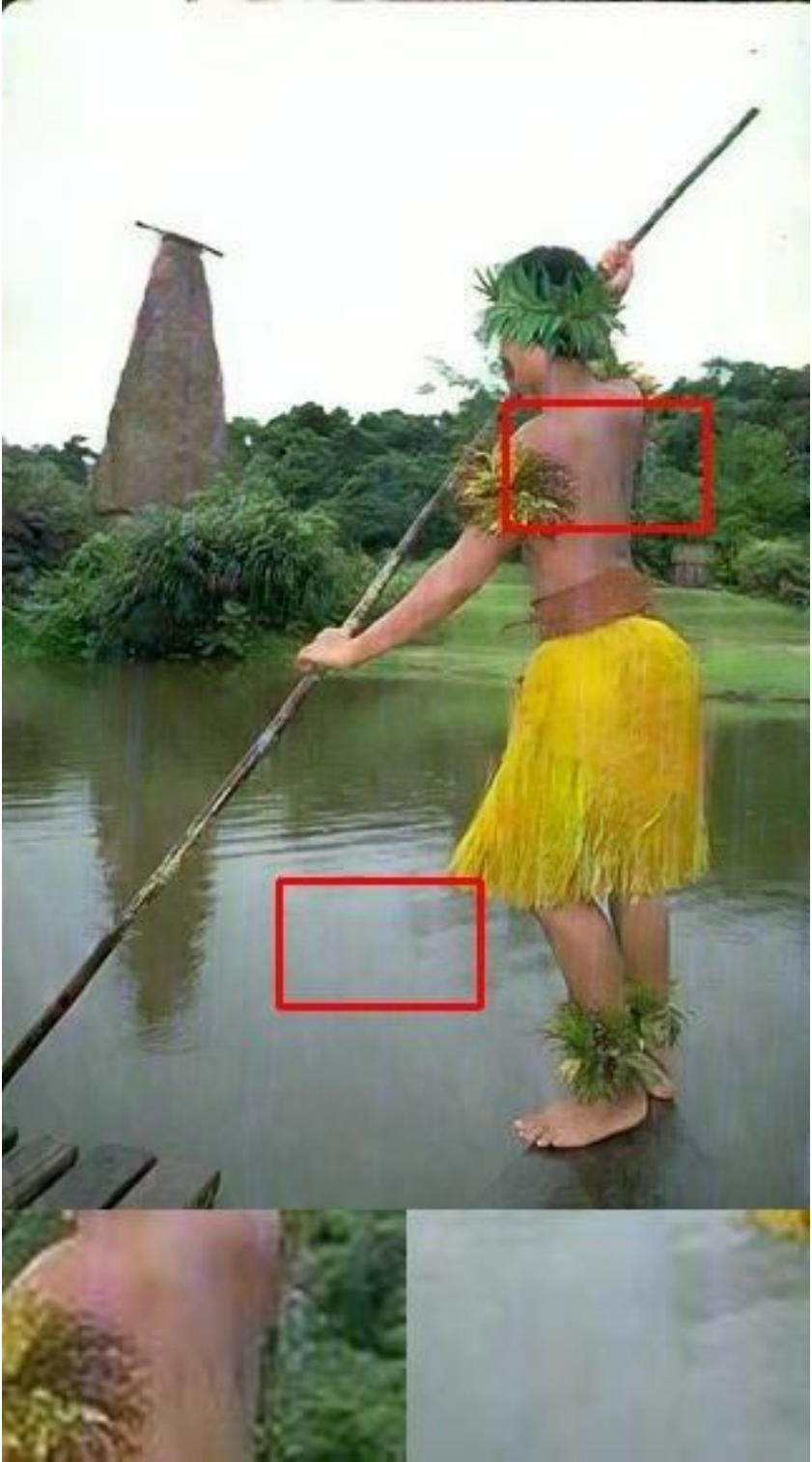}
		\subcaption{\scriptsize CVID$_2$}
	\end{subfigure}
	\begin{subfigure}[t]{0.115\textwidth}
		\includegraphics[width=1\textwidth]{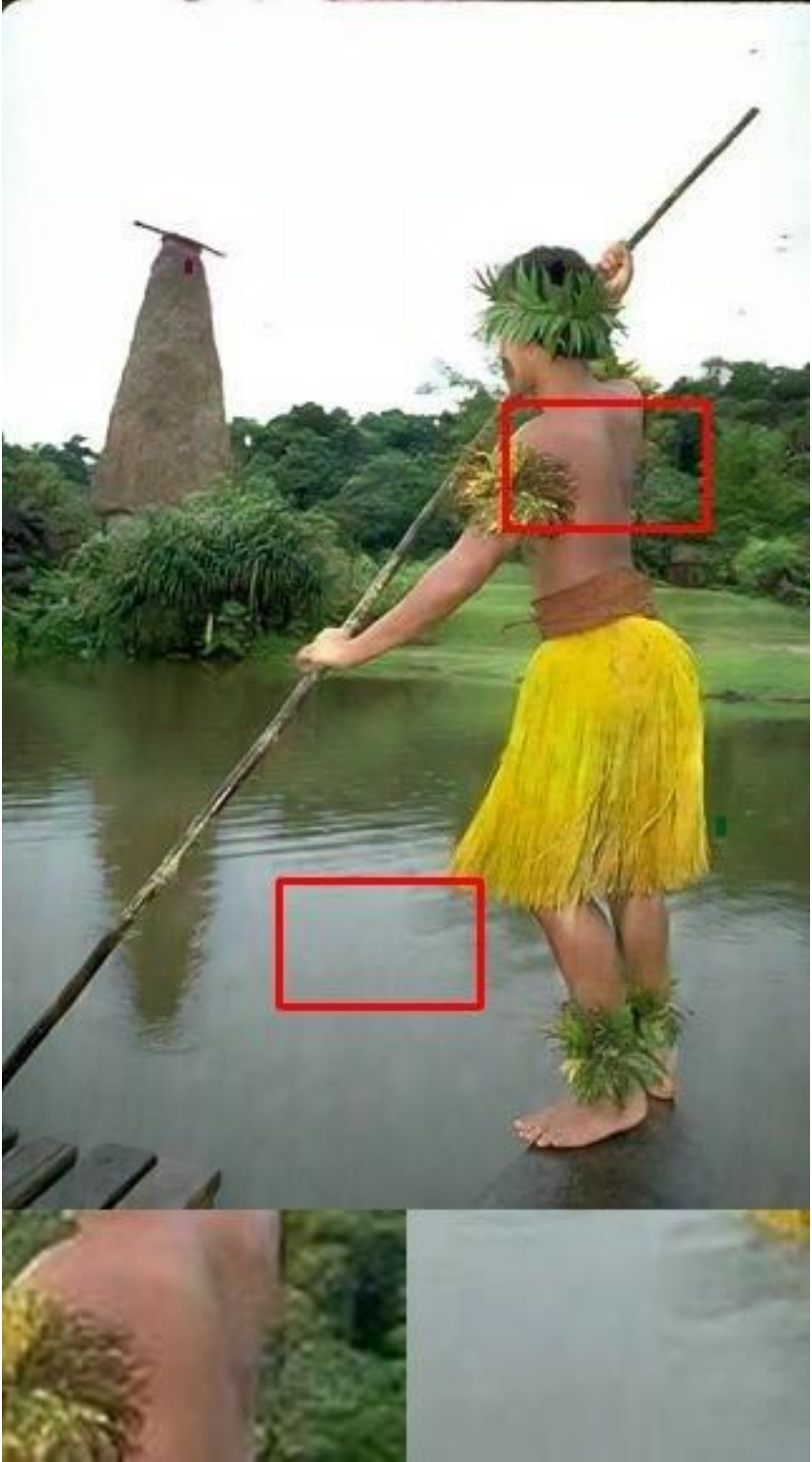}
		\subcaption{\scriptsize CVID$_{final}$}
	\end{subfigure}
	\caption{\textbf{Derained images of different image deraining methods}.\, The CVID$_{1}$ (f) and CVID$_{2}$ (g) are two candidate predictions by the proposed CVID network, and the CVID$_{final}$ (h) is their average as the final derained image.}
	\label{fig:example}
\end{center}
\vspace{-5mm}
\end{figure}

Despite their success on image deraining, these methods suffer from two major limitations.\, First, the real composition of rainy image cannot be fully reflected by the used composite models in current optimization based methods, and hence the regularizers are still insufficient in characterizing the background image and rain layer, limiting these methods from robust deraining performance on diverse scenarios.\, Second, the discriminative or generative learning based methods~\cite{JIN2020107143} mostly learn a mapping function and produce deterministic derained images (Figs.~\ref{fig:example}~(b)-(d)).\, However, it is difficult to define the optimal derained image for a real-world rainy image, due to the inherent ill-posed nature of image deraining.\,

\begin{figure*} [t]
	\begin{center}
	\includegraphics[width=1\textwidth]{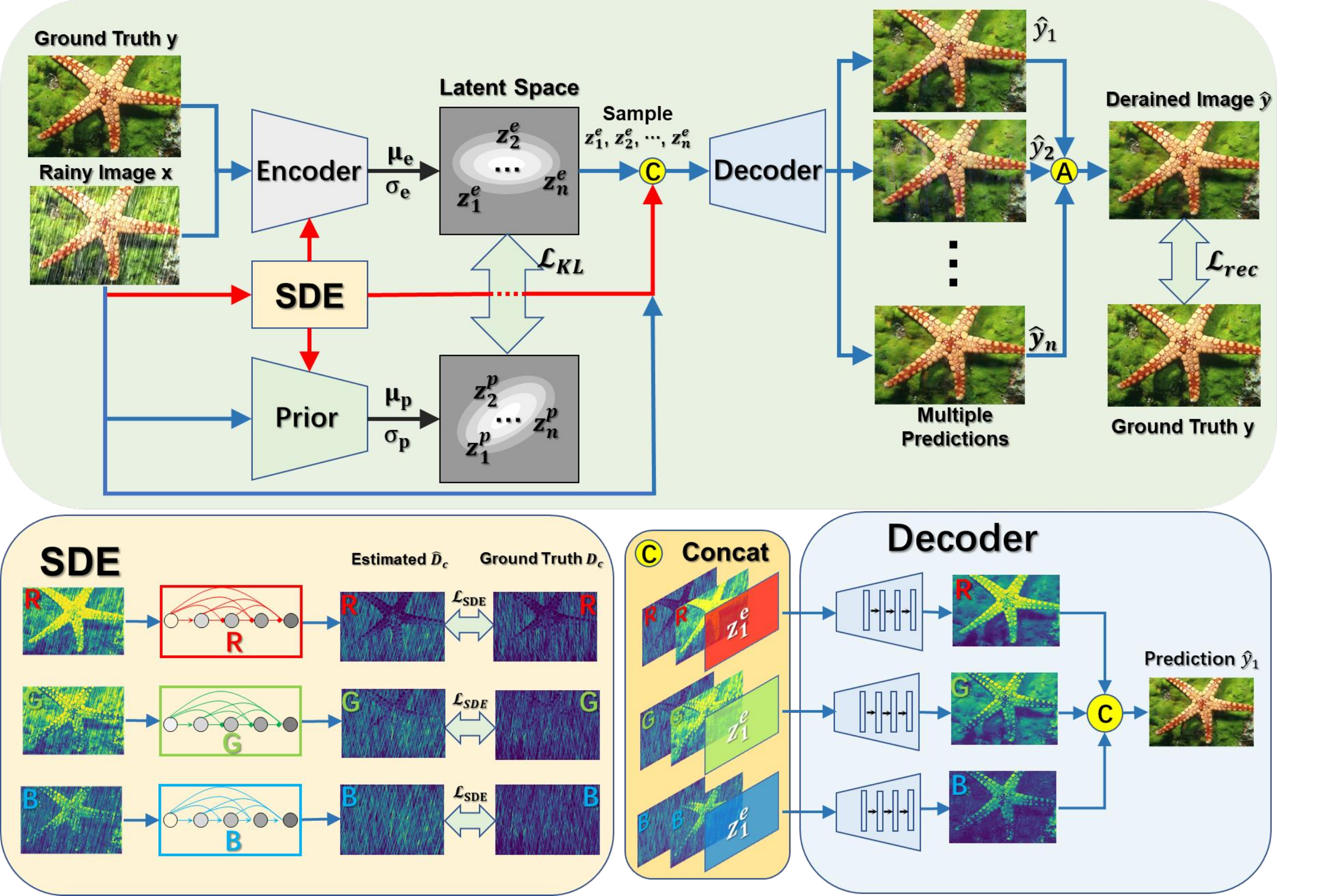}
	\vspace{-5mm}
    \caption{\textbf{The learning stage of CVID network}.\, The inputs of the encoder are the concatenation of rainy image $\mathbf{x}$, clean image $\mathbf{y}$, and estimated rain map $\hat{D}_c$ by the proposed SDE module.\, The decoder outputs the derained image $\mathbf{\hat{y}}$ based on estimated rain map by SDE, rainy image $\mathbf{x}$, and sampled $\{\mathbf{z}_{i}^e\}_{i=1}^{n}$ from the latent distribution $\mathcal{N}(\boldsymbol{\mu}_e, \boldsymbol{\sigma}_e)$.\, 
    The \textbf{black} arrows indicate the generation of latent space.\, 
    The blue arrows are the data flow of a standard CVAE model, while the red arrows are the additional data flow of our SDE module.
    Note that the encoder, SDE module, and decoder are performed channel-wisely \textsl{w.r.t.} the rainy image $\mathbf{x}$, clean image $\mathbf{y}$, and estimated rain map.\, 
    ``$\mathrm{C}$'' and ``$\mathrm{A}$'' denote the concatenation and averaging operations.}
	\label{fig:network}
	\end{center}
\vspace{-4mm}
\end{figure*}

\begin{figure*}[t]
	\includegraphics[width=1\textwidth]{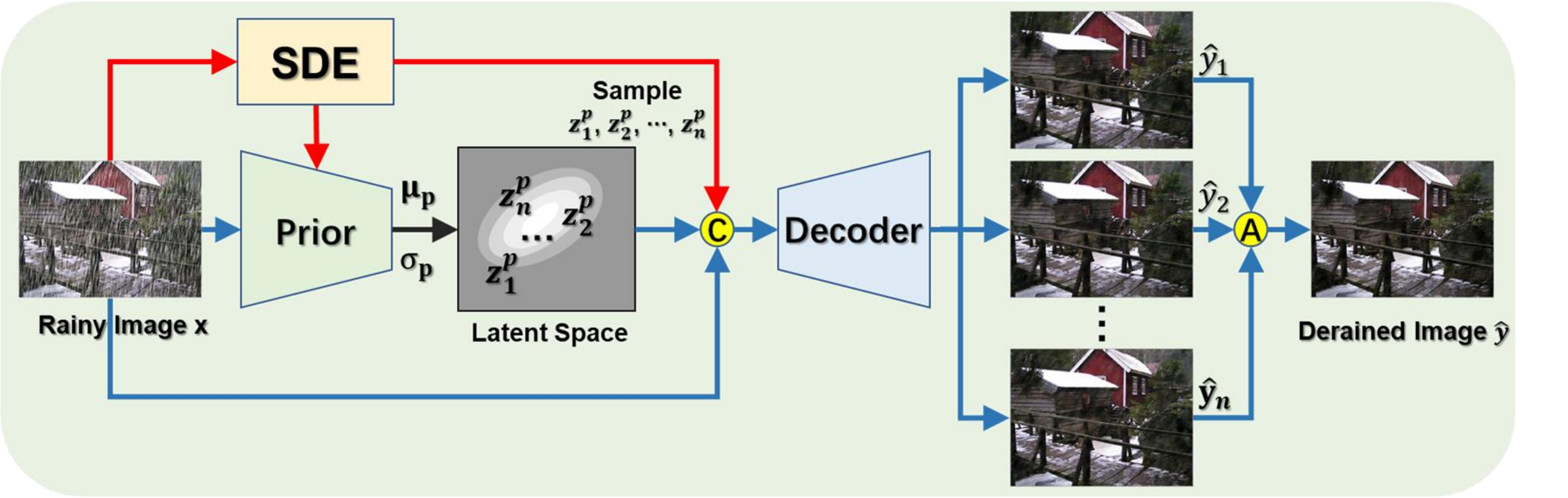}
    \caption{\textbf{The inference stage of CVID network.}\ We sample multiple latent variables $\mathbf{z}_{i}^p$ ($i=1, ..., n$) from the prior distribution $\mathcal{N}(\boldsymbol{\mu}_p, \boldsymbol{\sigma}_p)$ and adopt the Monte Carlo method~\cite{2007smcm.book} to perform a deterministic inference followed by weighted averaging.\, The \textbf{black} arrow indicates the process of generating latent space.\,
    ``$\mathrm{A}$'' and ``$\mathrm{C}$'' mean the averaging and concatenation operations.\, SDE is the spatial density estimation module, which will be introduced in \S\ref{sec:sde}.}
	\label{fig:inference}
    \vspace{-5mm}
\end{figure*}

In this paper, to address aforementioned challenges, we propose a Conditional Variational Image Deraining (CVID) network by leveraging the powerful generative capabilities of the recently developed Conditioanl Variational Auto-Encoder (CVAE)~\cite{sohn2015learning} framework.\, CVAE provides strong capability to model the latent distribution of image priors, from which the clean images can be generated.\, Instead of learning a deterministic mapping function in previous methods, we propose a CVAE based CVID network to simultaneously learn the latent representation of clean image priors and predict multiple possible derained images (Figs.~\ref{fig:example}~(f)-(g)).\, In the learning stage of CVID network, given pairs of clean and rainy images, the encoder learns to map these clean images into a latent distribution that shares common information for clean background images, while the decoder recover the derained images based on a sampled variable from the latent distribution space (as shown in Fig.~\ref{fig:network}).\, In the inference stage, we sample multiple latent variables from the latent distribution of clean image prior using the prior network, and adopt the Monte Carlo method~\cite{2007smcm.book} to perform deterministic prediction for each sampled variable (as shown in Fig.~\ref{fig:inference}).\, These predictions are averaged to produce the final derained image (Fig.~\ref{fig:example}~(h)).

One important observation of rainy images is that the rain streaks are usually unevenly distributed across the whole image, both in spatial locations and color channels (as shown in Fig.~\ref{fig:distribution}).\, Thus, how to remove the rain streaks in a spatially and channel-wisely adaptive manner should be handled seriously for image deraining.\, To tackle this problem, Zhang \textsl{et al.}~\cite{zhang2018density} proposed a density-aware deraining method by utilizing \textsl{global} rain density information.\, But this method produces inaccurate deraining results in \textsl{local} regions, and ignores the fact that rain distributions vary across different channels.\, To this end, we propose a spatial density estimation (SDE) module and a channel-wise (CW) scheme for more adaptive image deraining to our CVID network.\, The proposed SDE module and CW scheme jointly take a rainy image as input and output an estimated rain map for each channel, which indicates the intensity of rain on each pixel in each channel.\, Experiments on benchmark datasets demonstrate the advantages of our CVID network over previous contenders on image deraining, and the effectiveness of our SDE module and CW scheme.

In summary, our contributions are three-fold:
\begin{itemize}
    \item \textbf{A novel generative network which outputs multiple predictions for better image deraining performance}.\, We leverage the powerful Conditional Variational Auto-Encoder (CVAE) framework~\cite{sohn2015learning} for image deraining.\, The proposed CVID network effectively performs probabilistic deraining and produce multiple complementary derained predictions for better performance.\, As far as we know, our CVID network is the first work that tackles the image deraining problem under the CVAE framework.
    
    \item \textbf{Novel spatial attention module and channel-wise deraininig scheme}.\, We propose a spatial density estimation (SDE) module and a channel-wise (CW) scheme to endow our CVID network with the capability to perform spatially and channel-wisely adaptive deraining.\, Extensive ablation studies in \S\ref{Ablation} validate the effectiveness of the proposed SDE module and CW scheme.

    \item \textbf{Much better deraining performance on diverse datasets}.\, Experiments on three synthetic and one real-world rainy image datasets demonstrate that, the proposed CVID network achieves consistently superior results to previous state-of-the-art image deraining methods.\,
\end{itemize}

The rest of this paper is organized as follows.\, In \S\ref{sec:related}, we survey the related work.\, In \S\ref{sec:method}, we present the proposed CVID network for image deraining.\, Extensive experiments are conducted in \S\ref{sec:exp} to compare the proposed CVID network with state-of-the-art image deraining methods on synthetic and real-world image datasets.\, Conclusion is given in \S\ref{sec:con}.

%-------------------------------------------------------------------------

\section{Related Work}
\label{sec:related}
%\subsection{Single Image Deraining Methods}
In the past decade, numerous methods~\cite{fu2017clearing, zhang2017image, Wei_2019_CVPR} have been proposed to tackle the image deraining problem.\, Here, we briefly review the related work.

\noindent
\textbf{Optimization based methods}~\cite{kang2012automatic,chen2013generalized,ren2019pami} have been proposed for image deraining based on the fact that rainy images are composed of a clean background image layer and a rain layer.\, Image deraining
can be formulated by employing effective regularizers on both layers, and solved by proper optimization algorithms.\, Kang et al.~\cite{kang2012automatic} decomposed high frequency parts of rainy images into rainy and non-rainy components, and only processed the rainy component for rain streak removal.\, Luo et al.~\cite{luo2015removing} proposed a discriminative sparse coding framework based on image patches.\, Later, Chen et al.~\cite{chen2013generalized} proposed a low-rank appearance model for removing rain streaks.\, Similarly, Chang et al.~\cite{chang2017transformed} leveraged the low-rank property of rain streaks, which are removed via low-rankness based layer decomposition.\, However, since the real composition of rainy image cannot be fully explored by the composite models used in~\cite{kang2012automatic,chen2013generalized,luo2015removing}, the regularizers employed by these methods are insufficient in characterizing the background and rain layers, limiting these methods from robust deraining performance on diverse images.

\noindent
\textbf{Discriminative learning methods}.\, Recently, deep learning based methods have achieved promising performance on image deraining~\cite{fu2017clearing,li2018recurrent,Wei_2019_CVPR}.\, To the best of our knowledge, DerainNet~\cite{fu2017clearing} may be the first deep network developed for image deraining.\, Later, Deep Detail Network (DDN)~\cite{fu2017removing} was proposed to directly reduce the mapping range from input to output.\, The work of deep JOint Rain DEtection and Removal (JORDER) network~\cite{yang2017deep,yang2017deeppami} is developed for image deraining by using recurrent dilated networks.\, It can detect the rain region on each pixel of a rainy image, but does not reflect the density information of the pixels.\, The work of~\cite{zhang2017image} is also proposed for Density-aware Image Deraining using a Multi-stream Dense Network (DID-MDN).\, DID-MDN considers the global density of rain streaks, but ignores the density differences with respect to locally spatial location in the rainy image.\, The work of~\cite{li2018recurrent} introduced a recurrent squeeze-and-excitation context aggregation net (RESCAN) to tackle the problem of overlapping rain streak layers in image deraining.\, A simple baseline network is presented in~\cite{Ren_2019_CVPR} for single image deraining by preserving worthwhile deraining modules.\, A semi-supervised image deraining network is also developed in~\cite{Wei_2019_CVPR}.\, However, these methods do not consider the differences of rain density in different channels, and thus producing inaccurate results.

\noindent
\textbf{Generative methods}~\cite{li2016rain,zhang2017image,qian2018attentive} have also been developed for image deraining.\, Li et al.~\cite{li2016rain} proposed to use simple patch-based priors for both the background and rain layers.\, In~\cite{zhang2017image}, Zhang \textsl{et al.} utilized conditional generative adversarial networks (GANs) to prevent the background image from being degenerated of when extracted from rainy images, in which a learned discriminator network is employed as a guidance to synthesize rain-free images.\,
The work of~\cite{qian2018attentive} 
introduces the visual attention into both the generative and discriminative networks of GANs, and learns about raindrop regions and their surroundings for raindrop removal.\,

\noindent
\textbf{Conditional Variational Auto-encoder (CVAE)}~\cite{sohn2015learning} is a conditional generative model based on Variational Auto-encoder (VAE), which is originally proposed for structured prediction tasks, e.g., image segmentation and labelling.
Kohl \textsl{et al}.~\cite{kohl2018probabilistic} combined a U-Net~\cite{ronneberger2015u} with a CVAE that is capable of efficiently producing an unlimited number of plausible hypotheses, in order to handle inherent ambiguity of medical image segmentation.
Recently, Bao \textsl{et al}.~\cite{bao2017cvae} proposed the CVAE-GAN network to combine a VAE with a generative adversarial network, for fine-grained image generalization.
CVAE-GAN is essentially a conditional generative model, taking the fine-grained category label as input and generates images in a specific category. 
Ham \textsl{et al}.~\cite{Ham2018} incorporated perceptual loss into a VAE model, and demonstrated its effectiveness on image inpainting.
In our CVID, we use CVAE for supervised learning~\cite{sun2020learning}.
We model the latent distribution of clean images, and predict a specific clean image, which is treated as a condition, of the input rainy image.
In learning stage, we learn a prior network that takes a rainy image as input and estimates the latent distribution of its corresponding clean image.
In inference stage, with the sampled latent variable from the prior, the decoder network can recover the clean image from its rainy counterpart. 
In addition, we incorporate a density map estimation of the rainy image, as an extra condition to fully explore the power of CVAE for image deraining.
%\subsection{Video Deraining Methods}
%Compared single image deraining, video based methods additionaly leverage temporal information by analyzing the changes between adjacent frames.\, The work of Garg and Nayar~\cite{garg2004detection} is among the first for video deraining, which computes the average intensity of adjacent frames to remove rain from the static background~\cite{garg2004detection}.\, Other methods tackle the image deraining using Fourier techniques~\cite{barnum2010analysis}, Gaussian Mixture Models~\cite{bossu2011rain}, low-rank approximations~\cite{chen2013generalized}, or matrix completion algorithms\cite{kim2015video}.\, Recently, Ren \textsl{et al.}~\cite{ren2017video} divided rain streaks into sparse and dense ones, and proposed a matrix decomposition algorithm for video deraining.\, Liu \textsl{et al.}~\cite{liu2018erase} proposed a hybrid rain model to depict both rain streaks and  occlusions.\, In~\cite{d3rnet}, Liu \textsl{et al.} proposed a dynamic routing residue recurrent network for video deraining, integrating the hybrid model and useful motion segmentation context information.\, In this work, we focus on single image deraining task, and do not resort to temporal information in videos.
%Li \textsl{et al.}~\cite{li2018video} integrated multi-scale techniques into convolutional sparse coding framework for video rain streak removal.\,

\section{Learning Conditional Variational Image Deraining Network}
\label{sec:method}

In this section, we first present the proposed Conditional Variational Image Deraining (CVID) network developed under the Conditional Variational Auto-Encoder (CVAE) framework.\, In \S\ref{sec:Pre}, we provide the preliminaries of CVAE.\, Then we describe the learning stage of the proposed CVID network for image deraining in \S\ref{sec:dnet}.\, We present the proposed spatial density estimation (SDE) module and channel-wise (CW) deraining scheme in \S\ref{sec:sde} and \S\ref{sec:CID}, respectively.\, The optimization of CVID is provided in \S\ref{sec:optim} Finally, we introduce the inference stage of CVID in \S\ref{sec:Inf}

\subsection{Preliminaries on CVAE}
\label{sec:Pre}

Variational Auto-Encoder (VAE) is a powerful generative framework for learning the latent distribution of complex data~\cite{kingma2013auto, rezende2014stochastic, hoffman2013stochastic}.\, The generative process of a VAE is as follows: the encoder takes the data $\mathbf{x}$ as input and outputs a data-conditional distribution $q_{\phi}(\mathbf{z}|\mathbf{x})$ for a latent vector $\mathbf{z}$.\, A sample $\mathbf{z}\sim p_{\theta}$ is drawn from the code-conditional reconstruction distribution $p_{\theta}$, and then used by the decoder to determine the distribution $p_{\theta}(\mathbf{x}|\mathbf{z})$ over the input data $\mathbf{x}$.\, 
The objective of VAEs is to maximize the variational lower bound of $p_{\theta}(\mathbf{x})$:
\begin{equation}
\label{eq:vae}
\hspace{-3mm}
\log p_{\theta}(\mathbf{x})
\hspace{-1mm}
 \geq 
\hspace{-1mm}
 -D_{\rm KL}(q_{\phi}(\mathbf{z}|\mathbf{x})||p_{\theta}(\mathbf{z})) + \mathbb{E}_{q_{\phi}(\mathbf{z}|\mathbf{x})} \log p_{\theta}(\mathbf{x}|\mathbf{z}),
\end{equation}
where $D_{\rm KL}$ is the function of Kullback-Leibler (KL) divergence and $\mathbf{z} = g_{\phi}(\mathbf{x}, \boldsymbol{\epsilon}),\boldsymbol{\epsilon} \sim \mathcal{N}(\mathbf{0}, \mathbf{I})$.

Although VAEs have the innate capability of modeling latent distributions and preserving common features of clean images, it can only take in and output the same rainy image and cannot output a derained image from the rainy input.\, Thus, VAEs cannot be directly applied for image deraining.

Recently, Sohn et al.~\cite{sohn2015learning} have extended the VAEs to more powerful conditional VAEs (CVAEs), which model the latent variables and data, conditioned on side information, such as the clean image $\mathbf{y}$ of the rainy image $\mathbf{x}$.\, By taking the conditional information of clean image $\mathbf{y}$ into account, we can rewrite the lower bound of Eqn.~(\ref{eq:vae}) as: %to be maximized
\begin{equation}
\label{cvae}
\begin{aligned}
\mathcal{\widetilde L}_{\rm{CVAE}}= &-D_{\rm{KL}}(q_{\phi}(\mathbf{z}|\mathbf{x}, \mathbf{y})||p_{\theta}(\mathbf{z}|\mathbf{x}))
\\
&
+\mathbb{E}_{q_{\phi}(\mathbf{z}|\mathbf{x}, \mathbf{y})} \log p_{\theta}(\mathbf{y}|\mathbf{z}, \mathbf{x}),
\end{aligned}
\end{equation}
where $\mathbf{z}=g_{\phi}(\mathbf{x}, \mathbf{y}, \boldsymbol{\epsilon}), \boldsymbol{\epsilon} \sim \mathcal{N}(\mathbf{0}, \mathbf{I})$.\, Here, $p_{\theta}(\mathbf{z}|\mathbf{x})$ is assumed to be an isotropic Gaussian distribution and $p_{\theta}(\mathbf{x}|\mathbf{y}, \mathbf{z})$, while $q_{\phi}(\mathbf{z}|\mathbf{x}, \mathbf{y})$ are Gaussian distributions.\, 

Ever been introduced, CVAE has demonstrated its great power in diverse computer vision tasks, such as trajectory prediction~\cite{walker2016uncertain}, image colorization~\cite{deshpande2017learning}, image generation~\cite{esser2018variational}, and multi-modal human dynamic generation~\cite{yan2018mt}, \textsl{etc}.\,

Image deraining is a highly ill-posed problem, since it is non-trivial to define the optimal clean background for a real-world rainy image.\, CVAEs can generate multiple predictions of the derained image from the input rainy image.\, Thus, it is possible to obtain more accurate deraining results by integrating these predictions.
In this work, as far as we know, we are among the first to explore the generative capability of the CVAE model for single image deraining.

\subsection{Learning CVID Network for Image Deraining}
\label{sec:dnet}

In this work, we leverage the powerful generative ability of CVAEs as the backbone of our Conditional Variational Image Deraining (CVID) network for image deraining.\, 
The CVAE backbone is basicly consisted of an encoder, a prior network, and a decoder, as shown in Fig.~\ref{fig:network}.\,  
We set the filter size as $3$ and the number of convolution filters as $16$ in both the encoder and the prior network.\, In the last layer of the encoder and prior network, the first half is $\boldsymbol{\mu}$ and second half is $\boldsymbol{\sigma}$.\, We set the number of convolution filters as $1$.
For our CVID network,
we set the depth as $7$ for encoder, prior network and decoder. We employ the Leaky ReLU~\cite{xu2015empirical} as the activation function.\, Each layer is followed by Batch Normalization~\cite{ioffe2015batch}.
%and involves a clean image $\mathbf{y}$, a rainy image $\mathbf{x}$, and a latent variable $\mathbf{z}$.\, 

%
Conditioned on the rainy image $\mathbf{x}$, the encoder learns the latent distribution $\mathcal{N}(\boldsymbol{\mu}_e, \boldsymbol{\sigma}_e)$ that encrypts the information of the corresponding clean image $\mathbf{y}$.\,
To guarantee that the sampled latent variable $\mathbf{z}$ from the latent distribution and the input $\mathbf{x}$ are closely related during inference, we introduce a prior network (Fig.~\ref{fig:inference}) to make sure that the learned latent distribution is consistent with that obtained by inference.\, The prior network learns to map a rainy image $\mathbf{x}$ into a ``prior'' latent distribution $\mathcal{N}(\boldsymbol{\mu}_p, \boldsymbol{\sigma}_p)$ that encodes distribution information of the rainy image.\,
The goal of decoder is to reconstruct the derained image $\hat{\mathbf{y}}$ based on a sampled latent variable $\mathbf{z}_e$ from the ``encoder'' latent distribution $\mathcal{N}(\boldsymbol{\mu}_e, \boldsymbol{\sigma}_e)$, conditioned also on the rainy image $\mathbf{x}$.\,
Specifically, the input of decoder is the concatenation of the rainy image, the rain density estimation map, and the sampled $\mathbf{z}_e$. 
For the decoder, we set the filter size as $3$ and the number of deconvolution filters as $16$.
To compute the gradient more amenably, we use reparameterization techniques~\cite{kingma2013auto} to sample the latent variable $\mathbf{z}$ via $\mathbf{z} = \boldsymbol{\mu} (\mathbf{x}) + \boldsymbol{\epsilon} * \boldsymbol{\sigma} (\mathbf{x})$, where $\boldsymbol{\epsilon} $ is the sampled noise from a Gaussian distribution $\mathcal{N}(\mathbf{0}, \mathbf{I})$.

To learn the CVAE backbone network, we need to maximize the conditional variational lower bound defined in Eqn.\, (\ref{cvae}).\, The first term in Eqn.\, (\ref{cvae}) acts as a regularization term to minimize the difference between the data-conditional distribution $q_{\phi}(\mathbf{z}_e|\mathbf{x},\mathbf{y})$ and the prior distribution $p_{\theta}(\mathbf{z}_p|\mathbf{x})$.\, Here, we take Kullback-Leibler (KL) divergence as the penalty function to minimize the gap between the two Gaussian distributions $q_{\phi}(\mathbf{z_e}|\mathbf{x},\mathbf{y})$ and $p_{\theta} (\mathbf{z_p}|\mathbf{x})$.\, The second term in Eqn.\, (\ref{cvae}) is the reconstruction error measuring the information loss between the sampled latent code $\mathbf{z}_e$ and the clean image $\mathbf{y}$.\, We maximize the conditional log-likelihood $\mathbb{E}_{q_{\phi}(\mathbf{z_e}|\mathbf{x},\mathbf{y})}[\log p_{\theta}(\mathbf{\hat{y}}|\mathbf{x},\mathbf{z_e})]$ for accurate reconstruction.\, In practice, the error can be computed as the $\ell_2$ loss between the clean image $\mathbf{y}$ and the reconstructed image $\mathbf{\hat{y}}$.
%To mitigate the discrepancy in encoding of latent variables at learning and inference, we consider allocating larger weights on the negative KL divergence term of an objective function.

\noindent
\textbf{Loss for CVAE}.\,
The CVAE network is trained to maximize the conditional log-likelihood of the second term in Eqn.\, (\ref{cvae}).\, Since this objective function is intractable, we instead maximize its variational lower bound in Eqn.\, (\ref{cvae}).\, We minimize the KL divergence between the data-conditional distribution $q_{\phi}(\mathbf{z}|\mathbf{x}, \mathbf{y})$ and the prior distribution $p_{\theta}(\mathbf{z}|\mathbf{x})$, to mitigate the discrepancies between the encoding of latent variables at learning and inference stages:
\begin{equation}
    \mathcal{L}_{\rm KL} = \sum_{i=1}^{N} q_{\phi}(\mathbf{z}_i|\mathbf{x}_i, \mathbf{y}_i) \log(\frac{q_{\phi}(\mathbf{z}_i|\mathbf{x}_i, \mathbf{y}_i)}{p_{\theta}(\mathbf{z}_i|\mathbf{x}_i)}),
    \label{KL}
\end{equation}
where $q_{\phi}(\mathbf{z}_i|\mathbf{x}_i, \mathbf{y}_i)=\mathcal{N} (\boldsymbol{\mu}_e, \boldsymbol{\sigma}_e)$, $p_{\theta}(\mathbf{z}_i|\mathbf{x}_i)=\mathcal{N} (\boldsymbol{\mu}_p, \boldsymbol{\sigma}_p)$, and $N$ is the number of training images.

To maximize $\mathbb{E}_{q_{\phi}(\mathbf{z}|\mathbf{x},\mathbf{y})}[\log p_{\theta}(\mathbf{y}|\mathbf{x},\mathbf{z})]$ for the reconstruction of $\mathbf{x}$, we define the loss $\mathcal{L}_{\rm rec}$ 
%as the $\ell_2$ loss between clean image $\mathbf{y}$  and derained image $\mathbf{\hat{y}}$ 
as follows:
\begin{equation}
\setlength{\abovedisplayskip}{3pt}
    \mathcal{L}_{\rm rec} = \frac{1}{N}\sum_{i=1}^N \sum_{c\in\{r,g,b\}}|| \mathbf{y}_{i,c} - \mathbf{\hat{y}}_{i,c}||_{F}^{2},
\setlength{\belowdisplayskip}{3pt}
    \label{reconstruction}
\end{equation}
where $\mathbf{\hat{y}}_{i,c} = f^{\rm rec}_c(\mathbf{x}_{i,c}, \mathbf{y}_{i,c}, D_{i,c})$ is the CVAE associated with the $c$-th channel.\, The CVAE takes as the inputs each individual color channel of the rainy image $\mathbf{x}$, clean image $\mathbf{y}$ and the estimated rain density map $D_c$ in channel $c$ (Fig.~\ref{fig:network}, this part will be explained in \S\ref{sec:CID}), and outputs the derained image $\mathbf{\hat{y}}_c$ of channel $c$.\, In summary, the loss $\mathcal{L}_{\rm CVAE}$ is the sum of $\mathcal{L}_{\rm KL}$ and $\mathcal{L}_{\rm rec}$:
\begin{equation}
    \mathcal{L}_{\rm CVAE} = \mathcal{L}_{\rm rec} +\beta \mathcal{L}_{\rm KL},
    \label{cvae_loss}
\end{equation}
where $\beta > 0$ is a regularization parameter.

% algorithm---------------------------------------------------------------------------------------
\begin{algorithm}[t]
\small
\caption{Learning CVID Network for Image Deraining}
\label{alg:A}
\begin{algorithmic}
\STATE {\hrulefill}
\STATE {\textbf{Learning}: Input pairs of rainy and clean images $\left\{ \mathbf{x}_i, \mathbf{y}_i \right\}_{i=1}^N$}
\STATE {\qquad\qquad\ \ $\theta$, $\phi$  $\leftarrow$ Initialize parameters}
\REPEAT
\STATE {SDE: $\hat{D}_{i,c} \leftarrow  {\rm SDE}_{\theta}(\mathbf{x}_i)$}
\STATE {Encoder: 
$\left\{
        \begin{array}{l}
           \boldsymbol{\mu_e}, \boldsymbol{\sigma_e}  \leftarrow  E_{\phi}(\mathbf{x}_i, \mathbf{y}_i, \hat{D}_{i,c})  
           \\
           \mathbf{z}_{\bm{e}} \leftarrow \boldsymbol{\mu_e}(\mathbf{x}_i) + \boldsymbol{\epsilon} * \boldsymbol{\sigma_e}(\mathbf{x}_i),  \boldsymbol{\epsilon} \sim \mathcal{N} (\mathbf{0}, \mathbf{I})
        \end{array}
\right.$}

\STATE {Prior:
$\left\{
        \begin{array}{l}
           \boldsymbol{\mu_p}, \boldsymbol{\sigma_p} \leftarrow P_{\theta}(\mathbf{x}_i, \hat{D}_{i,c}) 
           \\
            \mathbf{z}_{\bm{p}} \leftarrow \bm{\mu}_p(\mathbf{x}_i) + \boldsymbol{\epsilon} * \boldsymbol{\sigma_p}(\mathbf{x}_i),  \boldsymbol{\epsilon} \sim \mathcal{N} (\mathbf{0}, \mathbf{I})
        \end{array}
\right.$}

\STATE {Decoder: $\mathbf{\hat{y}_i}  \leftarrow D_{\theta}(\mathbf{x}_i, \mathbf{z}, \hat{D}_{i,c})$ }
\STATE{$\mathbf{g}$ $\leftarrow$  $\nabla_{\theta, \phi}$$\mathcal{L}(\theta, \phi; \mathbf{x}, \mathbf{y}, \boldsymbol{\epsilon}$)}
\STATE{$\theta$, $\phi$ $\leftarrow$ Update parameters using gradients $\mathbf{g}$}
\UNTIL{convergence}

\RETURN{$\theta$, $\phi$}
\STATE {\hrulefill}
\STATE {\textbf{Inference}: Input rainy image $\mathbf{x}$}

\STATE {SDE:  $\hat{D}_c \leftarrow \mathrm{SDE}_{\theta}(\mathbf{x})$ }

\STATE {Prior:
$\left\{\begin{array}{l}
           \boldsymbol{\mu_p}, \boldsymbol{\sigma_p} \leftarrow P_{\theta}(\mathbf{x},  \hat{D}_c) \\
            \mathbf{z}_p \leftarrow \boldsymbol{\mu_p}(\mathbf{x}) + \boldsymbol{\epsilon} * \boldsymbol{\sigma_p}(\mathbf{x}),  \boldsymbol{\epsilon} \sim \mathcal{N} (\mathbf{0}, \mathbf{I})
        \end{array}
\right.$}

\STATE {Decoder: $\mathbf{\hat{y}}  \leftarrow  \frac{1}{n} \sum\limits^{n}_{j=1} D_{\theta}(\mathbf{x}, \mathbf{z}, \hat{D}_c)$}
\RETURN {Derained image $\mathbf{\hat{y}}$ }
\end{algorithmic}
\end{algorithm}

\begin{figure*}[ht] 
\begin{center}
    \begin{subfigure}{0.24\textwidth}
		\includegraphics[width=1\textwidth]{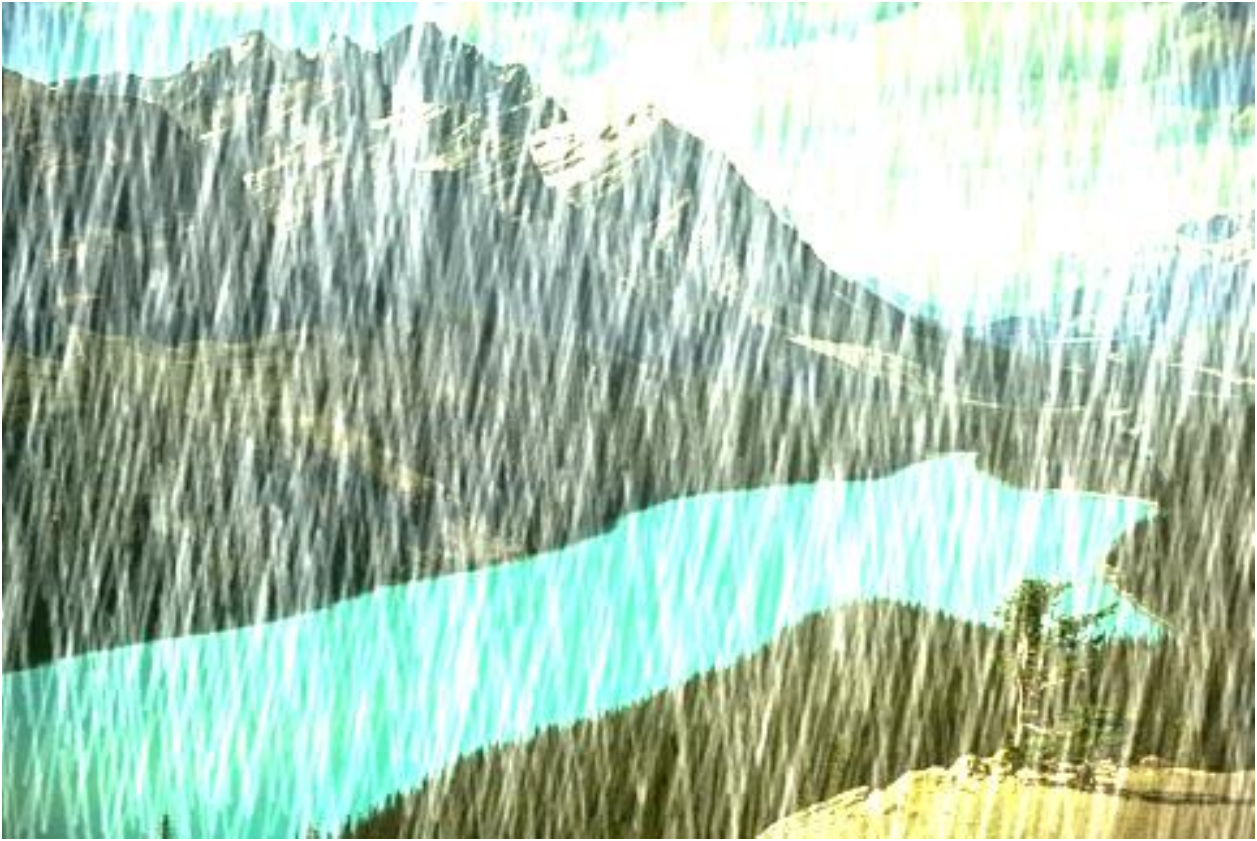}
		\subcaption{\footnotesize Synthetic rainy image}
	\end{subfigure}	
	\begin{subfigure}{0.24\textwidth}
		\includegraphics[width=1\textwidth]{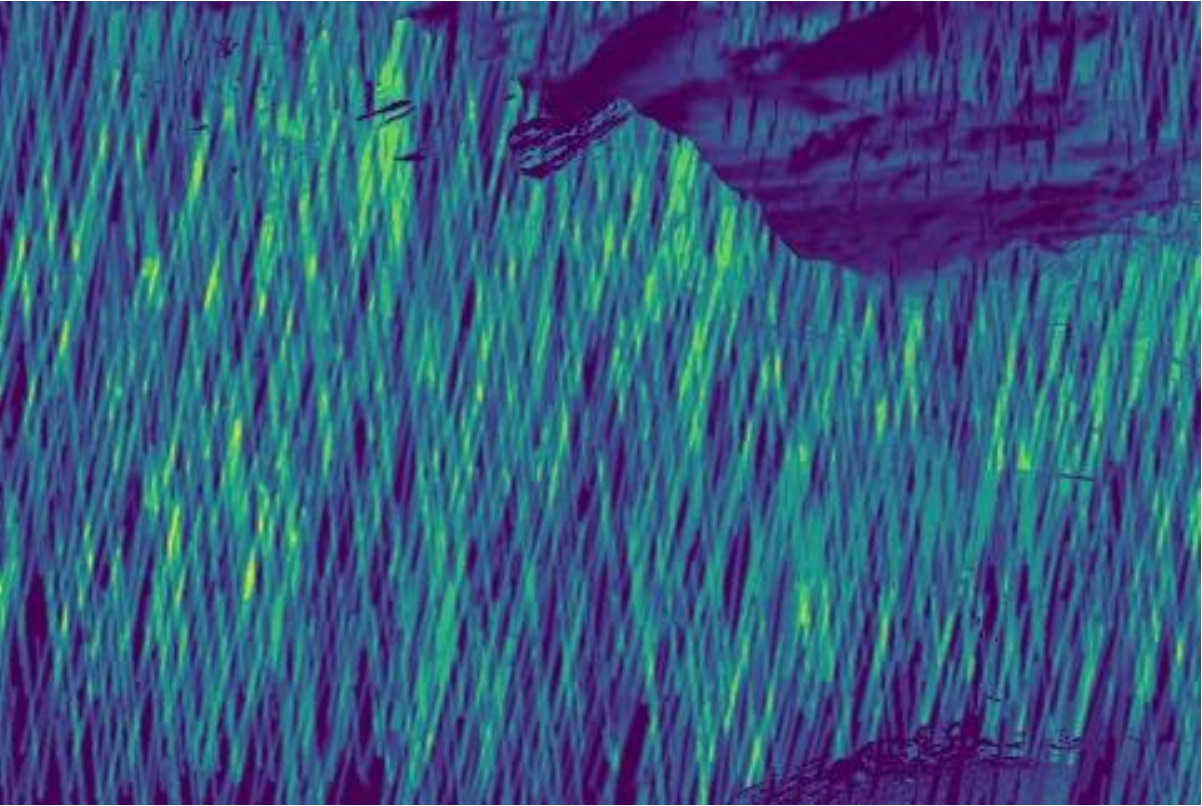}
		\subcaption{Rain map in R channel}%$D_r$
	\end{subfigure}	
	\begin{subfigure}{0.24\textwidth}
		\includegraphics[width=1\textwidth]{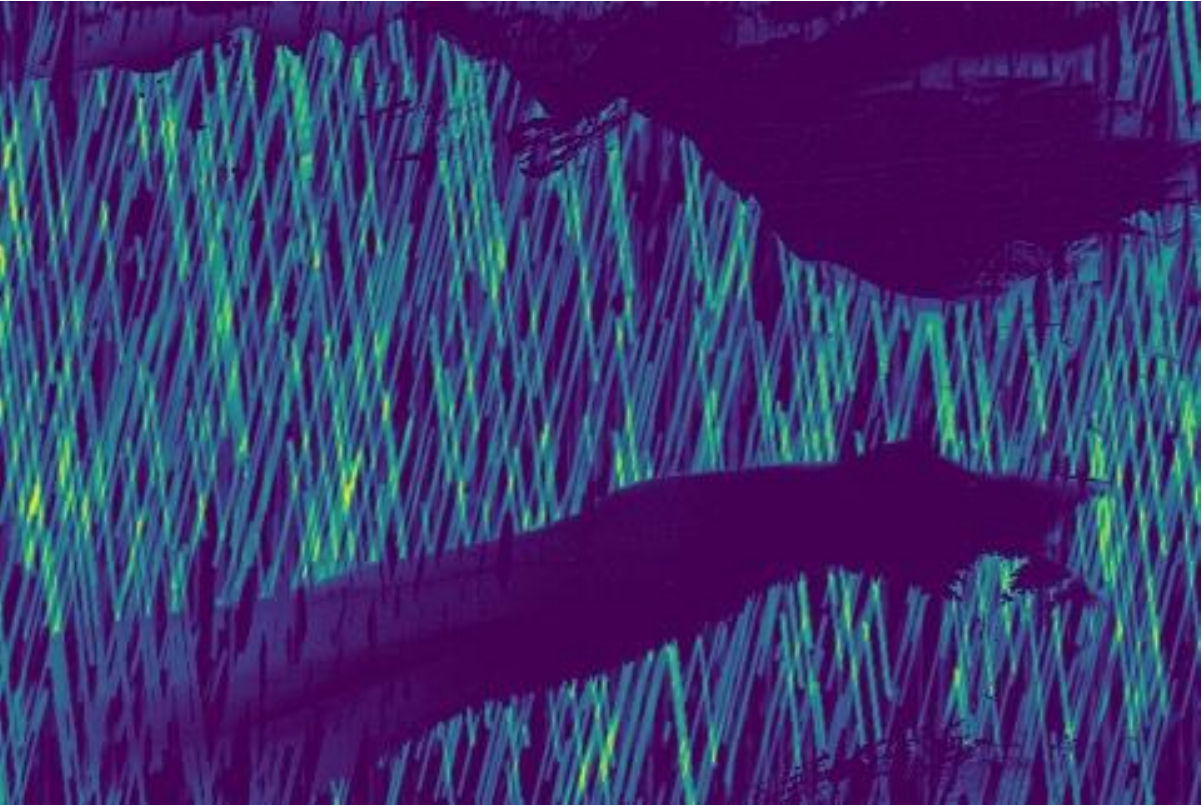}
		\subcaption{Rain map in G channel}% $D_g$
	\end{subfigure}	
	\begin{subfigure}{0.24\textwidth}
		\includegraphics[width=1\textwidth]{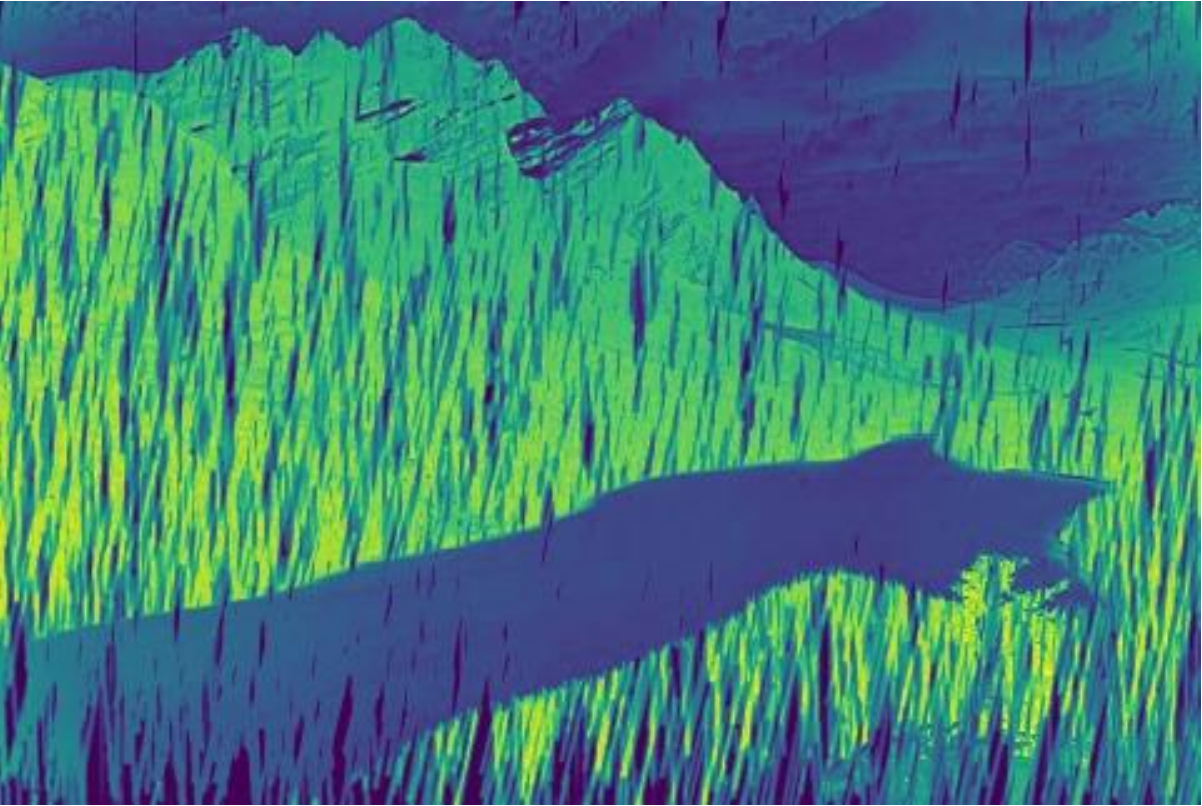}
		\subcaption{Rain map in B channel}% $D_b$
	\end{subfigure}
	\begin{subfigure}{0.24\textwidth}
		\includegraphics[width=1\textwidth]{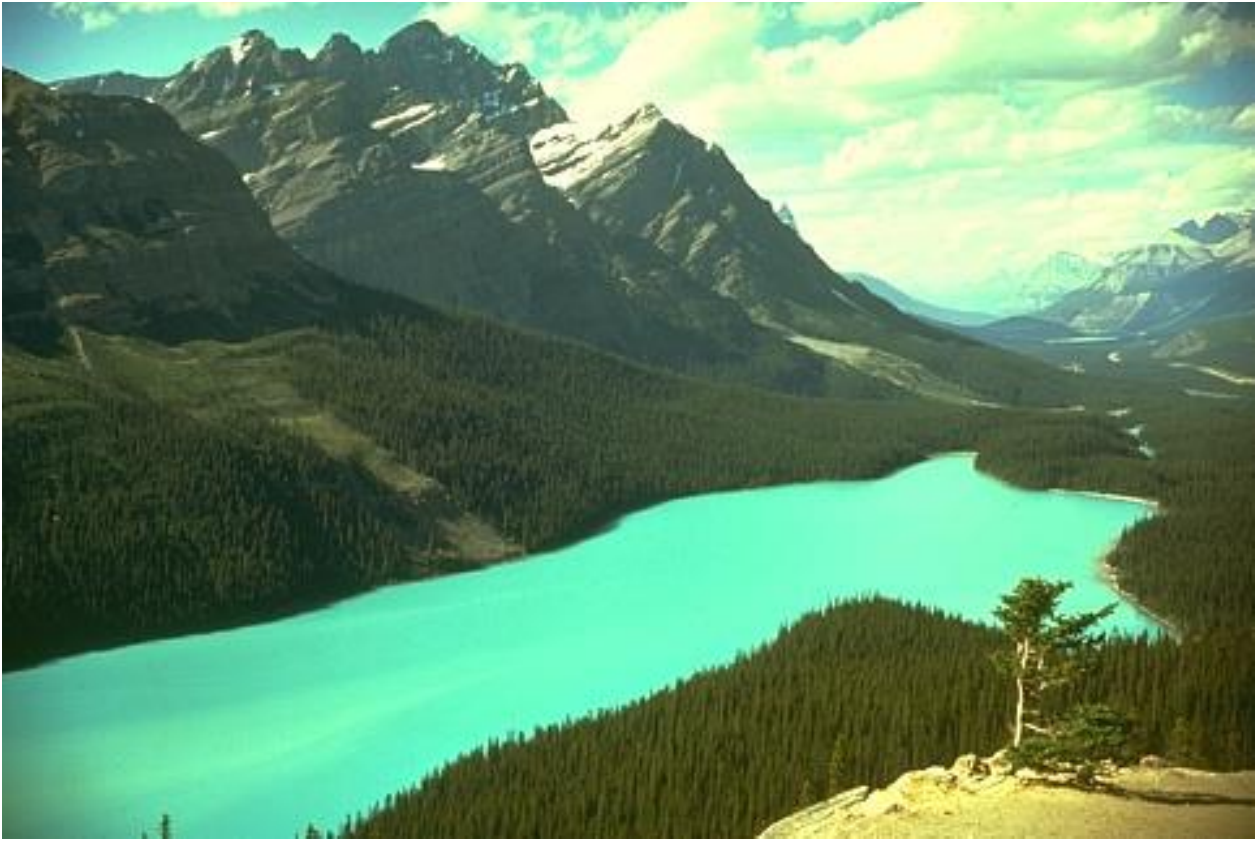}
		\subcaption{Ground truth}
	\end{subfigure}	
    \begin{subfigure}{0.24\textwidth}
		\includegraphics[width=1\textwidth]{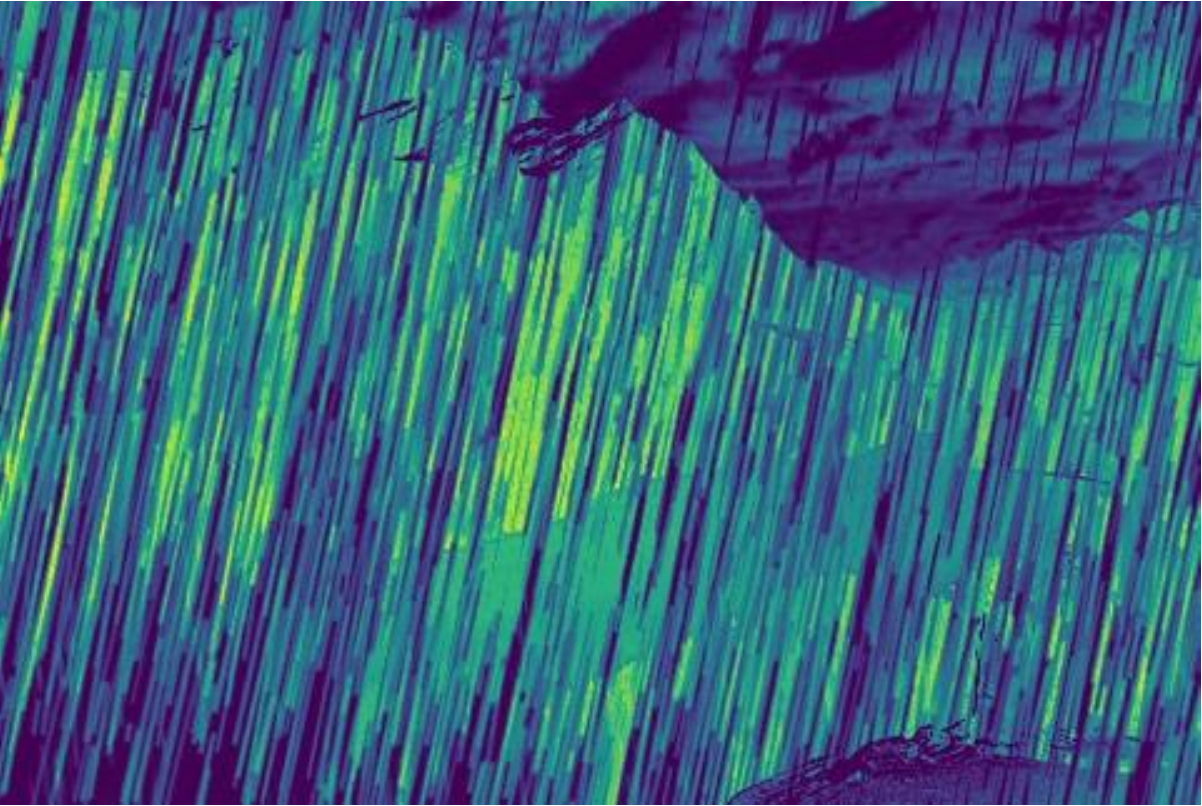}
		\subcaption{Estimated rain in R channel}%$\hat D_r$
	\end{subfigure}	
	\begin{subfigure}{0.24\textwidth}
		\includegraphics[width=1\textwidth]{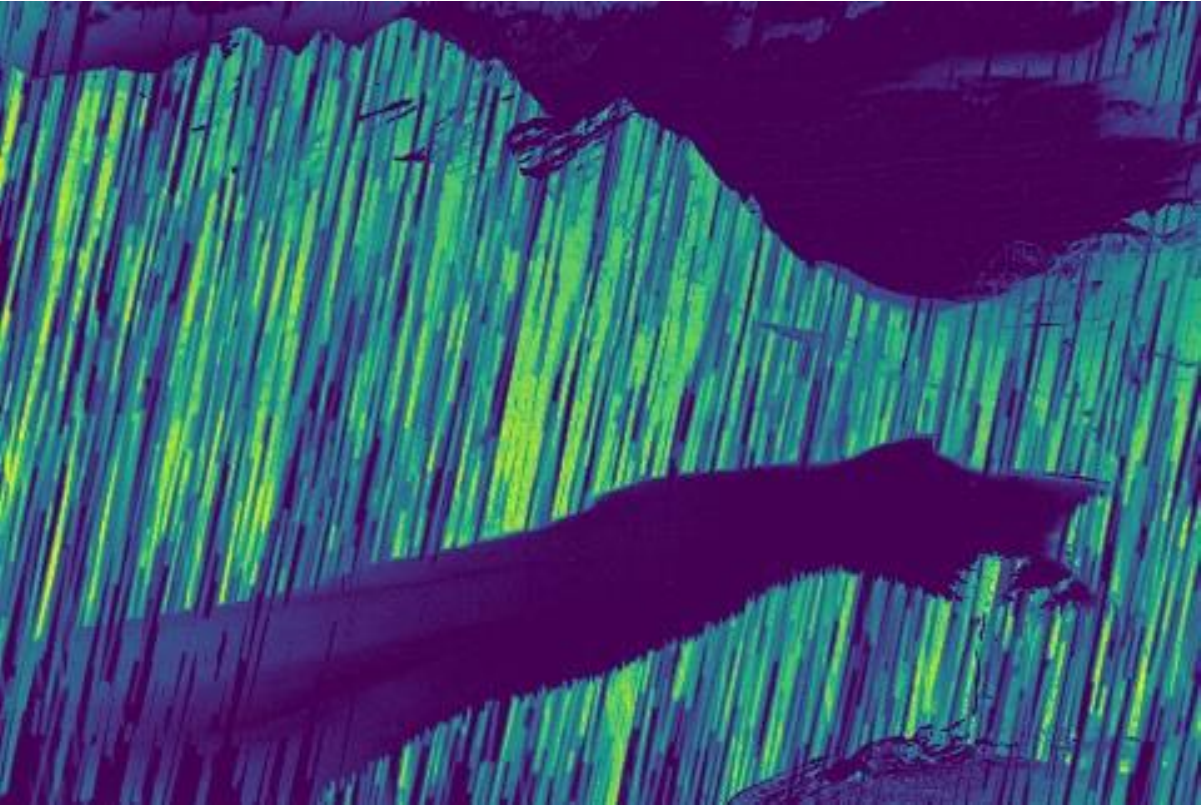}
		\subcaption{Estimated rain in G channel}%$\hat D_g$
	\end{subfigure}	
	\begin{subfigure}{0.24\textwidth}
		\includegraphics[width=1\textwidth]{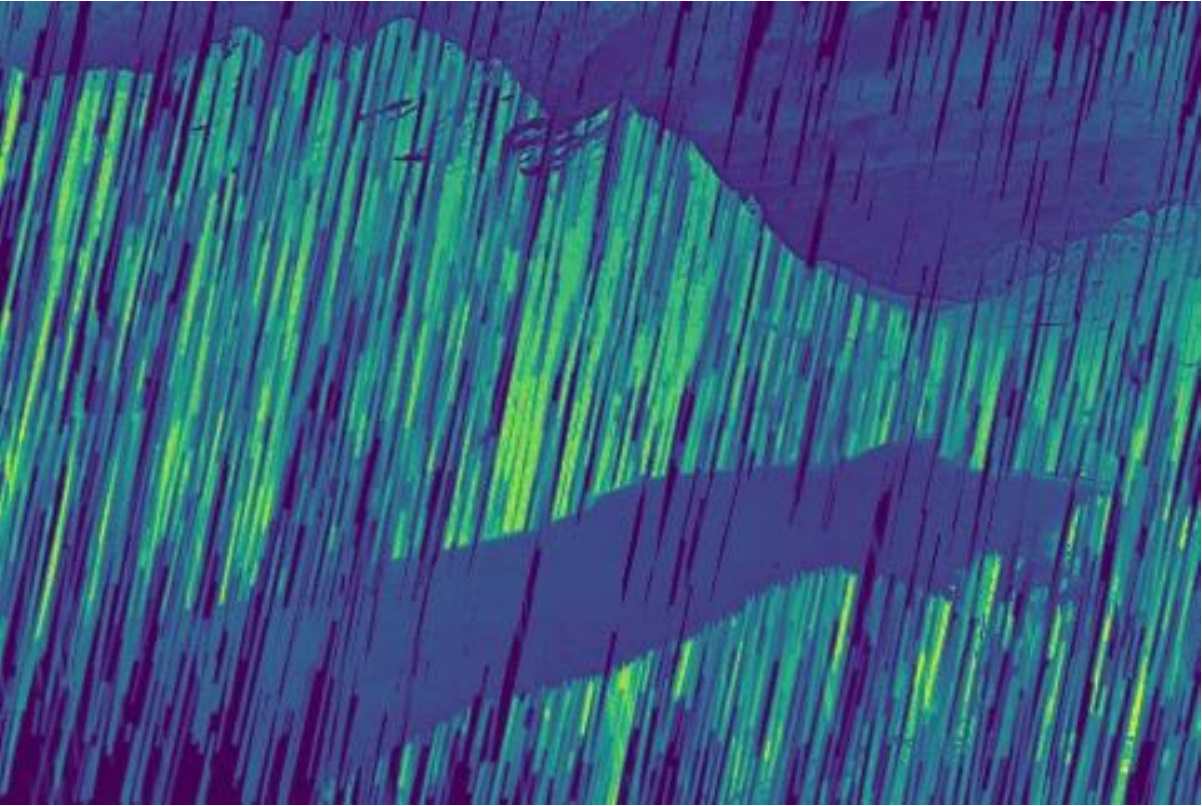}
		\subcaption{Estimated rain in B channel}%$\hat D_b$
	\end{subfigure}
\end{center}
    \vspace{-3mm}
	\caption{\textbf{Illustration of rain maps and estimated rain density maps on $R$, $G$, $B$ channels, respectively}.\, (a) and (e) are clean and corresponding rainy images, respectively.\, (b), (c) and (d) show rain maps on $R$, $G$, $B$ channels, respectively.\, (f), (g) and (h) are the estimated density maps by the proposed SDE module for three color channels, respectively.}
	\label{fig:distribution}
    \vspace{-3mm}
\end{figure*}

\subsection{Proposed Spatial Density Estimation Module}
\label{sec:sde}
The rain streaks are usually unevenly distributed in a rainy image.\, The methods ignoring the spatial variance of rain distribution will inevitably produce inaccurate deraining results.\, Although global density estimation is considered in~\cite{zhang2018density} by grading rain strength into different levels, inaccurate deraining results are still unavoidable in local regions.\, Besides, since the rain streaks are randomly distributed in the rainy image, it is difficult to locate rainy regions consistently in diverse images.\, 
%This challenge largely hinders the performance of current image deraining methods~\cite{fu2017removing, yang2017deep, zhang2018density, li2018recurrent, Wei_2019_CVPR, Ren_2019_CVPR}.

% \label{anet}
% \begin{figure}[t] \small
% 	\centering
% 		\includegraphics[width=0.48\textwidth]{desnse-crop.pdf}	
% 	\caption{\textbf{Spatial density estimation module.}}
% 	\label{fig:dense}
% \end{figure}

In this work, we propose a spatial density estimation (SDE) module for our CVAE backbone network, to learn a density estimation map for the input rainy image and make it spatially adaptive for image deraining.\, %In this way, the pixels with strong rain streaks will be restored with more strength, while those with weak rain streaks will be restored with less strength.\, 
The proposed SDE module is implemented as a compact densely-connected convolutional block with five layers~\cite{huang2017densely}.\,
The input of each layer is obtained by concatenating the output of all previous layers.
The filter size is set as $3$ and the number of filters as $16$.\, 
Each convolutional layer is followed by the batch normalization (BN)~\cite{ioffe2015batch} and the ReLU~\cite{maas2013rectifier} activation operations.\, For the last layer, we use the Sigmoid activation function to make the density estimation maps within $[0, 1]$.\, 

The learning of the density estimation maps is performed in a fully supervised manner.\,
It takes the whole rainy image as input and outputs one density estimation map for each color channel. 
Specifically, we subtract a rainy image $\mathbf{x}$ from its corresponding clean image $\mathbf{y}$ (``ground truth''), and produce a residual image denoted as $\hat{Re}$.\, $\hat{Re}_c$ indicates the $c$-th color channel of $\hat{Re}$, where $c\in\{R, G, B\}$, and $\hat{Re}_c(x)$ denotes a pixel value of position $x$ on each channel.\, $\hat{Re}_c(x)=0$ indicates that there is no rain at position $x$, while $\hat{Re}_c(x)\ne0$ indicates the intensity of rain at this pixel.\, Based on the residual map $\hat{Re}$, we generate the ground truth image for supervised learning of density estimation maps using $ D_c(x)$, 
\begin{equation}
\label{att-label}
D_c(x) =
    \left\{
        \begin{array}{cc}
          0 & \hat{Re}_c(x) = 0 \\
          \sigma(\hat{Re}_{c}(x)) & \hat{Re}_c(x)\ne0
        \end{array},
    \right
    .
\end{equation}
where $D_c$ is the ground truth for the $c$-th channel, $\sigma$ is the Sigmoid activation function.

We plot an example of $D_c$ in Figs.~\ref{fig:distribution} (b), (c), and (d) for the $R$, $G$ and $B$ channels of a rainy image, respectively.\, As can be seen, the rain streaks are distributed randomly across spatial locations and the three channels.\, The main reason is that, the light emitted from different sources are in different strength, such as the sunlight, the white floors, and the green water.\, We also plot the density maps estimated by the proposed SDE module in Figs.~\ref{fig:distribution} (f), (g), and (h), for the $R$, $G$, and $B$ channels of the rainy image, respectively.\, As can be seen, the density maps are very close to those of the ground truth maps shown in Figs.~\ref{fig:distribution} (b), (c), and (d).\, This indicates that the SDE module can accurately locate the rain regions of three color channels.\, Therefore, the proposed SDE module embeded CVAE network can obtain adaptive deraining performance by accurately estimating the rain strength via the density maps.

\noindent
\textbf{Loss for SDE}.\,
The SDE module is also trained in a fully supervised manner.\, It takes the rainy image $\mathbf{x}$ as input and estimates the density maps $D_c$ ($c \in \{R, G, B\} $) for each color channel.\, The loss function of the SDE module is as follows:
\begin{equation}
\mathcal{L}_{\rm SDE} = \frac{1}{N}\sum_{i=1}^N \sum_{c\in\{R, G, B\}}||D_{i,c} - \hat{D}_{i,c}||_{F}^{2},
\label{sde_loss}
\end{equation}
where $\hat{D}_{i,c} = f^D_c(\mathbf{x}_{i, c})$ and $f^D_c(\cdot)$ is the SDE module associated with the $c$-th channel.\, Penalizing the loss Eqn.\, (\ref{sde_loss}) aims to minimize the difference between the estimated density maps and ground truth maps.\, The obtained density estimation maps are then input to the CVAE backbone network for adaptive deraining performance on diverse local regions.

\subsection{Proposed Channel-wise Deraining Scheme}
\label{sec:CID}

\begin{figure*}[t]
	\centering	
	\begin{subfigure}[t]{0.24\textwidth}
		\includegraphics[width=1\textwidth]{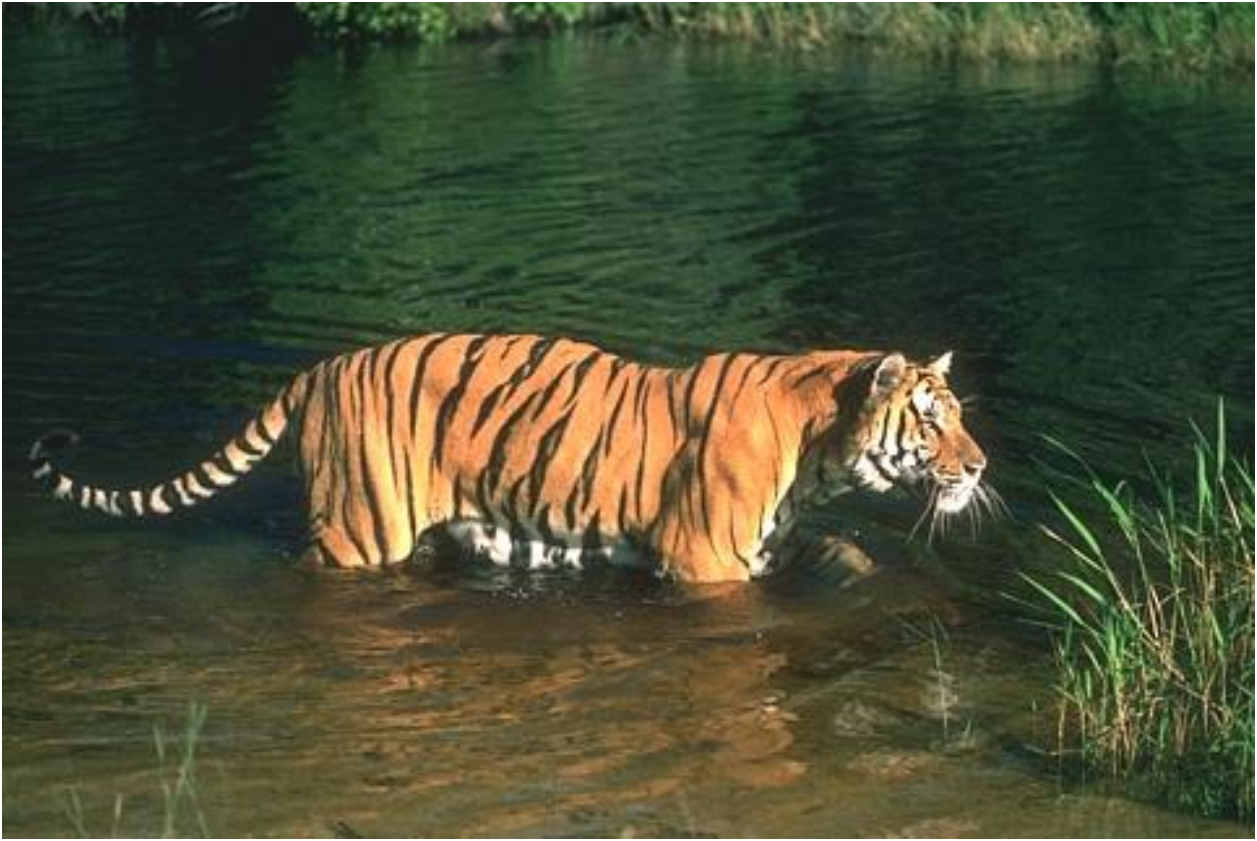}
	\end{subfigure}	
	\begin{subfigure}[t]{0.24\textwidth}
		\includegraphics[width=1\textwidth]{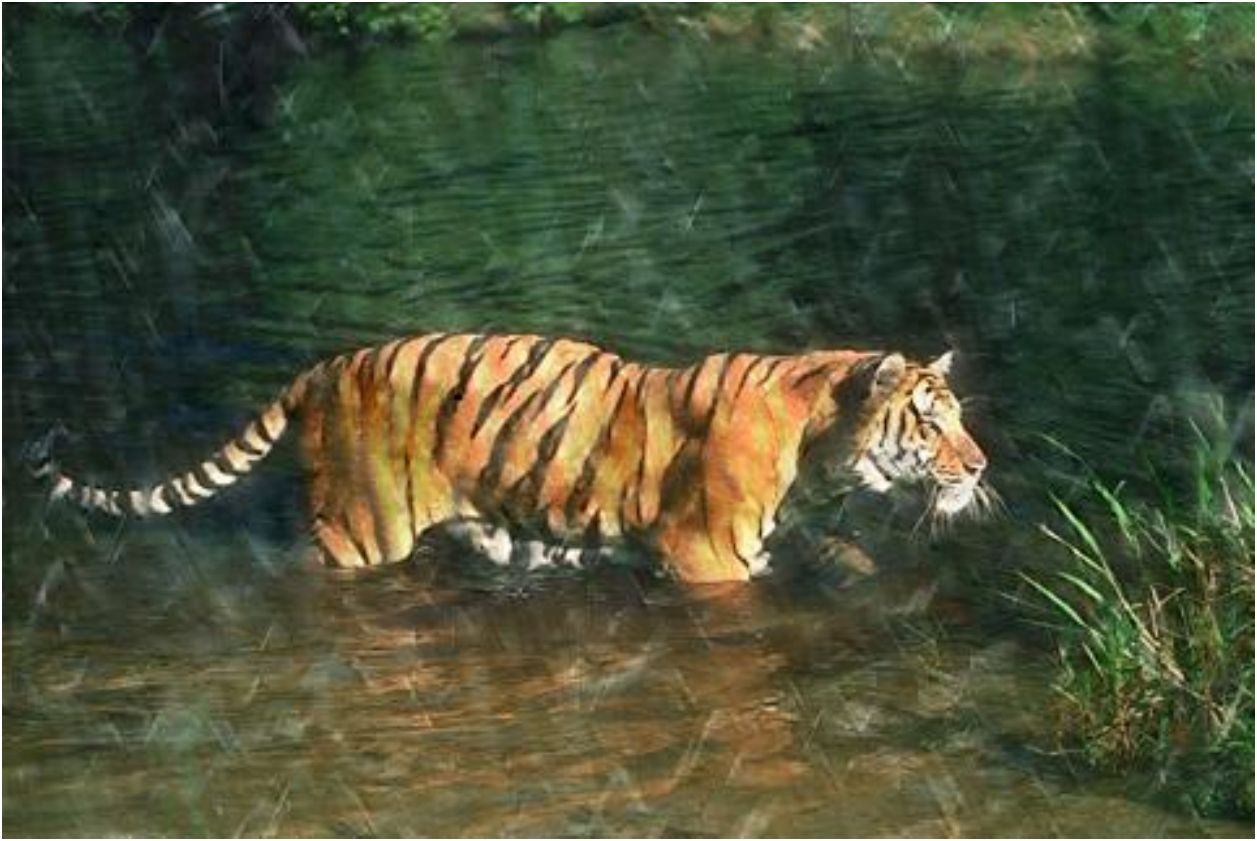}
	\end{subfigure}	
	\begin{subfigure}[t]{0.24\textwidth}
		\includegraphics[width=1\textwidth]{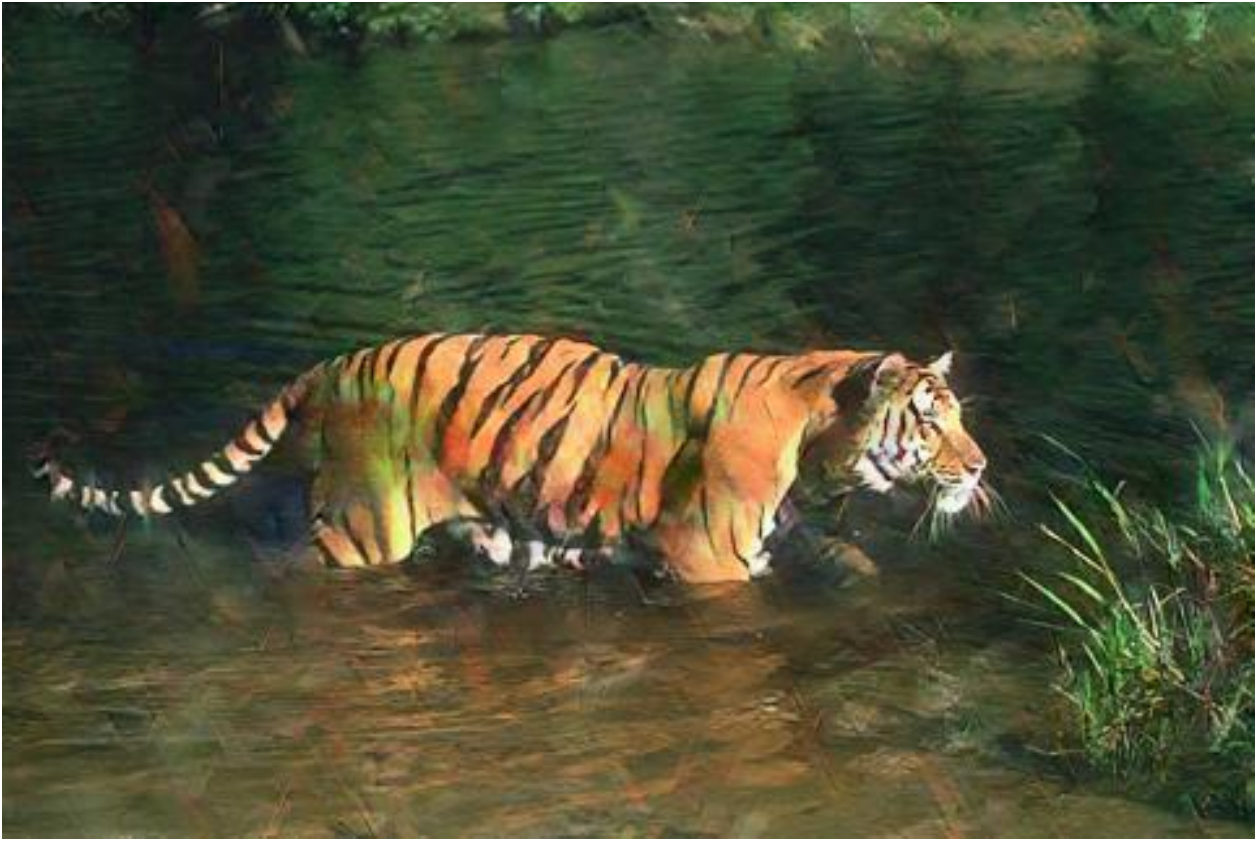}
	\end{subfigure}
	\begin{subfigure}[t]{0.24\textwidth}
		\includegraphics[width=1\textwidth]{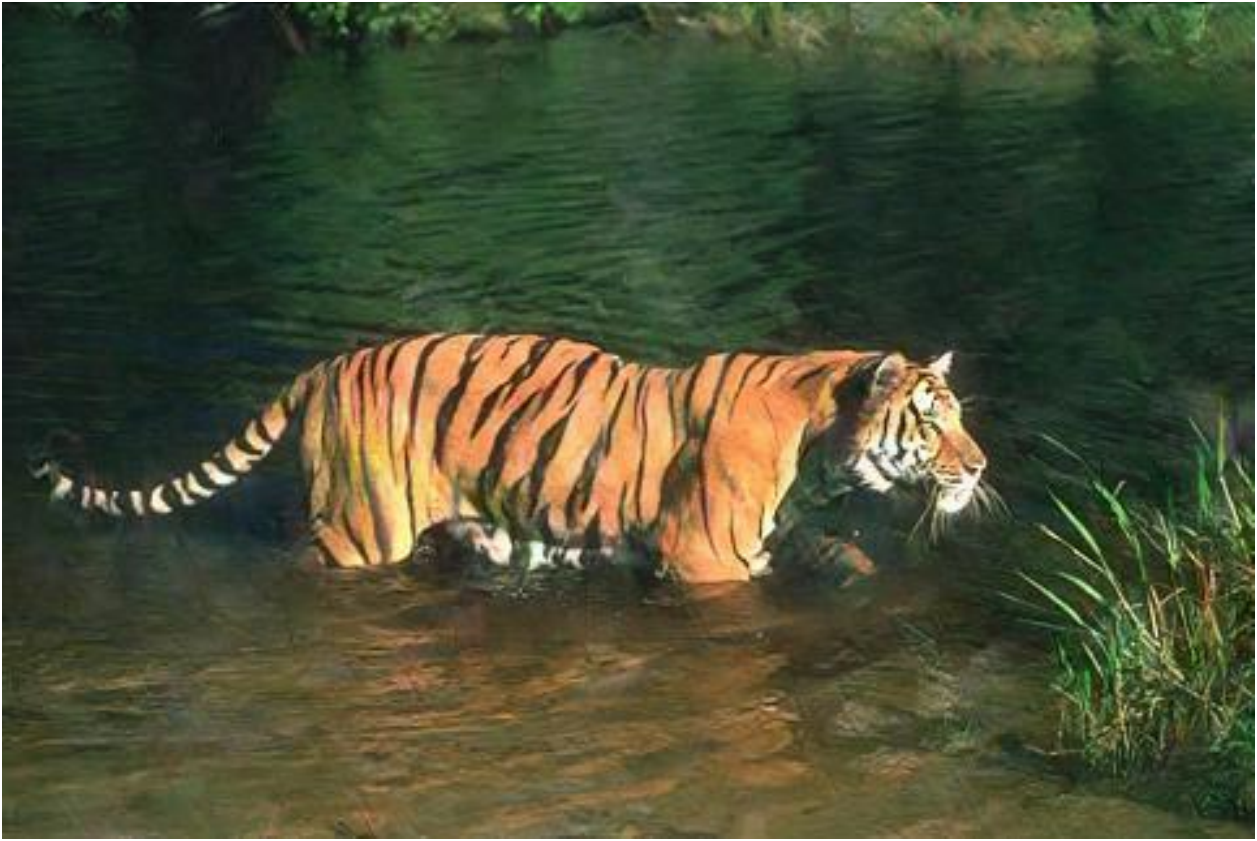}
	\end{subfigure}	
    \begin{subfigure}[t]{0.24\textwidth}
		\includegraphics[width=1\textwidth]{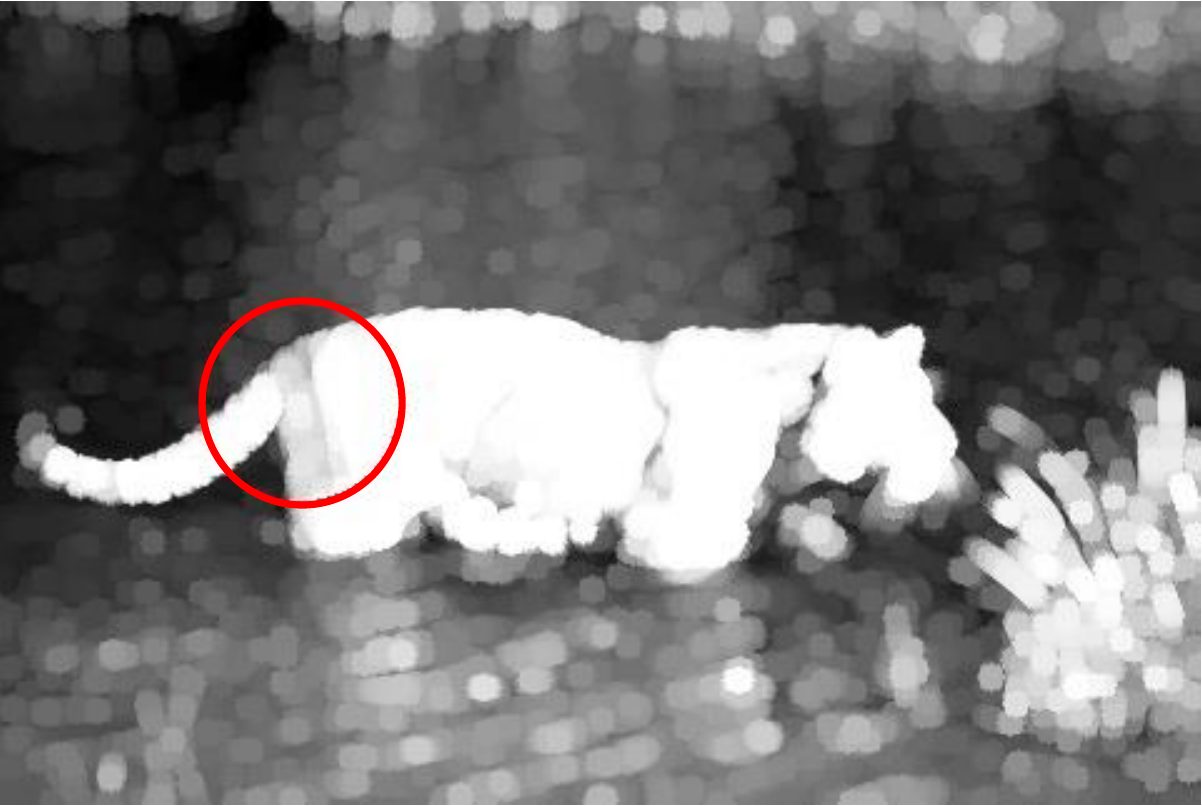}
	\end{subfigure}	
	\begin{subfigure}[t]{0.24\textwidth}
		\includegraphics[width=1\textwidth]{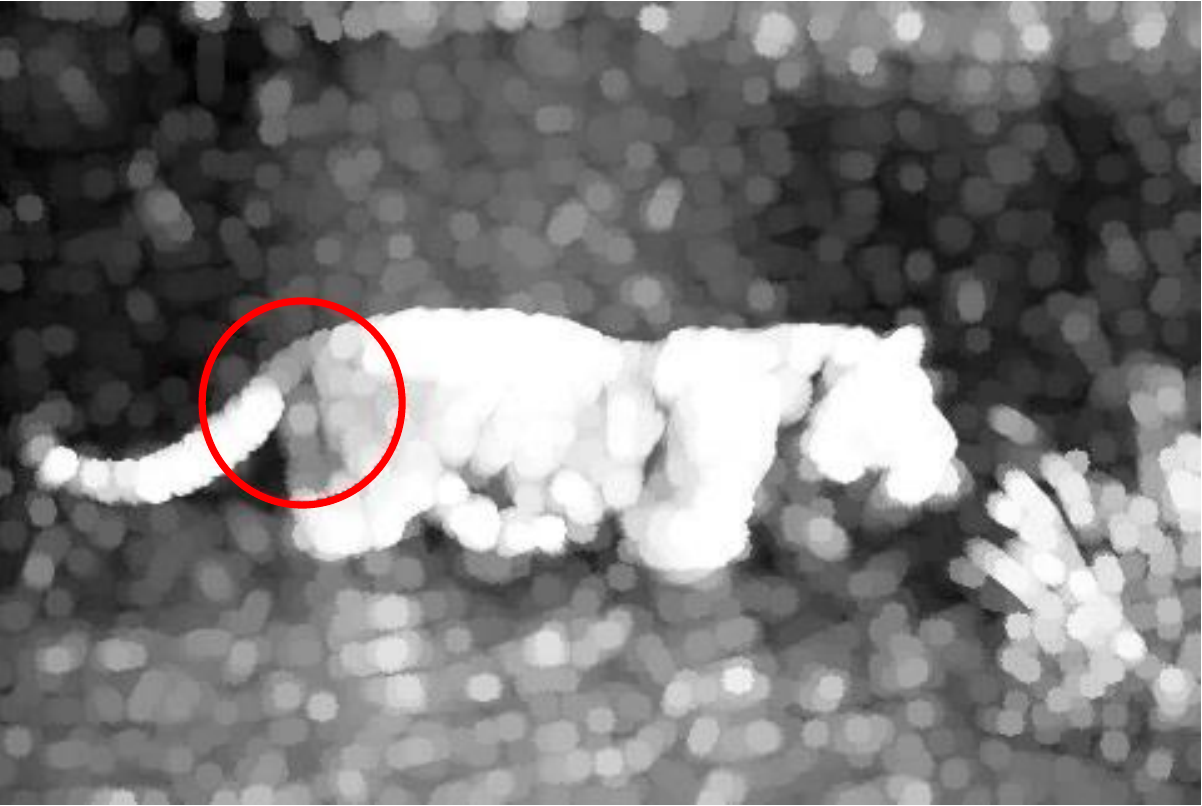}
	\end{subfigure}	
	\begin{subfigure}[t]{0.24\textwidth}
		\includegraphics[width=1\textwidth]{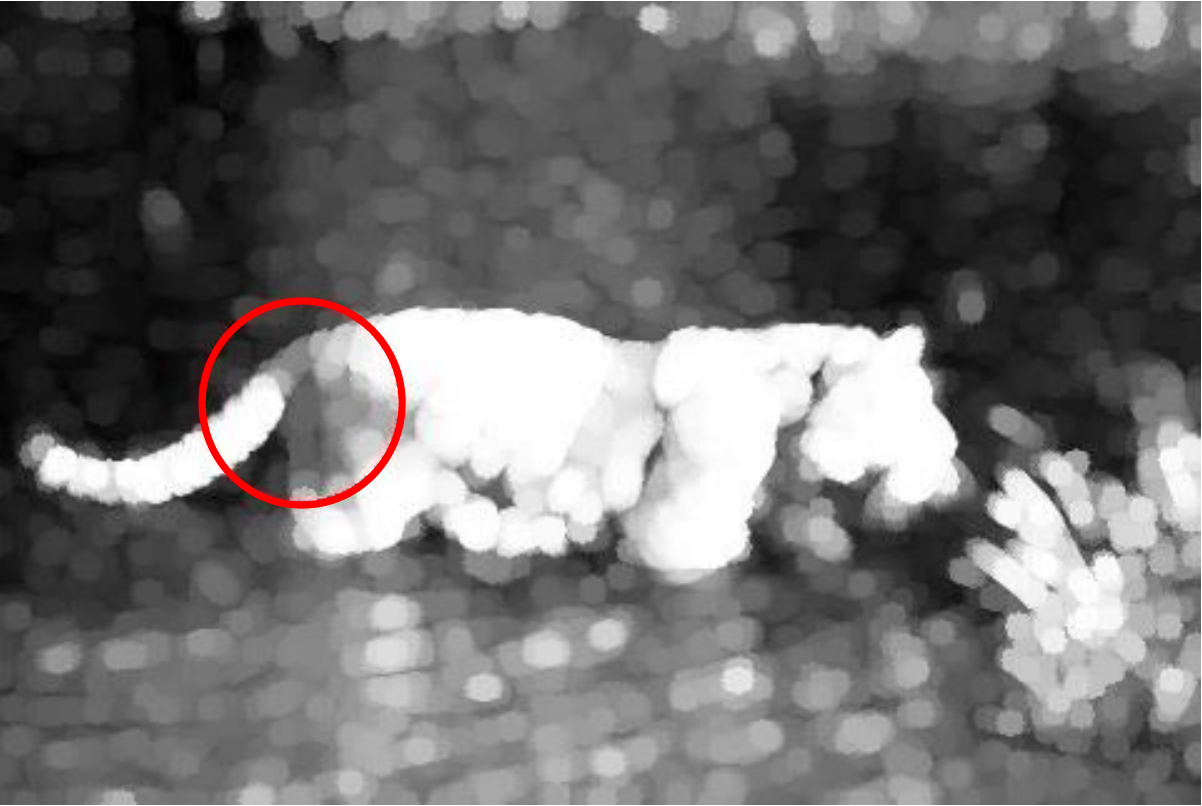}
	\end{subfigure}
	\begin{subfigure}[t]{0.24\textwidth}
		\includegraphics[width=1\textwidth]{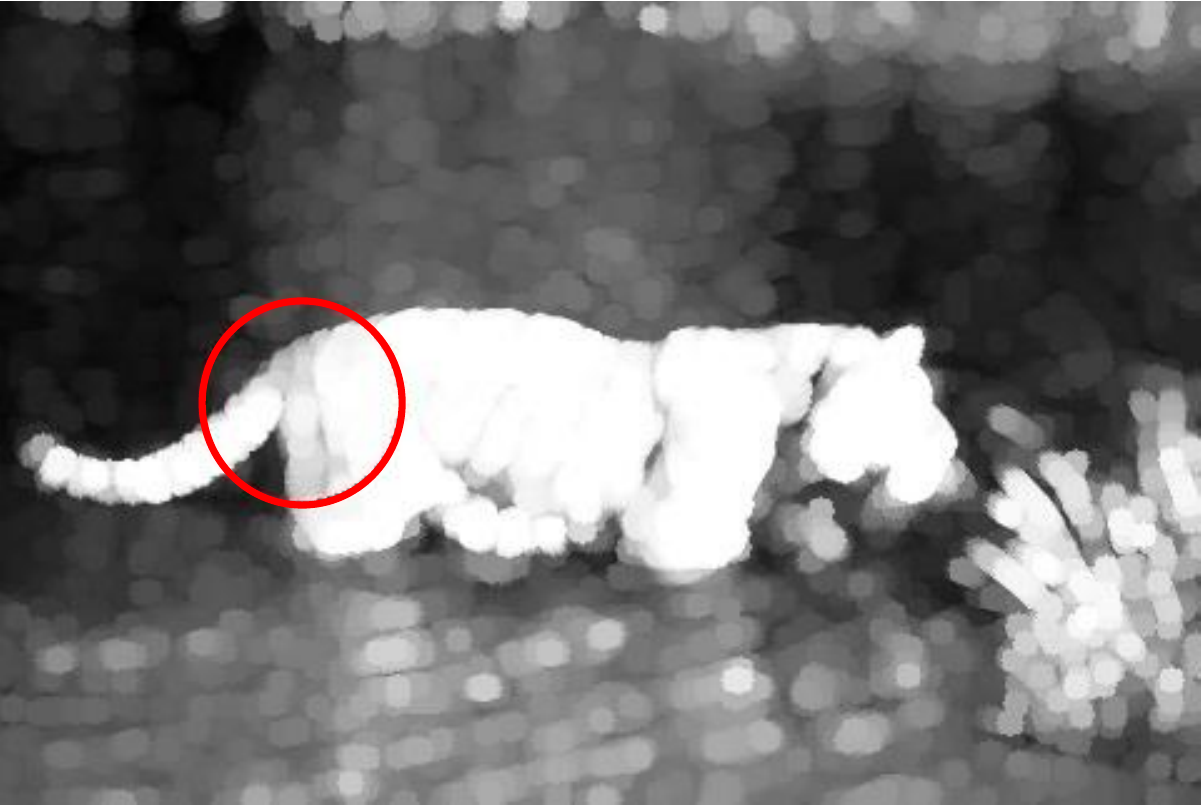}
	\end{subfigure}
	\begin{subfigure}[t]{0.24\textwidth}
		\includegraphics[width=1\textwidth]{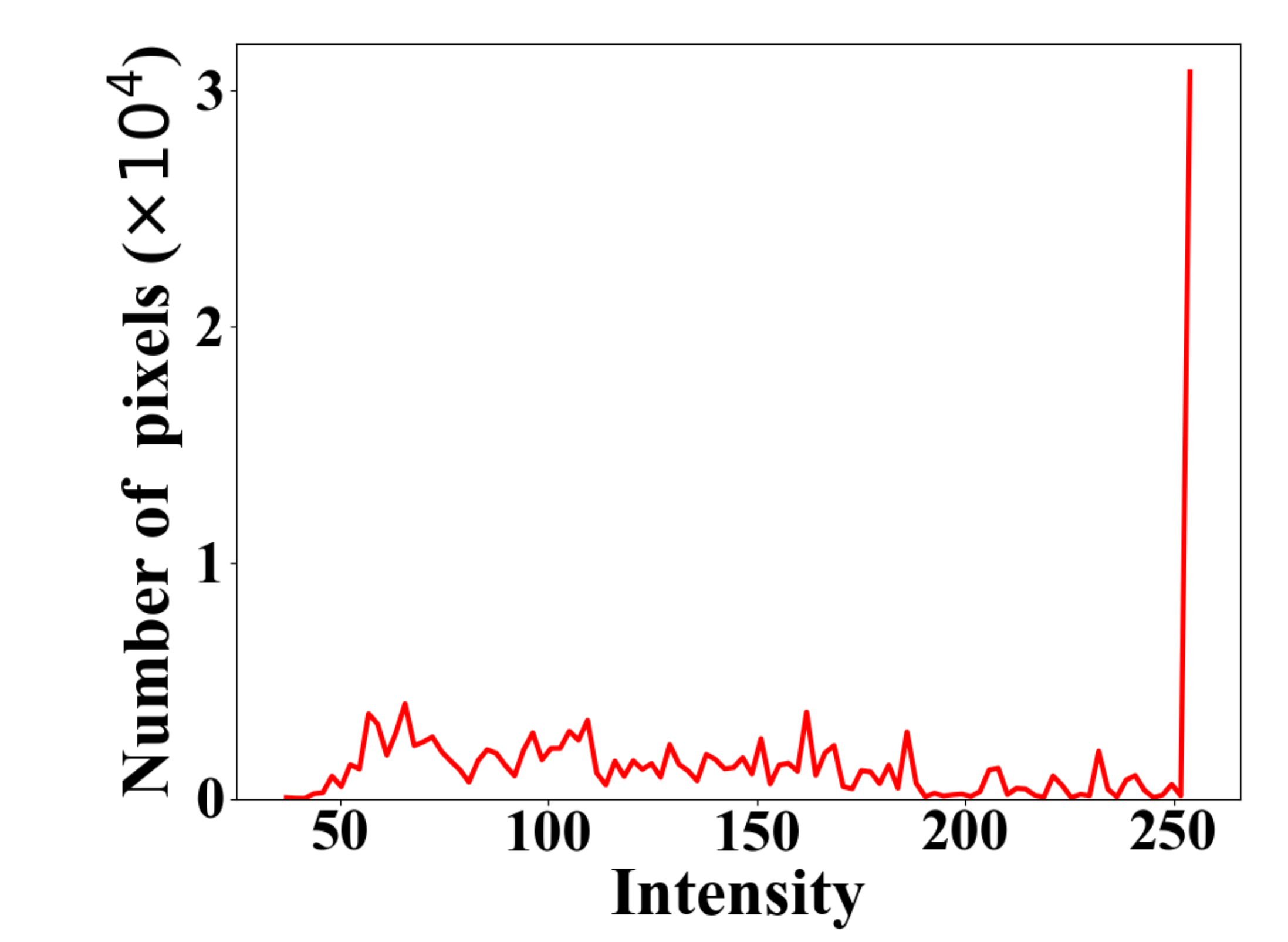}
		\subcaption{\normalsize Ground Truth}
	\end{subfigure}	
	\begin{subfigure}[t]{0.24\textwidth}
		\includegraphics[width=1\textwidth]{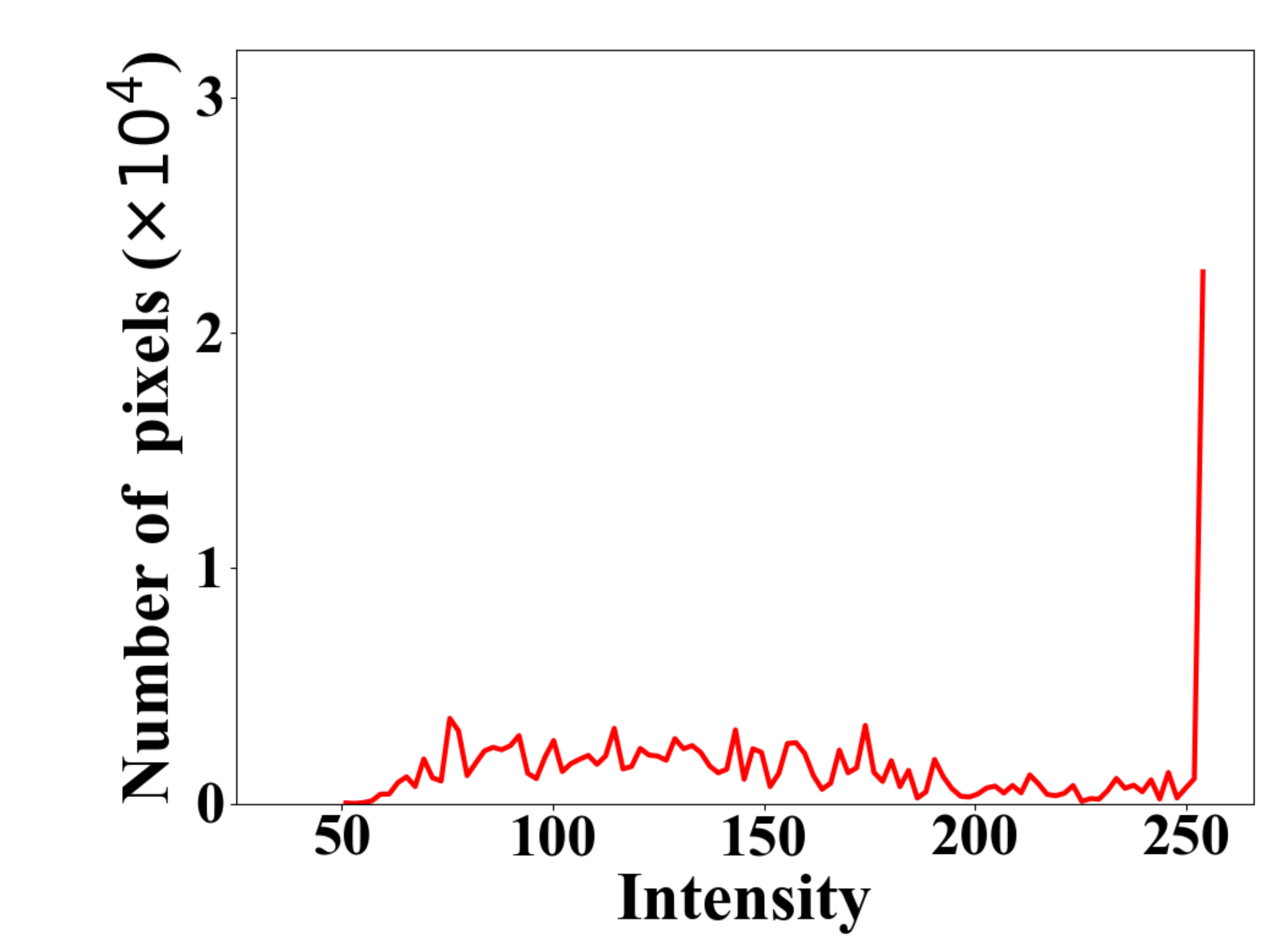}
		\subcaption{\normalsize DDN \cite{fu2017removing}}
	\end{subfigure}	
	\begin{subfigure}[t]{0.24\textwidth}
		\includegraphics[width=1\textwidth]{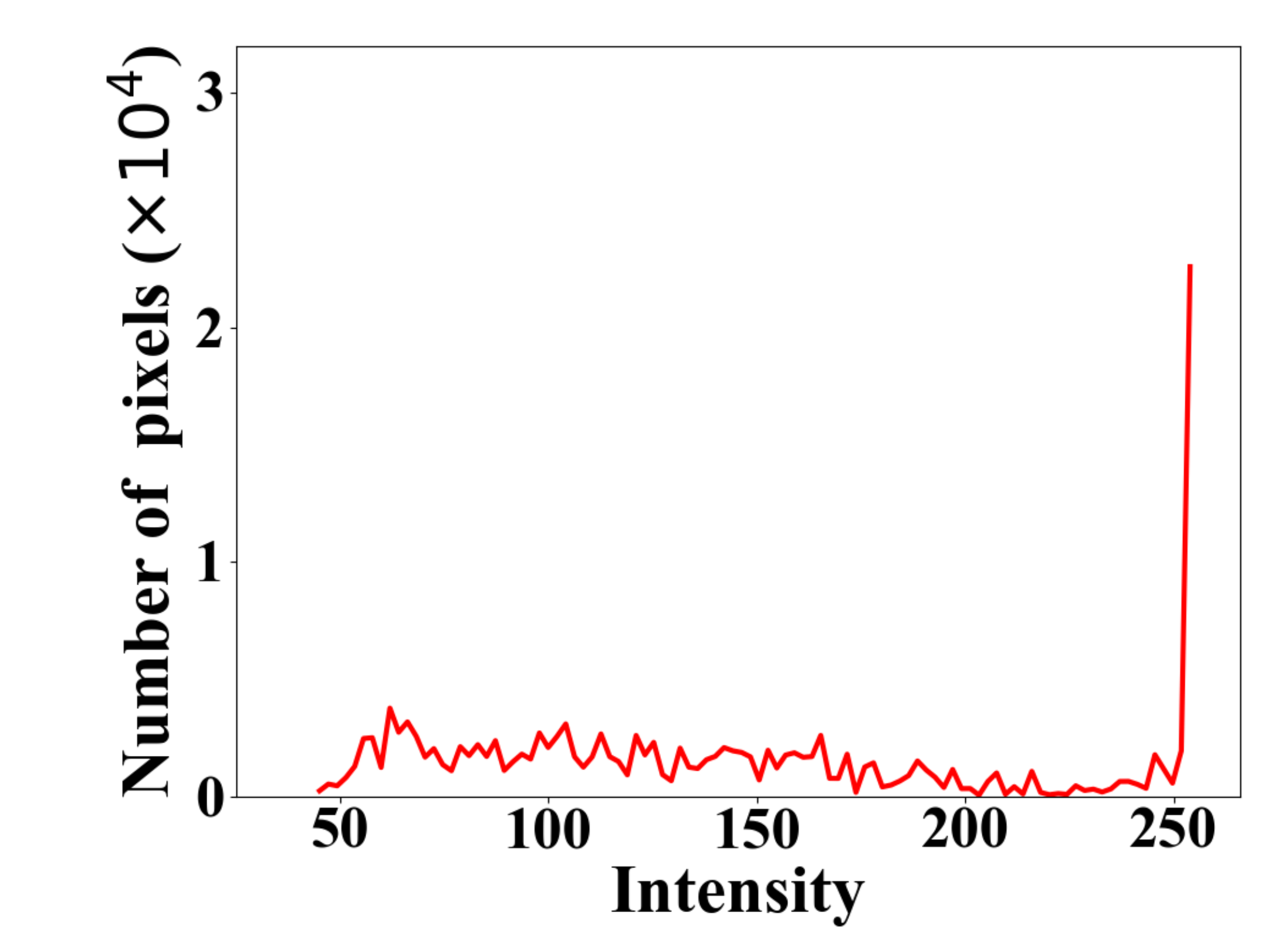}
		\subcaption{\normalsize JORDER \cite{yang2017deeppami}}
	\end{subfigure}	
	\begin{subfigure}[t]{0.24\textwidth}
		\includegraphics[width=1\textwidth]{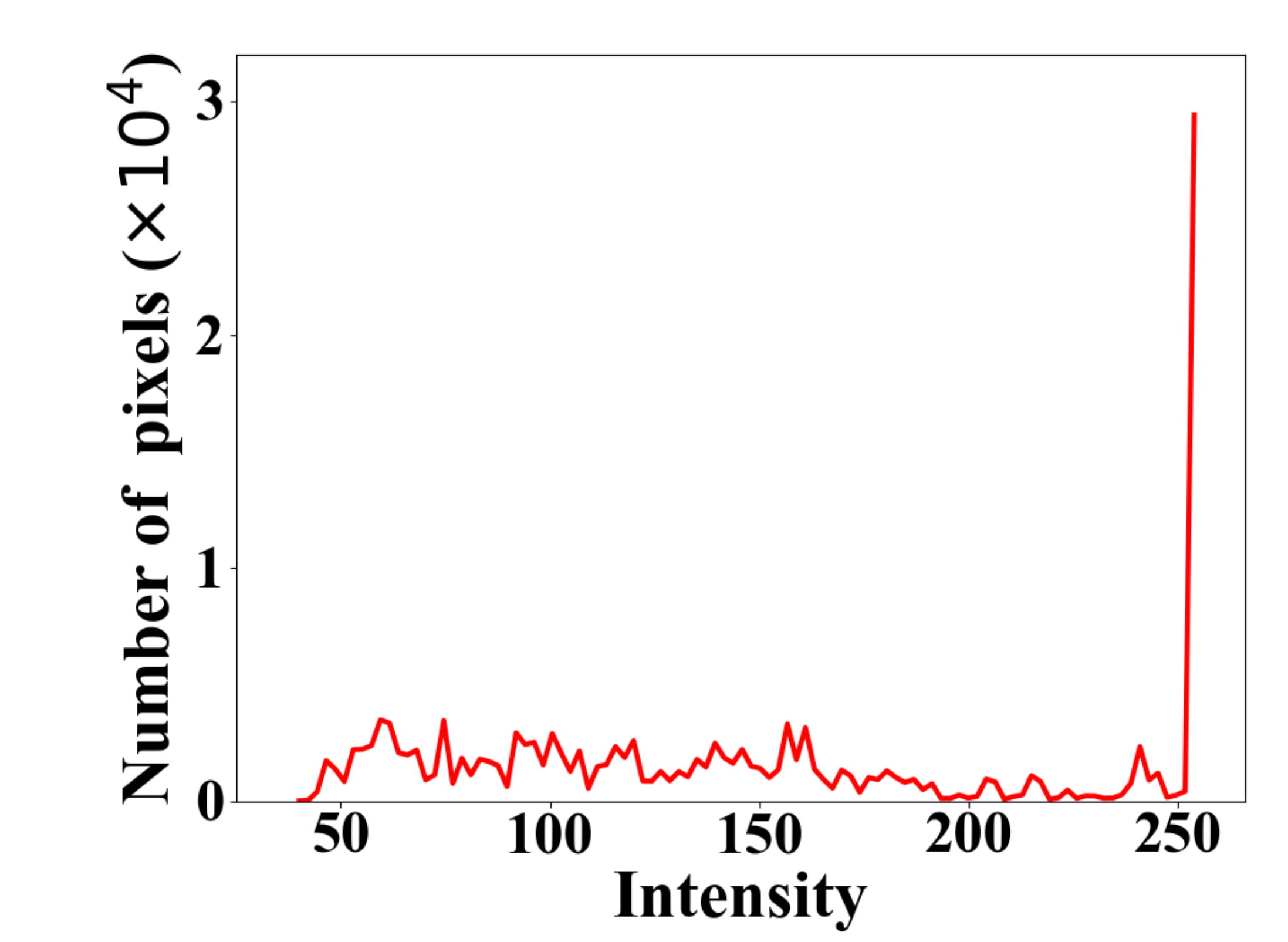}
		\subcaption{\normalsize \bfseries CVID (Ours) }
	\end{subfigure}
	\vspace{-2mm}
	\caption{\textbf{Top}:\ ground truth (a) and derained images (b)-(d) by different methods.\, \textbf{Middle}:\ the corresponding bright channels.\, \textbf{Bottom}:\ the intensity distribution for the bright channel of the derained image.}
	\label{fig:bright channel}
\vspace{-4mm}
\end{figure*}

Aside from the spatially uneven distribution, the rain density is also in distinct distribution for different color channels.\, This point is largely ignored by previous image deraining methods.\, Inspired by the bright channel prior (BCP)~\cite{fu2013variational}, we propose a channel-wise (CW) deraining scheme to further boost the CVAE backbone network for image deraining.\, Until now, our proposed Conditioanl Variational Image Deraining (CVID) network is carried out, by employing the CVAE backbone with the proposed SDE module and CW deraining scheme.\, The BCP prior~\cite{fu2013variational} describes that in natural scenes, for each pixel there at least exists one color channel with high intensity.\, Specifically, the BCP prior is defined as
\begin{equation} 
\label{BCP}
J^{bright}(x) = \underset{y \in \Omega (x)}{\max}(\underset{c \in \{R, G, B\}}{\max} J^{c}(y)),
\end{equation}
where $J^{c}$ is the $c$-th color channel of image $J$ and $\Omega (x)$ is a local patch centered at location $x$.\, The intensity of $J^{bright}$ should be close to $1$ (intensity is in $[0,1]$), except in a situation lacking light or dominated by shadow~\cite{fu2013variational}.\, With the BCP prior (\ref{BCP}), we propose a proposition to validate our CW deraining strategy as follows (the proof is provided in \S\ref{appendix}):
\begin{prop}
Denote $\bar{B}$ and $B$ as images derained without and with distinguishing different color channels, respectively.\, $\|\cdot\|_0$ is the $\ell_{0}$ norm, counting the number of non-zero values.\, Then, the intensity of the pixels in $\bar{B}$ is much lower than $B$.\, That is, the number of brightest pixels in $\bar{B}$ tends to be less than that in $B$.\, That is, we have
\begin{equation}
\label{corollary}
\|1- B\|_0 \leq \|1- \bar{B}\|_0.
\end{equation}
\end{prop}
Note that the intensity of the brightest pixels is 1 and the less than or equal to sign can be satisfied if and only if the distribution of rain streaks on each channel is identical.\, To provide a more intuitive illustration, in Fig.~\ref{fig:bright channel}, we compare the results of our method with previous representative methods, e.g., DDN~\cite{fu2017removing} and JORDER~\cite{yang2017deep,yang2017deeppami}, that do not separate the three color channels.\, From the middle row it can be seen that, the intensities of the brightest pixels of the derained image obtained by DDN and JORDER are mostly lower than that of our CVAE network, as indicated by the red circle.\, Fig. \ref{fig:bright channel} (bottom row) shows the intensity distribution for the bright channels of the derained image.\, One can see that the derained image with our channel-wise scheme contains more brightest pixels than others, demonstrating the effectiveness of the proposed channel-wise scheme.\, The advantages of separately processing each color channel are also validated in other low-level vision tasks~\cite{he2010single,twsc,mcwnnm}.\,
%the middle row shows the bright channels of the corresponding derained images and the ground truth clean image, where we can see that

\subsection{Optimization}
\label{sec:optim}
The proposed CVID network is optimized by jointly minimizing the negative conditional variational lower bound~(\ref{cvae}) and the loss of the SDE module defined in Eqn.\, (\ref{sde_loss}).\, Specifically, we formulate the objective function of our CVID network as an integration of the CVAE loss in Eqn.\, (\ref{cvae_loss}) and the SDE loss in Eqn.\, (\ref{sde_loss}):
\begin{equation}
\label{final_loss}
\setlength{\abovedisplayskip}{3pt}
\mathcal{L}_{\rm CVID} = \mathcal{L}_{\rm CVAE} + \lambda \mathcal{L}_{\rm SDE},
\setlength{\belowdisplayskip}{3pt}
\end{equation}
where $\lambda>0$ is a regularization parameter to balance the importance of $\mathcal{L}_{\rm SDE}$ and $\mathcal{L}_{\rm CVAE}$.\, We observe that CVID constantly achieves satisfactory performance when we treat $\mathcal{L}_{\rm SDE}$ and $\mathcal{L}_{\rm CVAE}$ equally, i.e., $\lambda = 1$.\, In our CVID network, the CVAE loss $\mathcal{L}_{\rm CVAE}$ (\ref{cvae_loss}) and SDE loss $\mathcal{L}_{\rm SDE}$ (\ref{sde_loss}) are jointly minimized by gradient decent via backward error propagation in an end-to-end manner.\,

\subsection{Inference Stage}
\label{sec:Inf}

To obtain a deterministic output during inference, we draw $n$ latent codes $\{\mathbf{z}_j^{p}\}_{j=1}^{n}$ from the prior distribution $p_{\theta}(\mathbf{z}_p|\mathbf{x})$ learned by the prior network, and simply take the average of the $n$ posteriors as the final prediction.\, Specifically, we compute the marginal likelihood of the latent clean image $\mathbf{y}$ using the Monte Carlo method~\cite{2007smcm.book}:
\vspace{-2mm}
\begin{equation}
    p_{\theta}(\mathbf{y}|\mathbf{x}) \approx \frac{1}{n} \sum^n_{j=1}p_{\theta}(\mathbf{y}|\mathbf{x}, \mathbf{z}_j^{p}), \mathbf{z}_j^{p} \sim p_{\theta}(\mathbf{z}_{p}|\mathbf{x}).
      \vspace{-2mm}
\label{infer}
\end{equation}
For the second term of (\ref{cvae}), we use the Monte Carlo sampling to estimate its conditional log-likelihoods (CLL).\, Initial experiments demonstrate that 100 samples are enough to obtain an accurate estimation of the CLL.\, 
%(please refer to (\ref{infer})).\,
In Algorithm~\ref{alg:A}, we summarize the learning and inference procedures of the proposed CVID network for image deraining.\,

\begin{figure*}[t]
    \centering
	\begin{subfigure}[t]{0.23\textwidth}
		\centering
		\includegraphics[width=1\textwidth]{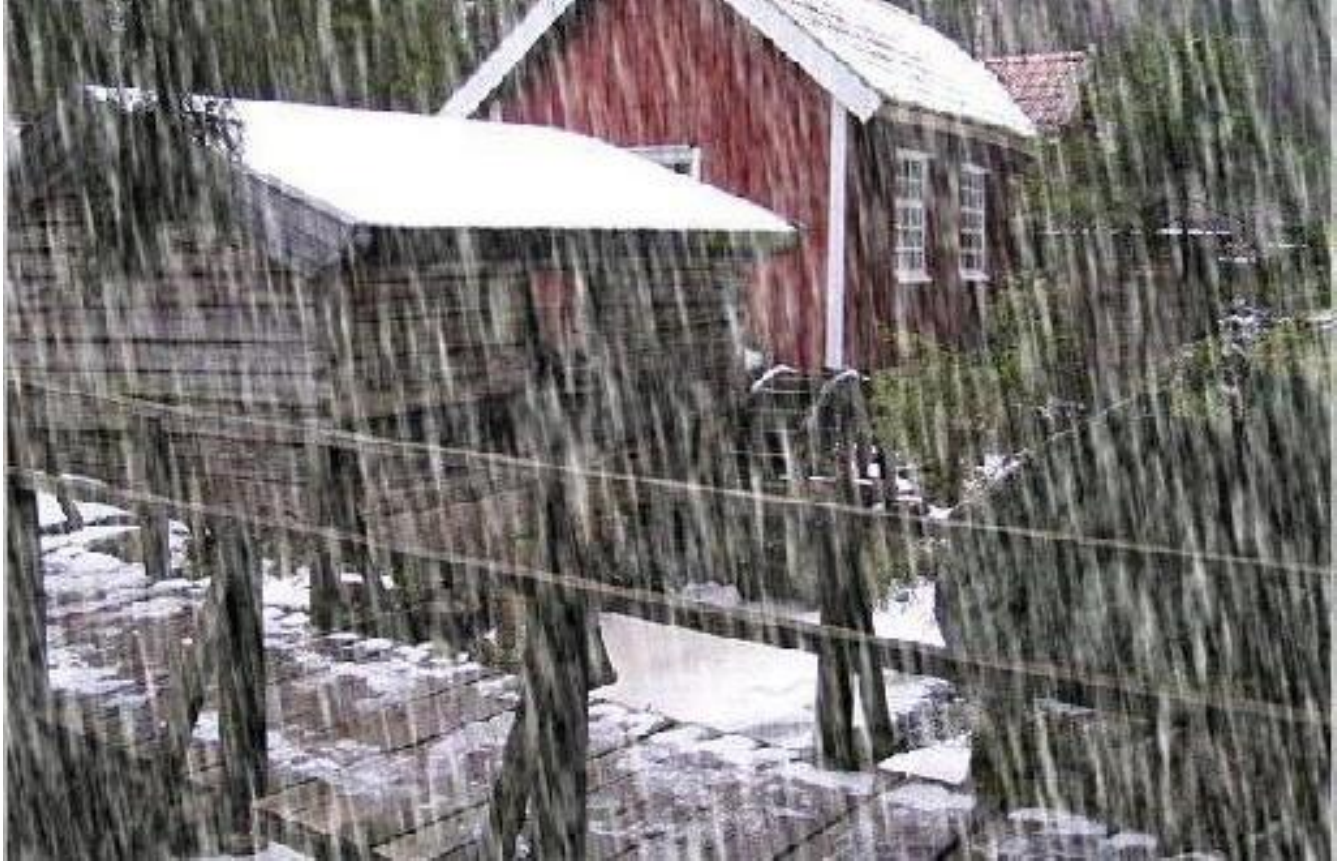}
		 \subcaption*{Rainy Image: 22.97/0.4834}
	\end{subfigure}	
	\begin{subfigure}[t]{0.23\textwidth}
		\centering
		\includegraphics[width=1\textwidth]{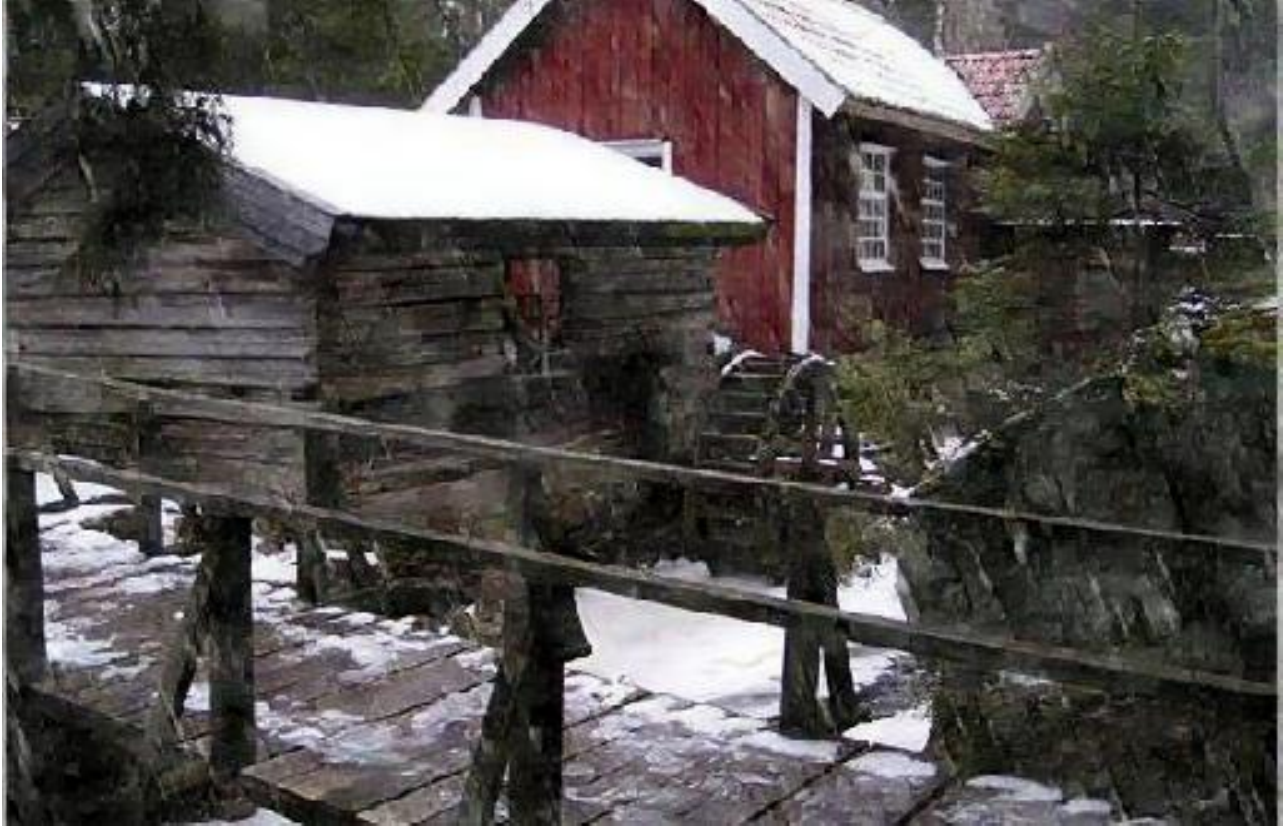}
		     \subcaption*{LP~\cite{li2016rain}: 27.74/0.7357}
	\end{subfigure}	
	\begin{subfigure}[t]{0.23\textwidth}
		\centering
		\includegraphics[width=1\textwidth]{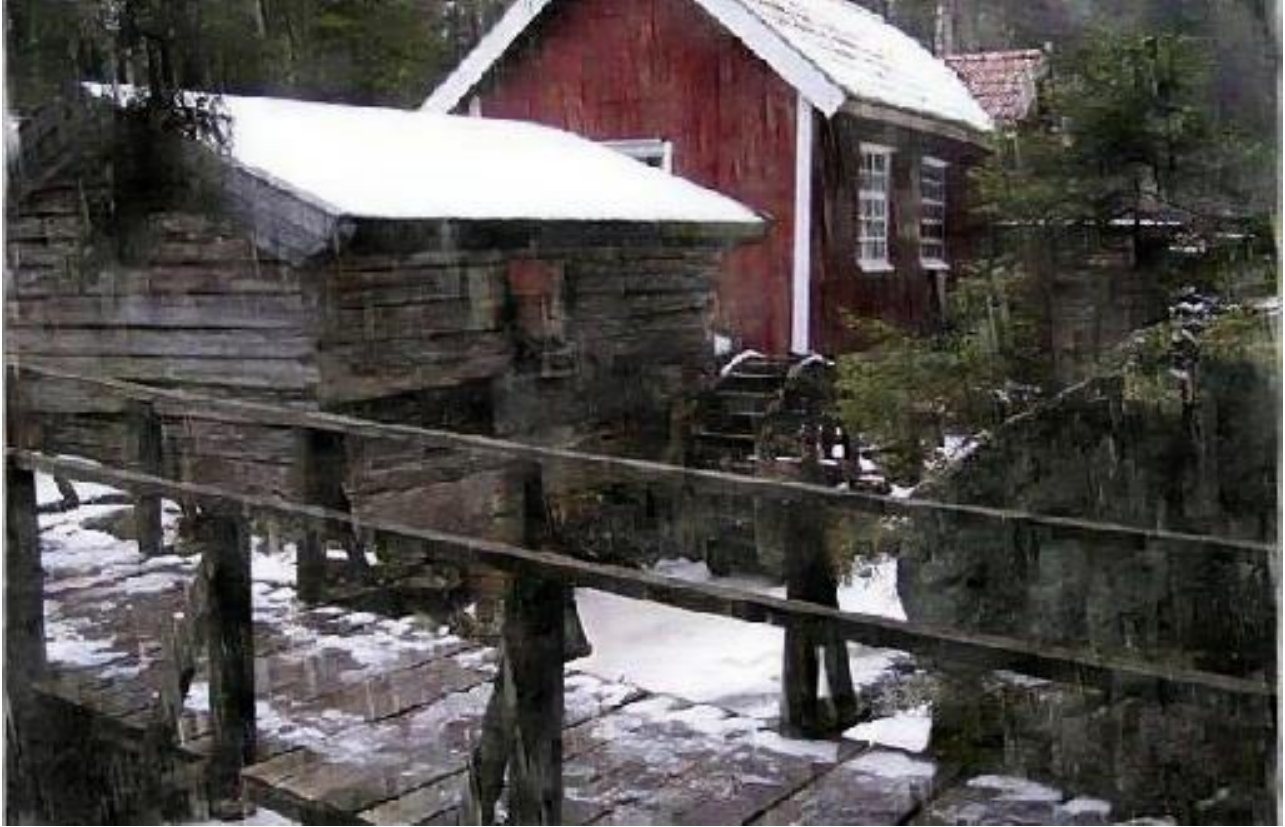}
		\subcaption*{DDN~\cite{fu2017removing}: 27.89/0.7623}
	\end{subfigure}	
	\begin{subfigure}[t]{0.23\textwidth}
		\centering
		\includegraphics[width=1\textwidth]{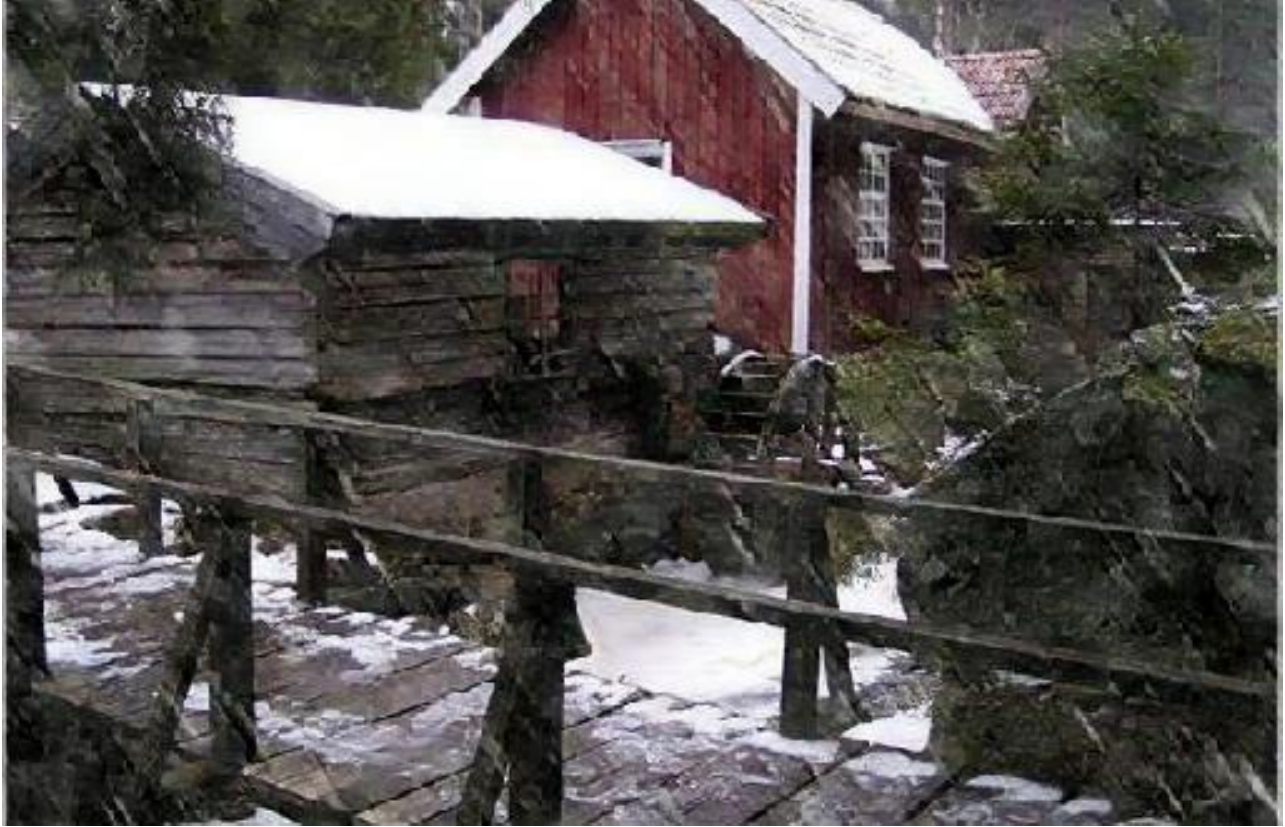}
		\subcaption*{JORDER~\cite{yang2017deeppami}: 28.32/0.7509}
	\end{subfigure}	
	\begin{subfigure}[t]{0.23\textwidth}
		\centering
		\includegraphics[width=1\textwidth]{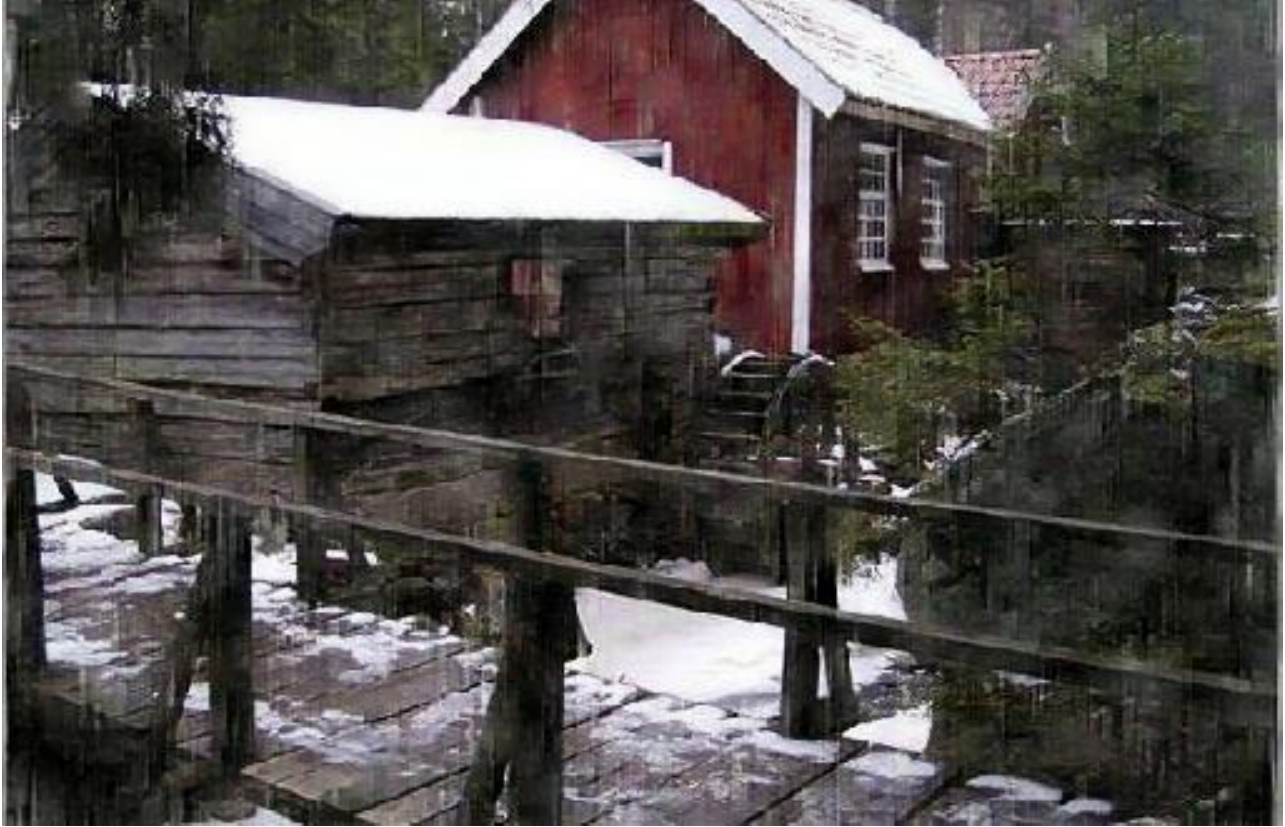}
		\subcaption*{DID-MDN~\cite{zhang2018density}: 29.98/0.8593}
	\end{subfigure}	
    \begin{subfigure}[t]{0.23\textwidth}
		\centering
		\includegraphics[width=1\textwidth]{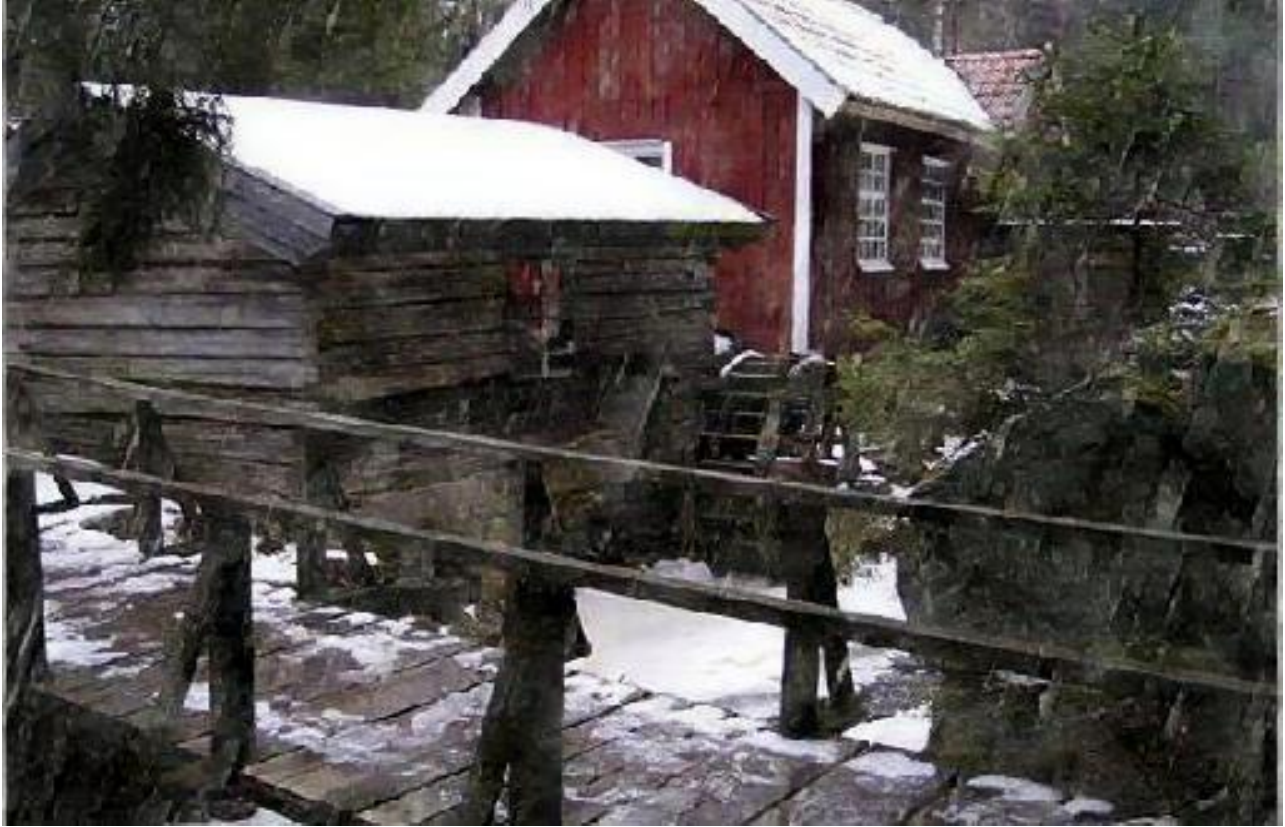}
		\subcaption*{RESCAN~\cite{li2018recurrent}: 29.82/0.8460}
	\end{subfigure}	
	\begin{subfigure}[t]{0.23\textwidth}
		\centering
		\includegraphics[width=1\textwidth]{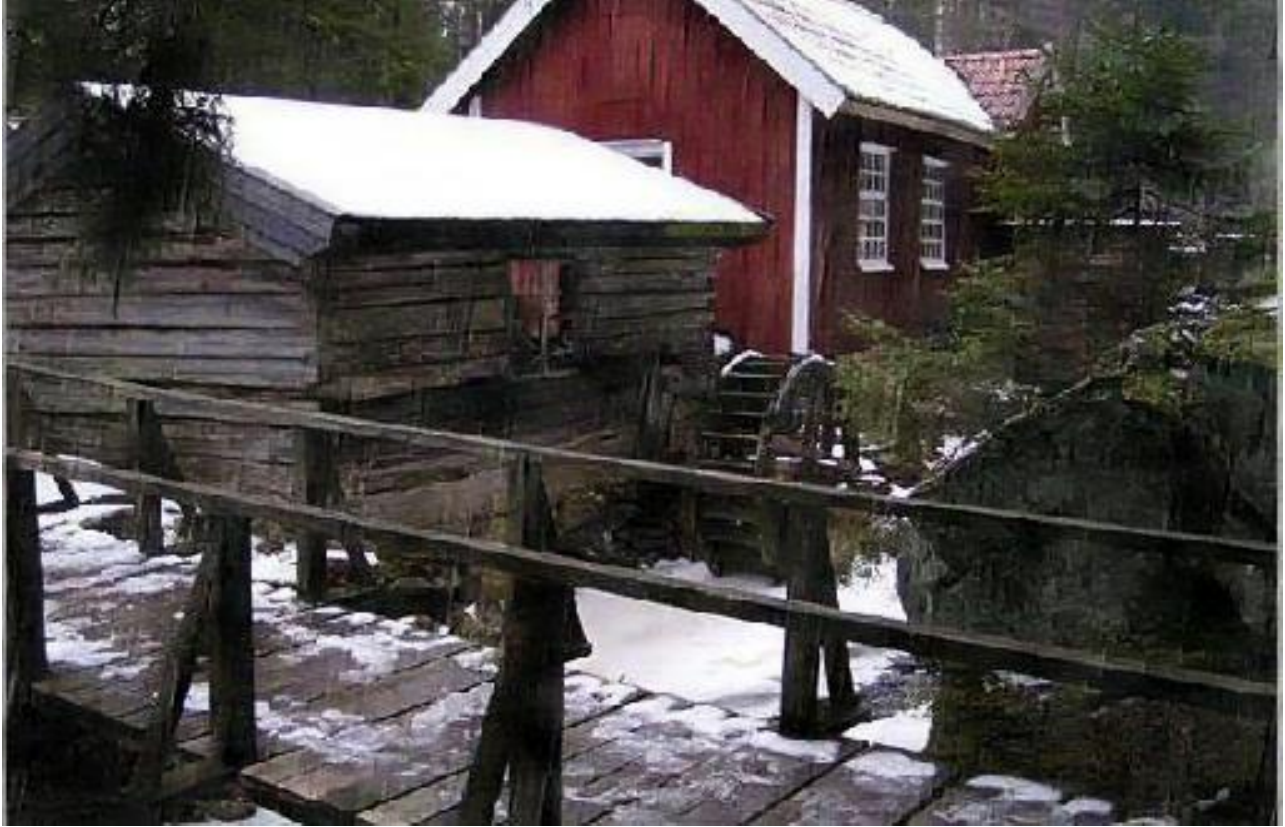}
		\subcaption*{\textbf{CVID} (\textbf{Ours}): \textbf{31.75}/\textbf{0.8820}}
	\end{subfigure}	
	\begin{subfigure}[t]{0.23\textwidth}
		\centering
		\includegraphics[width=1\textwidth]{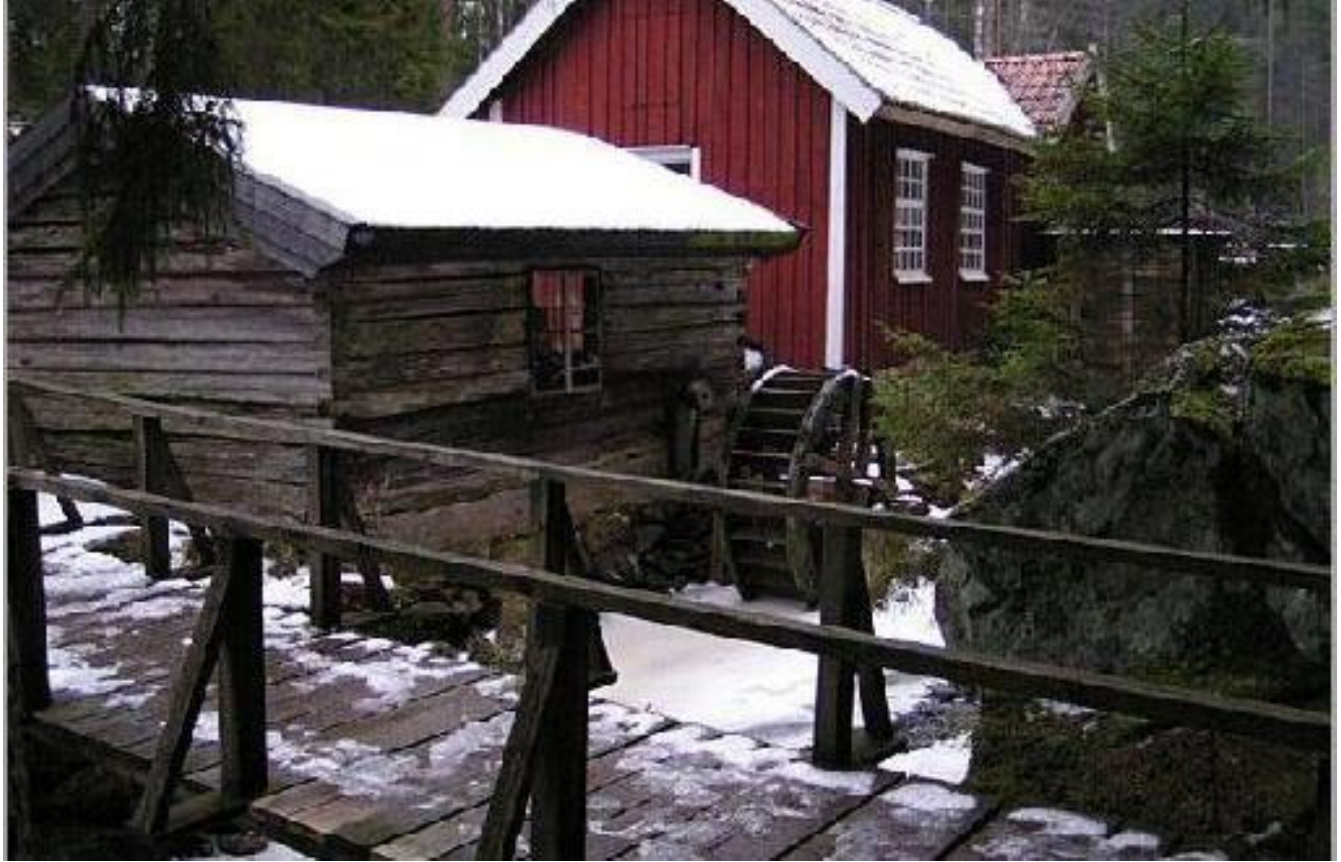}
		\subcaption*{Ground Truth}
	\end{subfigure}
	\vspace{-2mm}
	\caption{\textbf{Derained images and PSNR (dB)/SSIM by different methods} on a synthetic rainy image from \textit{D1}~\cite{fu2017removing}.}
	\vspace{-2mm}
    \label{fig:syne1}
\end{figure*}

\begin{figure*}[t]
    \centering
	\begin{subfigure}[t]{0.23\textwidth}
		\centering
		\includegraphics[width=1\textwidth]{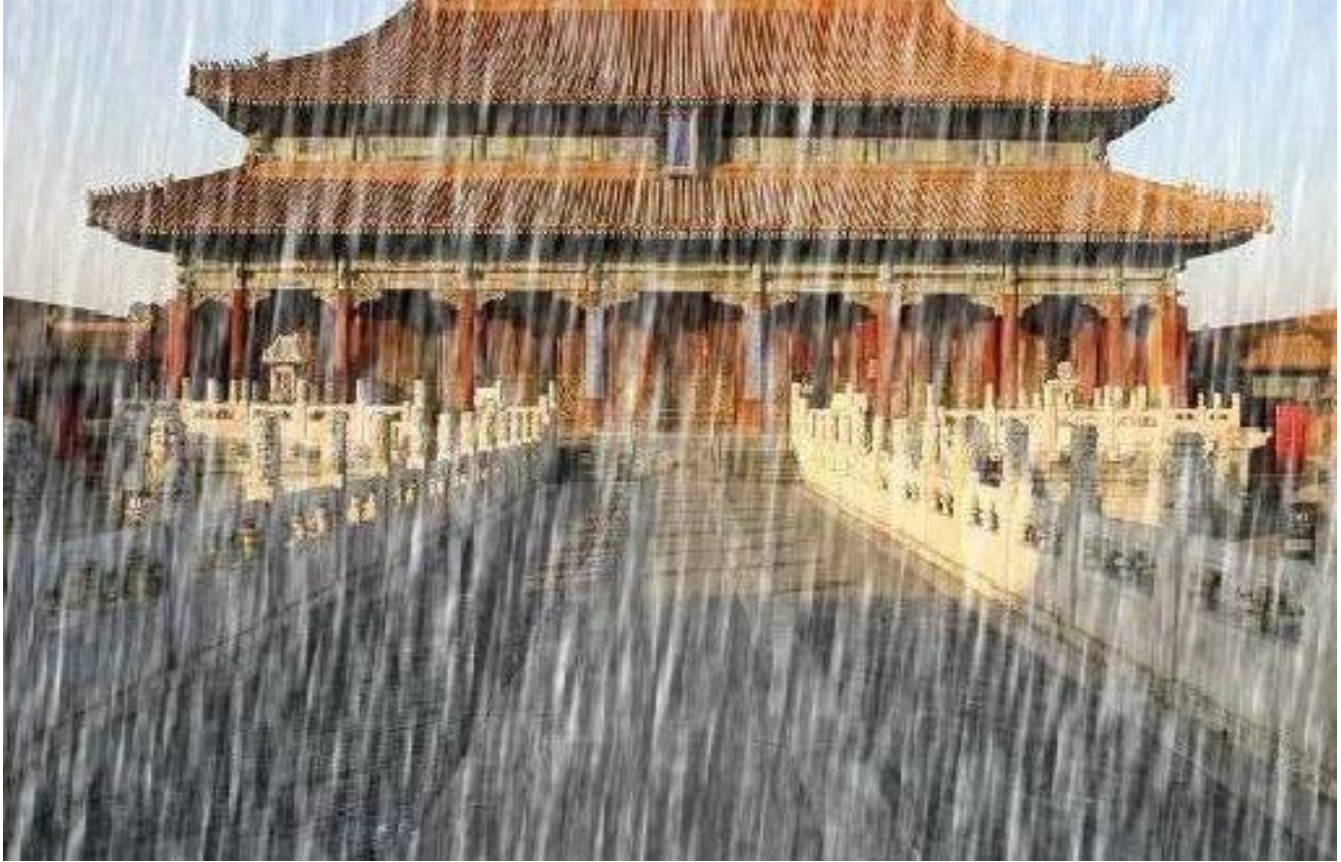}
		 \subcaption*{Rainy Image: 23.17/0.4587}
	\end{subfigure}	
	\begin{subfigure}[t]{0.23\textwidth}
		\centering
		\includegraphics[width=1\textwidth]{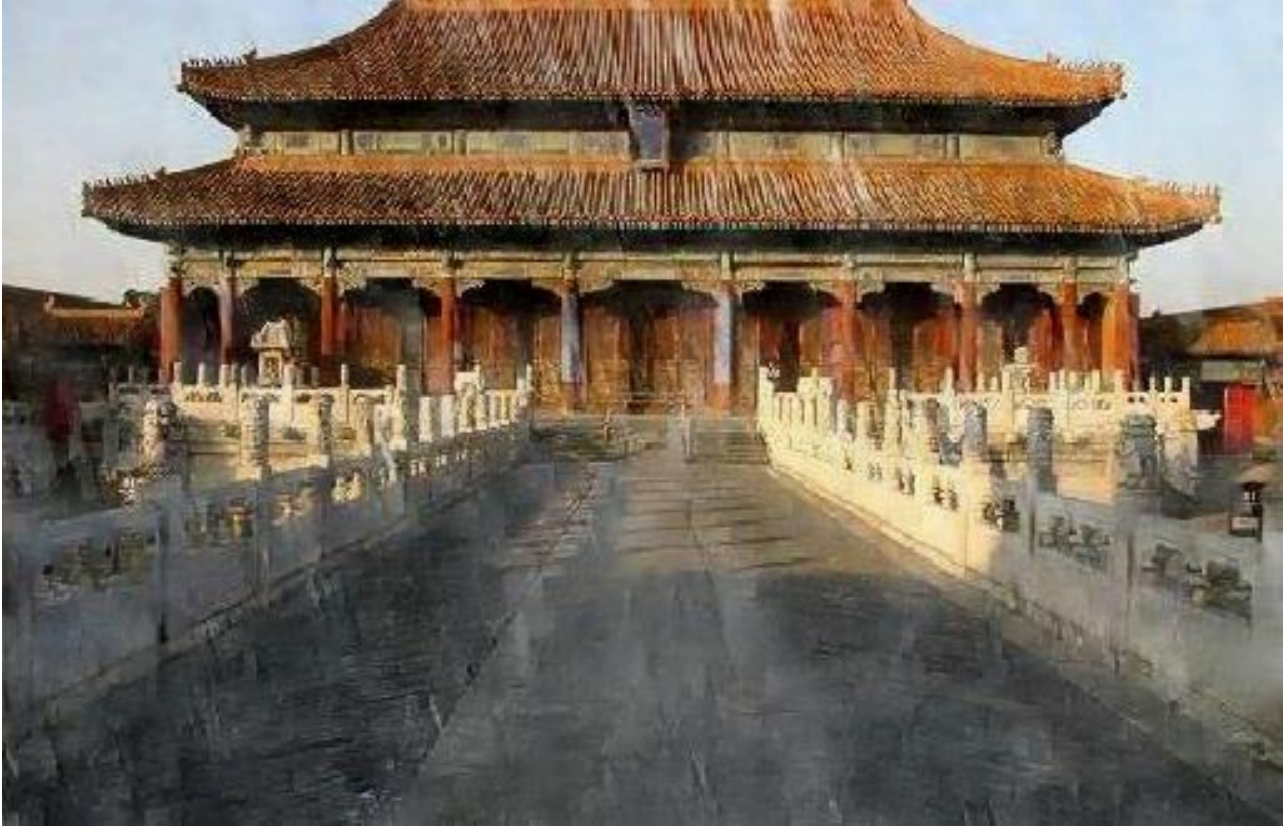}
		     \subcaption*{LP~\cite{li2016rain}: 25.37/0.7318}
	\end{subfigure}	
	\begin{subfigure}[t]{0.23\textwidth}
		\centering
		\includegraphics[width=1\textwidth]{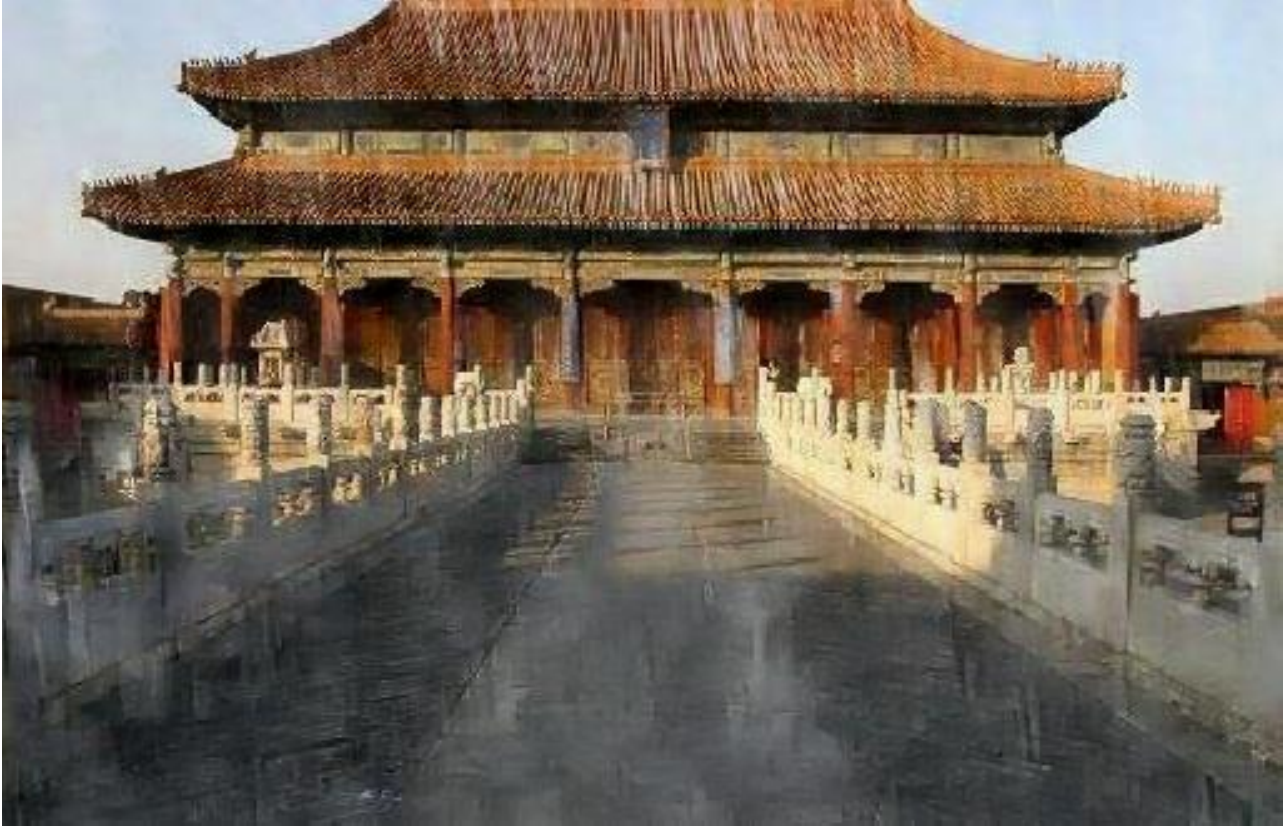}
		\subcaption*{DDN~\cite{fu2017removing}: 26.73/0.7538}
	\end{subfigure}	
	\begin{subfigure}[t]{0.23\textwidth}
		\centering
		\includegraphics[width=1\textwidth]{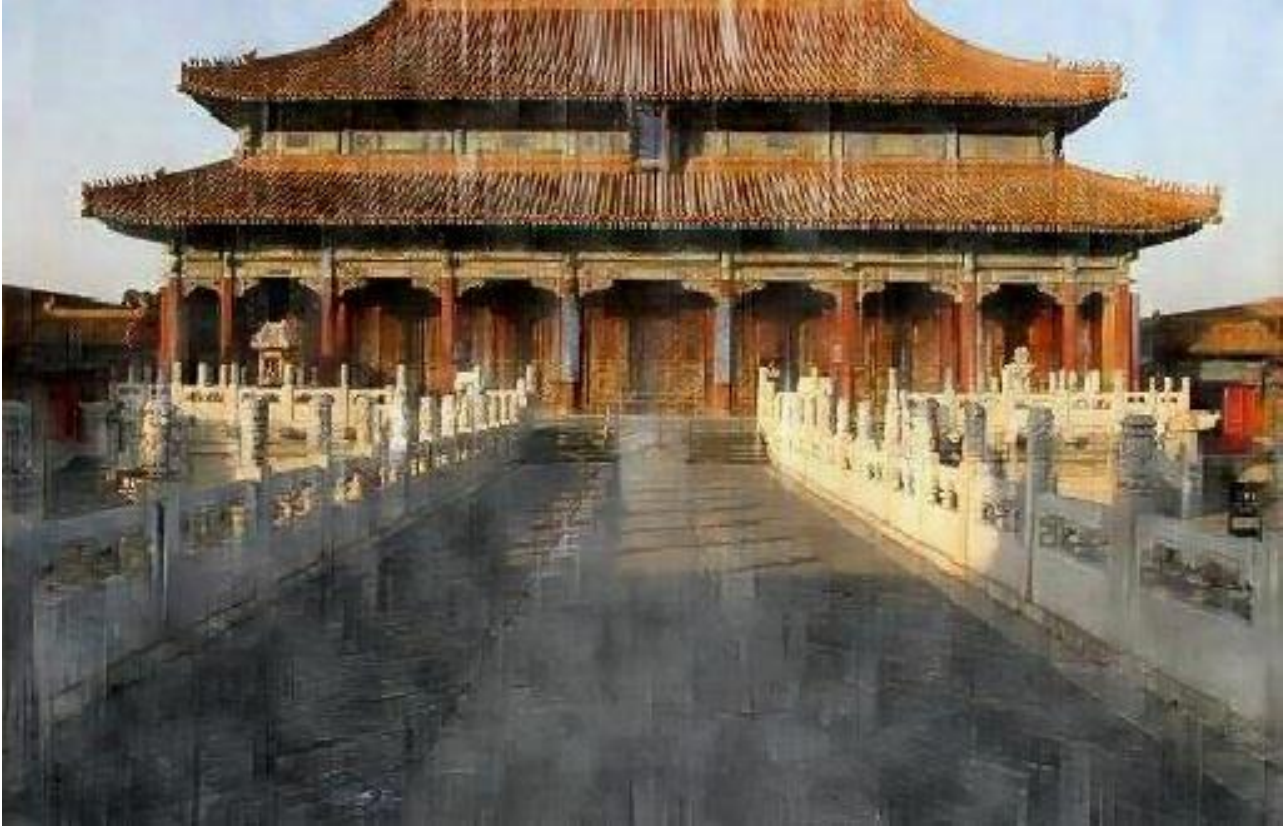}
		\subcaption*{JORDER~\cite{yang2017deeppami}: 26.65/0.7493}
	\end{subfigure}	
	\begin{subfigure}[t]{0.23\textwidth}
		\centering
		\includegraphics[width=1\textwidth]{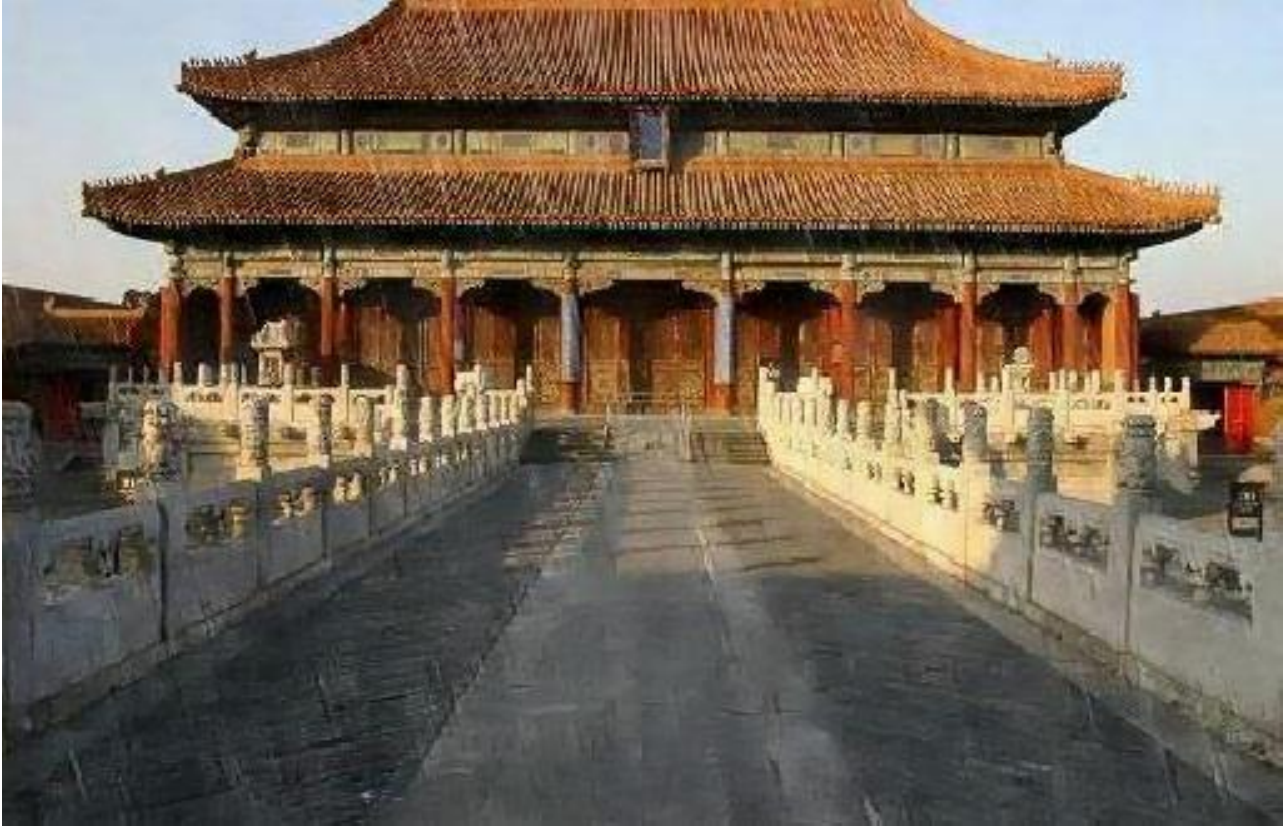}
		\subcaption*{DID-MDN~\cite{zhang2018density}: 29.72/0.8625}
	\end{subfigure}	
    \begin{subfigure}[t]{0.23\textwidth}
		\centering
		\includegraphics[width=1\textwidth]{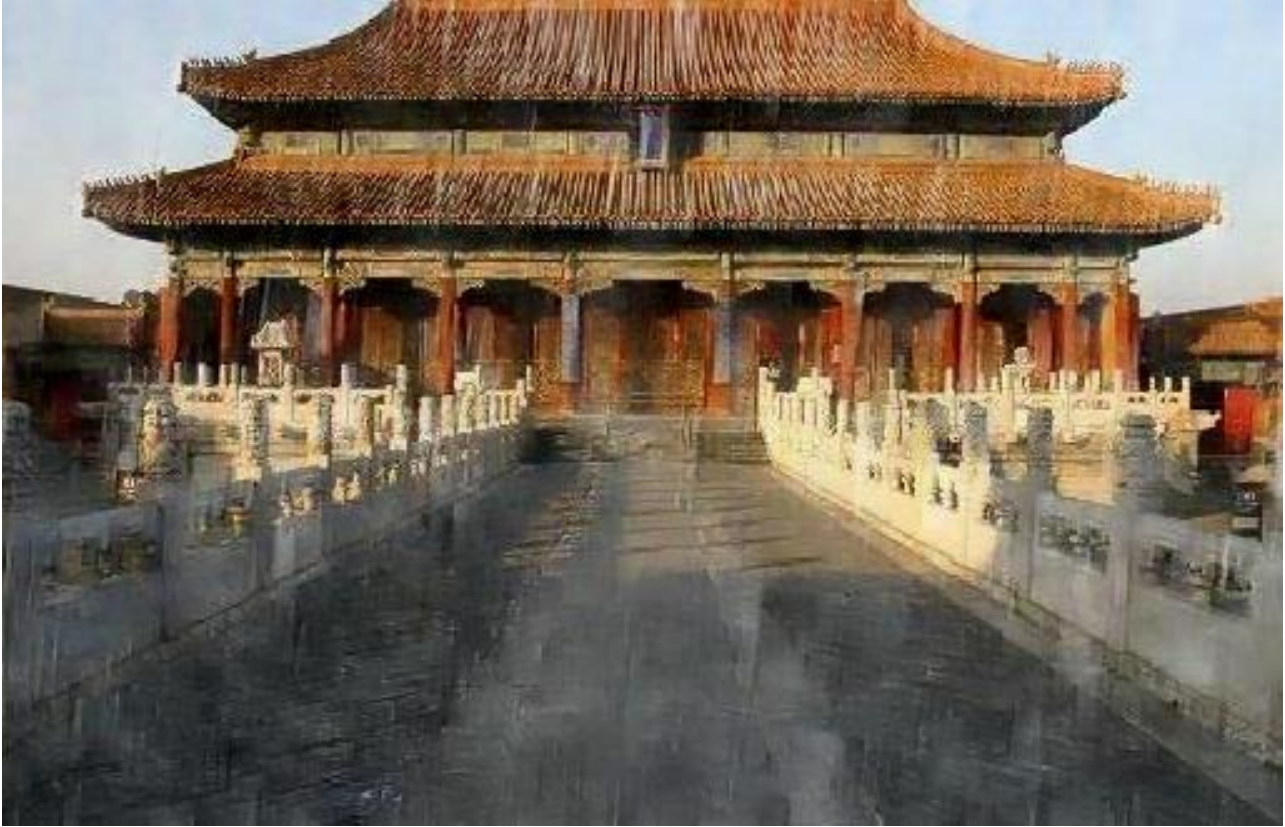}
		\subcaption*{RESCAN~\cite{li2018recurrent}: 28.37/0.8521}
	\end{subfigure}	
	\begin{subfigure}[t]{0.23\textwidth}
		\centering
		\includegraphics[width=1\textwidth]{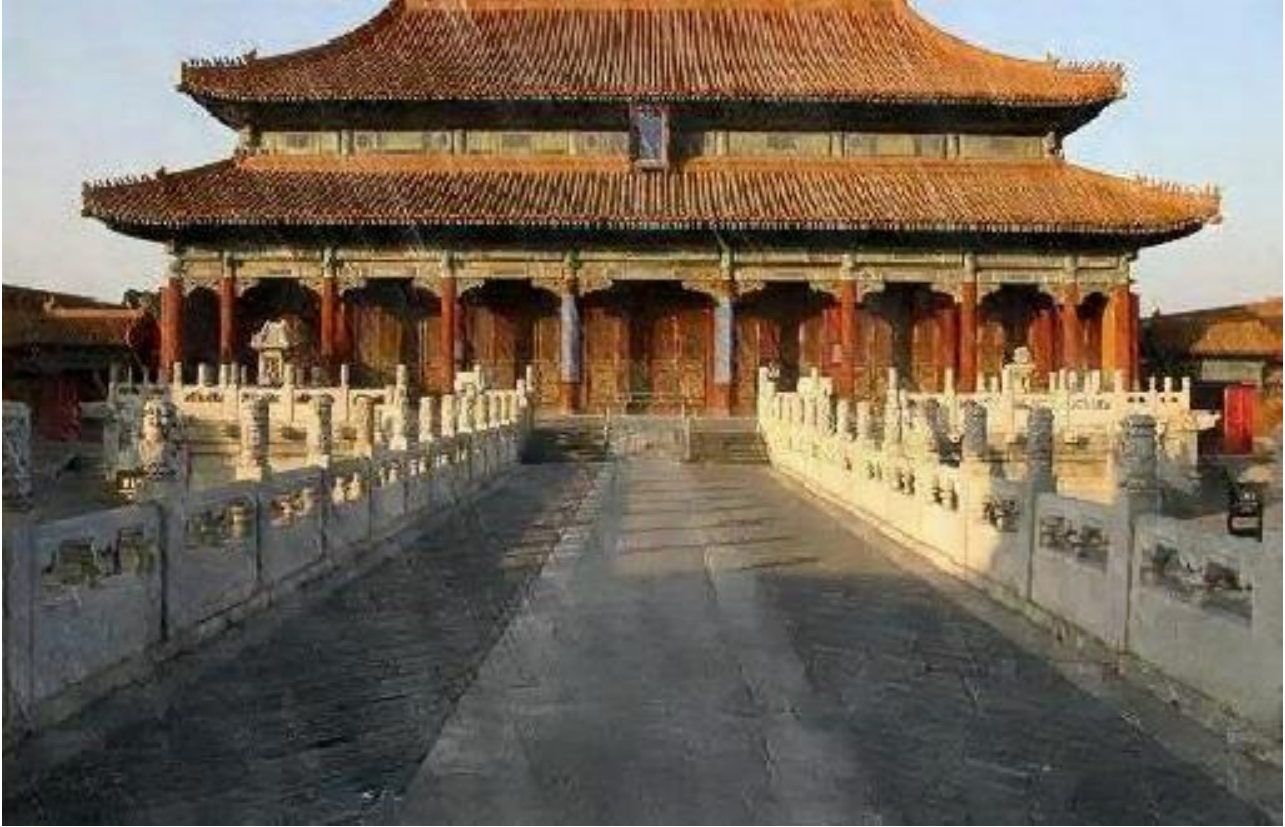}
		\subcaption*{\textbf{CVID} (\textbf{Ours}): \textbf{30.27}/\textbf{0.8921}}
	\end{subfigure}	
	\begin{subfigure}[t]{0.23\textwidth}
		\centering
		\includegraphics[width=1\textwidth]{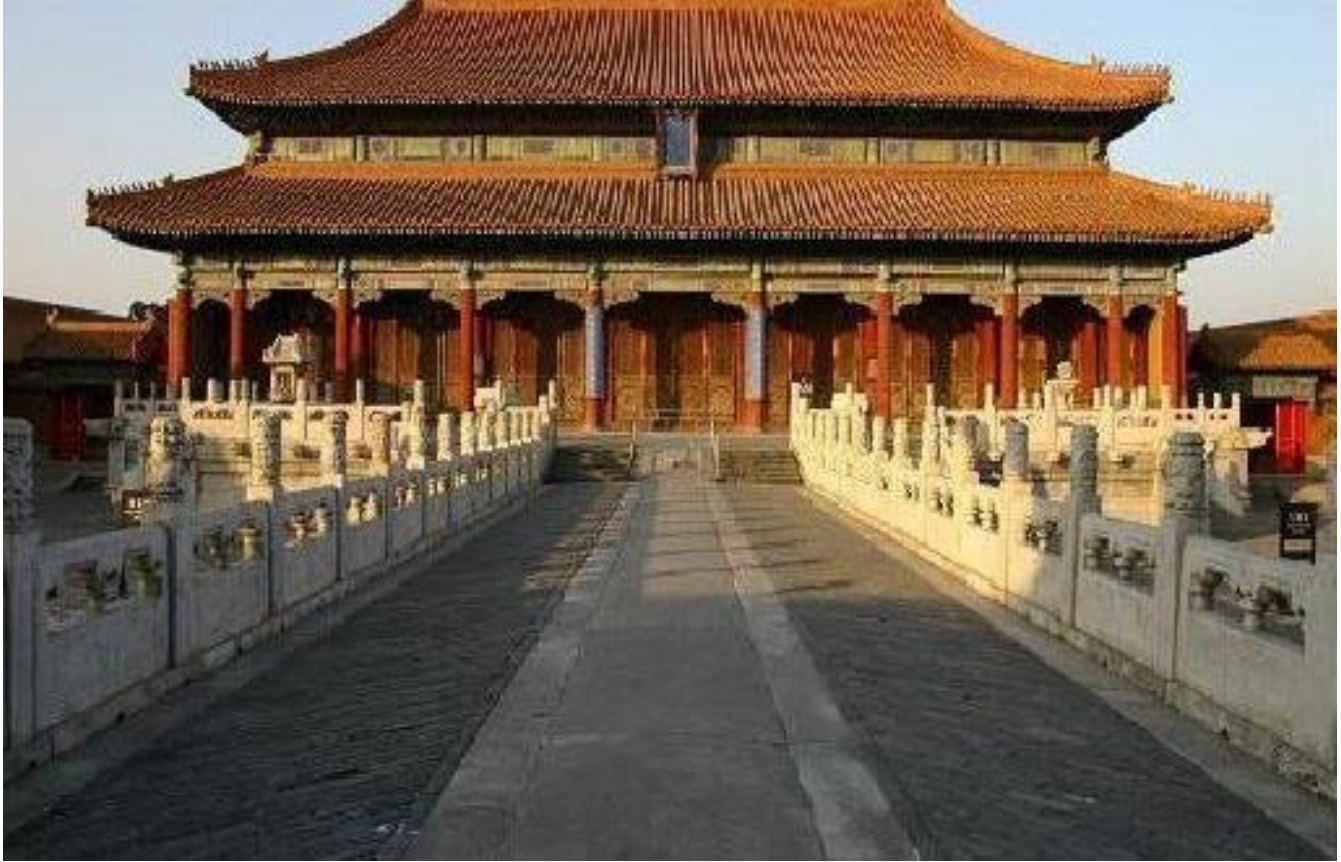}
		\subcaption*{Ground Truth}
	\end{subfigure}
	\vspace{-2mm}
	\caption{\textbf{Derained images and PSNR (dB)/SSIM by different methods} on a synthetic rainy image from \textit{D2}~\cite{yang2017deeppami}}
	\vspace{-1mm}
    \label{fig:syne2}
\end{figure*}

\begin{table*}[htbp]
\begin{center}
%\normalsize
\footnotesize
\begin{tabular}{r||c|c|c|c|c|c|c|c|c|c|c|c}
\Xhline{1pt}
\rowcolor[rgb]{ .85,  .9,  0.95}
%\rowcolor[rgb]{ .9,  .9,  .9}
\multicolumn{1}{c||}{Dataset} & \multicolumn{3}{c|}{\textsl{D1}~\cite{fu2017removing}} &\multicolumn{3}{c|}{\textsl{D2} \textsl{Rain100L}~\cite{yang2017deeppami}} &\multicolumn{3}{c|}{\textsl{D2} \textsl{Rain100H}~\cite{yang2017deeppami}} &\multicolumn{3}{c}{\textsl{D3}~\cite{zhang2018density}}
\\
\hline
\rowcolor[rgb]{ .85,  .9,  0.95}
\multicolumn{1}{c||}{Metric} &PSNR$\uparrow$ &SSIM$\uparrow$ & NIQE$\downarrow$ &PSNR$\uparrow$ &SSIM$\uparrow$ & NIQE$\downarrow$ &PSNR$\uparrow$ &SSIM$\uparrow$ & NIQE$\downarrow$ &PSNR$\uparrow$ &SSIM$\uparrow$ & NIQE$\downarrow$ \\ 
\hline 
Rainy Input &19.31 &0.7695 & 9.17 &23.52 &0.8332 & 7.93 &12.13 &0.3702 & 14.27 &21.15 &0.7781 & 10.15 \\
\rowcolor[rgb]{ .9, .9, .9}
GMM~\cite{li2016rain} &24.35 &0.8312 & 4.83 &32.02 &0.9137 & 3.14 &14.26 &0.5444 & 7.76 &25.23 &0.8514 & 5.37 \\
JORDER~\cite{yang2017deeppami} &22.36 &0.8405 & 4.97 &36.02 &0.9712 & 2.74 &23.45 &0.7382 & 5.39 &24.32 &0.8622 & 5.64 \\
\rowcolor[rgb]{ .9, .9, .9}
DDN\ \ ~\cite{fu2017removing} &25.63 &0.8851 & 3.92 &33.75 &0.9213 & 3.09 &22.26 &0.6928 & 6.89 &27.33 &0.8978 & 4.98 \\
ID-GAN \cite{zhang2017image} &26.31 &0.8932 & 3.43 &35.89 &0.9631 & 2.96 &23.15 &0.7120 & 6.33 &28.18 &0.9102 & 4.01 \\
\rowcolor[rgb]{ .9, .9, .9}
DID-MDN~\cite{zhang2018density} &26.07 &0.9092 & 3.47 &35.73 &0.9602 & 3.02 &23.25 &0.7315 & 6.18 &27.95 &0.9087 & 3.71 \\
RESCAN~\cite{li2018recurrent}  &25.45 &0.8812 & 3.90 &37.27 &0.9813 & 2.39 &26.45 &0.8458 & 4.96 &26.19 &0.8712 & 4.01 \\ 
\hline
\rowcolor[rgb]{ .95, .95, .95}
\textbf{CVID (Ours)}  &\textbf{28.96} &\textbf{0.9375} & \textbf{2.99} &\textbf{37.83} &\textbf{0.9882} & \textbf{2.13} &\textbf{27.93} &\textbf{0.8765} & \textbf{4.71} &\textbf{30.97} &\textbf{0.9374} & \textbf{3.33}
\\
\hline
\end{tabular}
\end{center}
\vspace{-3mm}
\caption{\textbf{Comparison of PSNR (dB), SSIM~\cite{wang2004image}, and NIQE~\cite{wang2004image} results} by different methods on three synthetic datasets.}
\vspace{-4mm}
\label{ssim_1}
\end{table*}

\begin{figure*}[t]
 \begin{center}
     \begin{subfigure}[t]{0.16\textwidth}
		\includegraphics[width=1\textwidth]{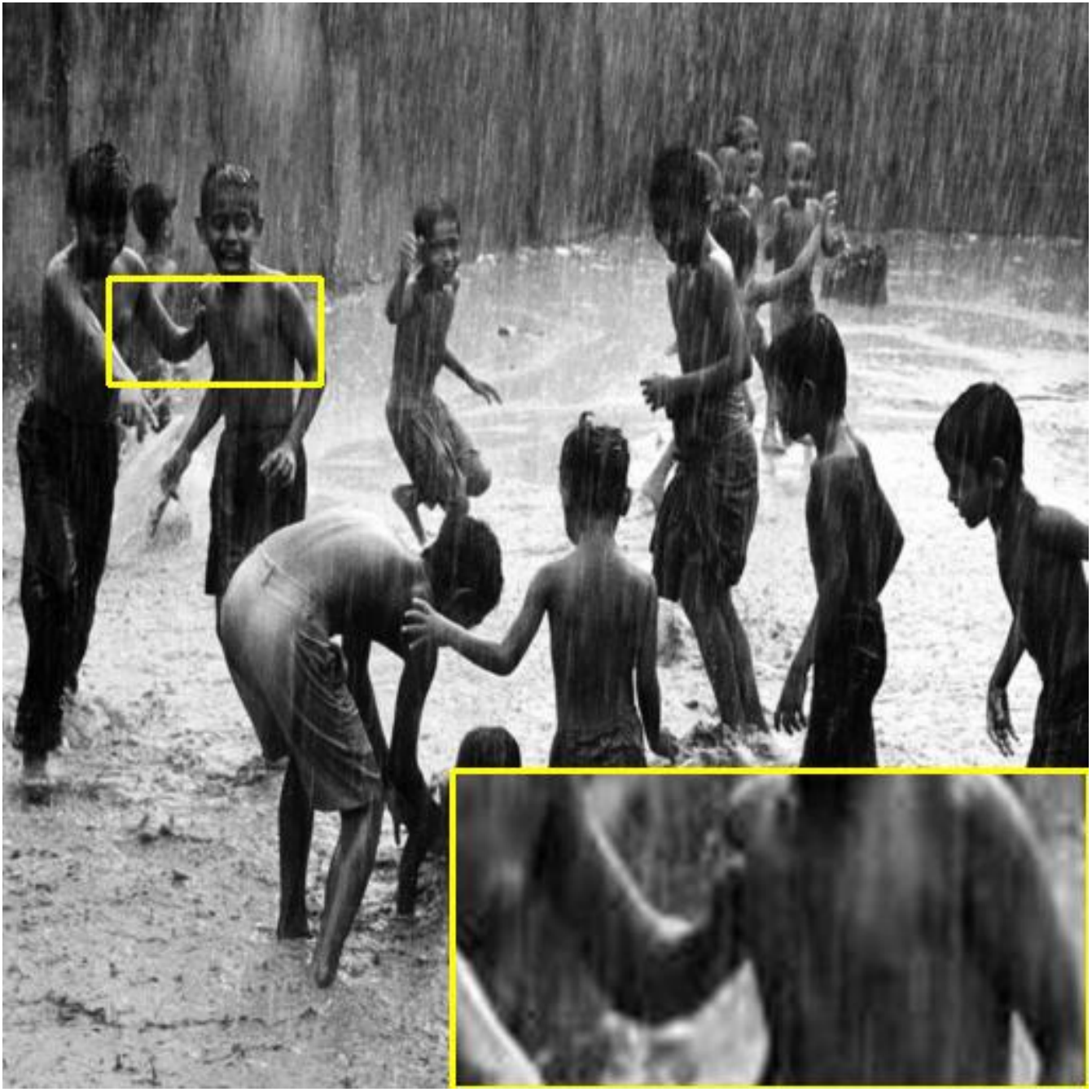}
	\end{subfigure}	
	\begin{subfigure}[t]{0.16\textwidth}
		\includegraphics[width=1\textwidth]{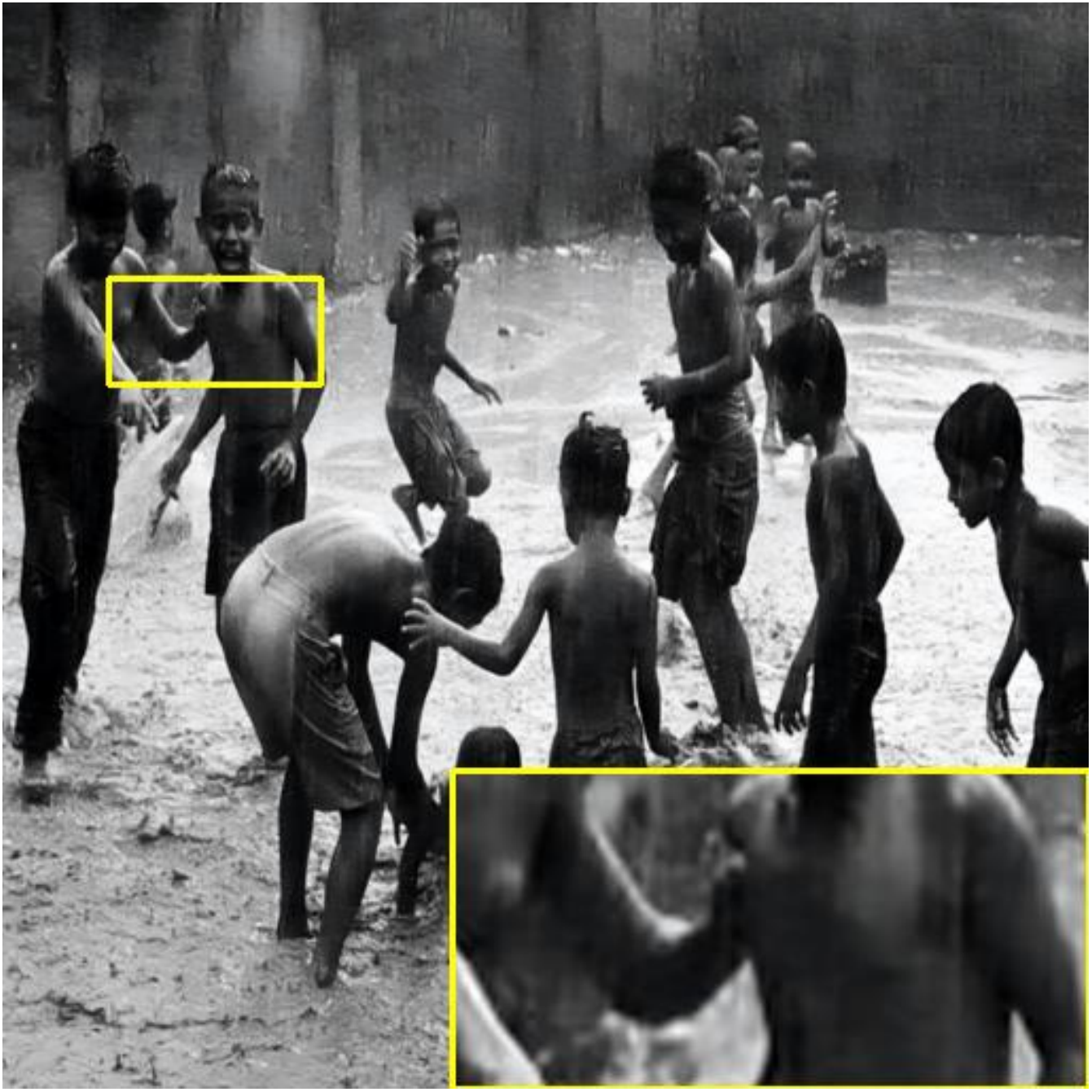}
	\end{subfigure}	
	\begin{subfigure}[t]{0.16\textwidth}
		\includegraphics[width=1\textwidth]{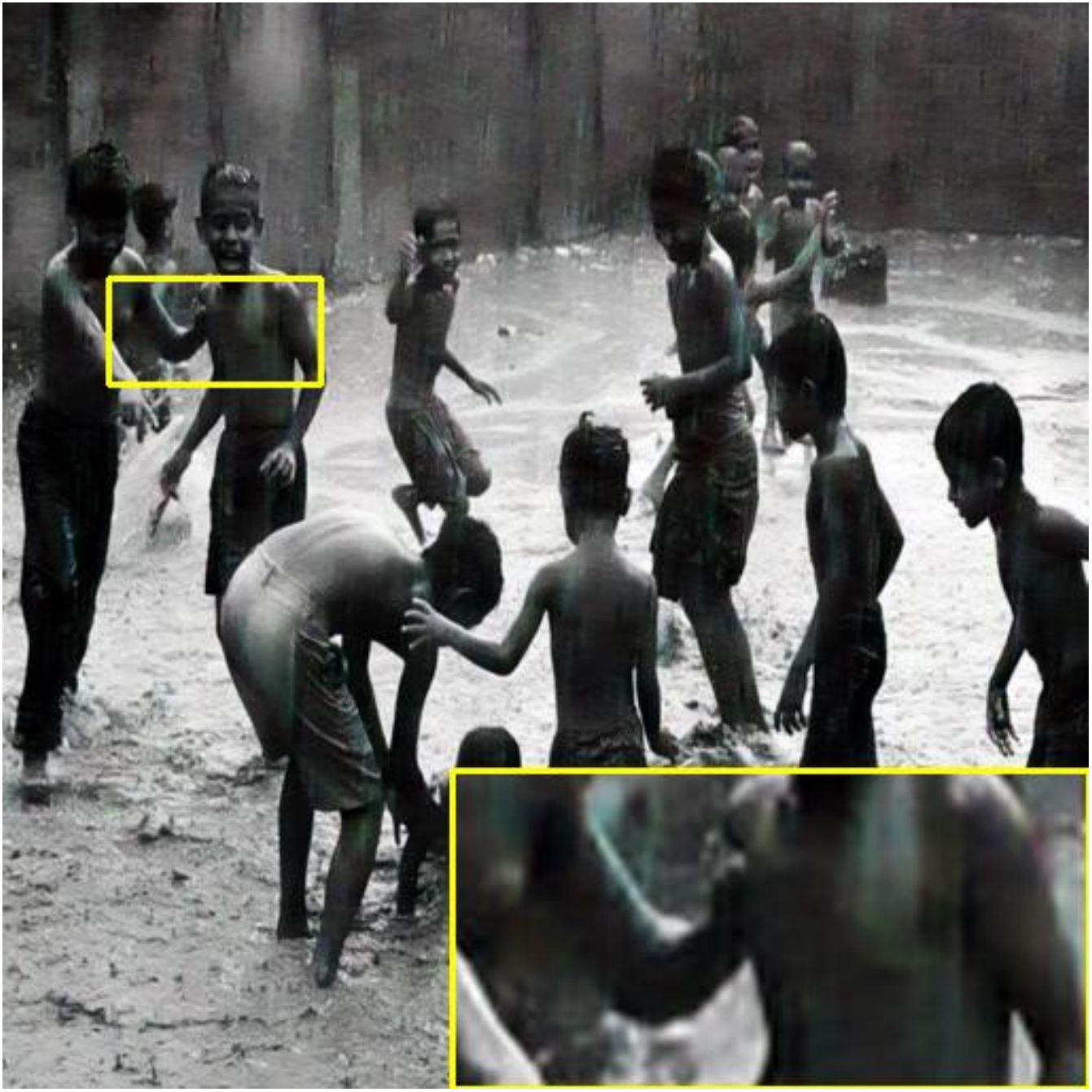}
	\end{subfigure}	
	\begin{subfigure}[t]{0.16\textwidth}
		\includegraphics[width=1\textwidth]{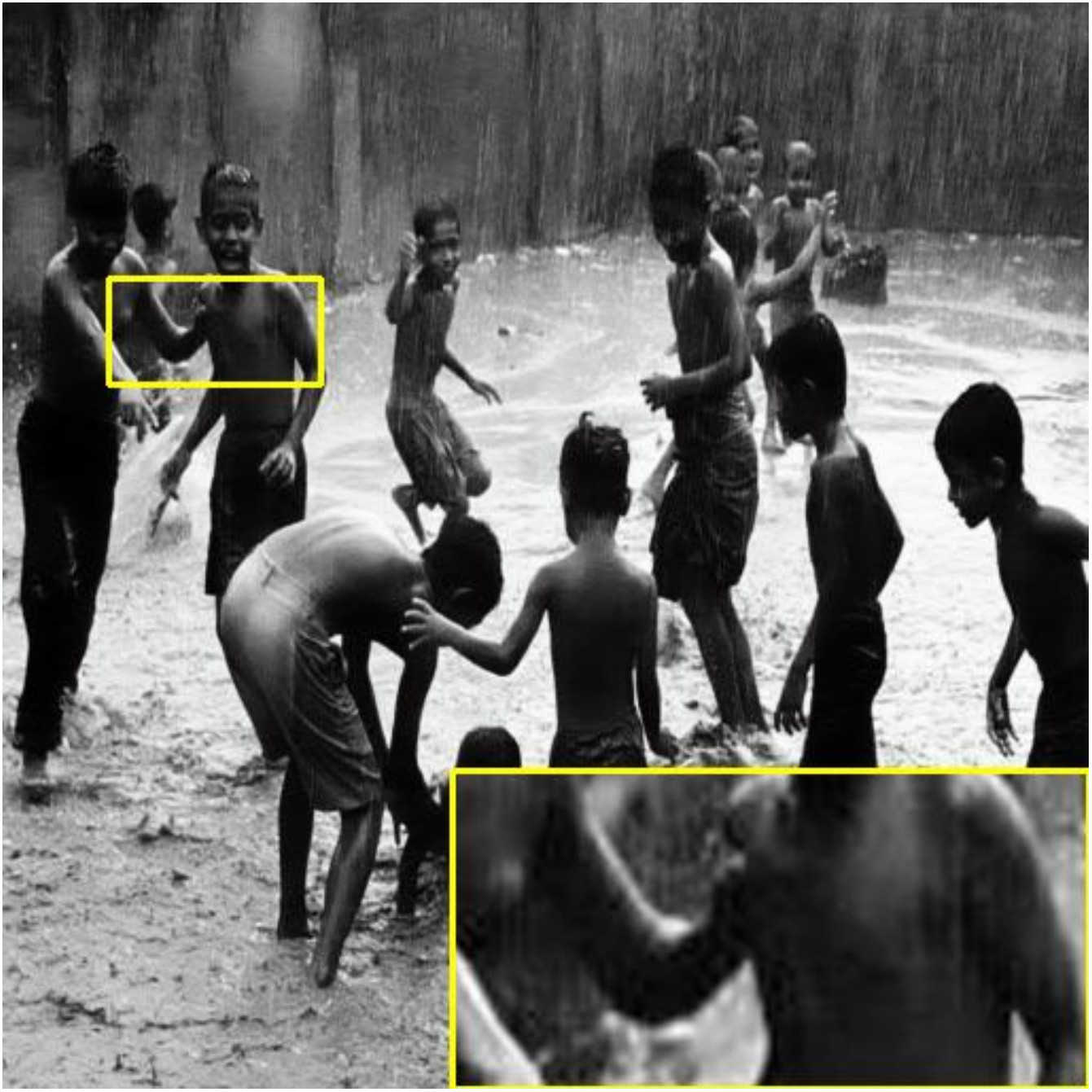}
	\end{subfigure}
    \begin{subfigure}[t]{0.16\textwidth}
		\includegraphics[width=1\textwidth]{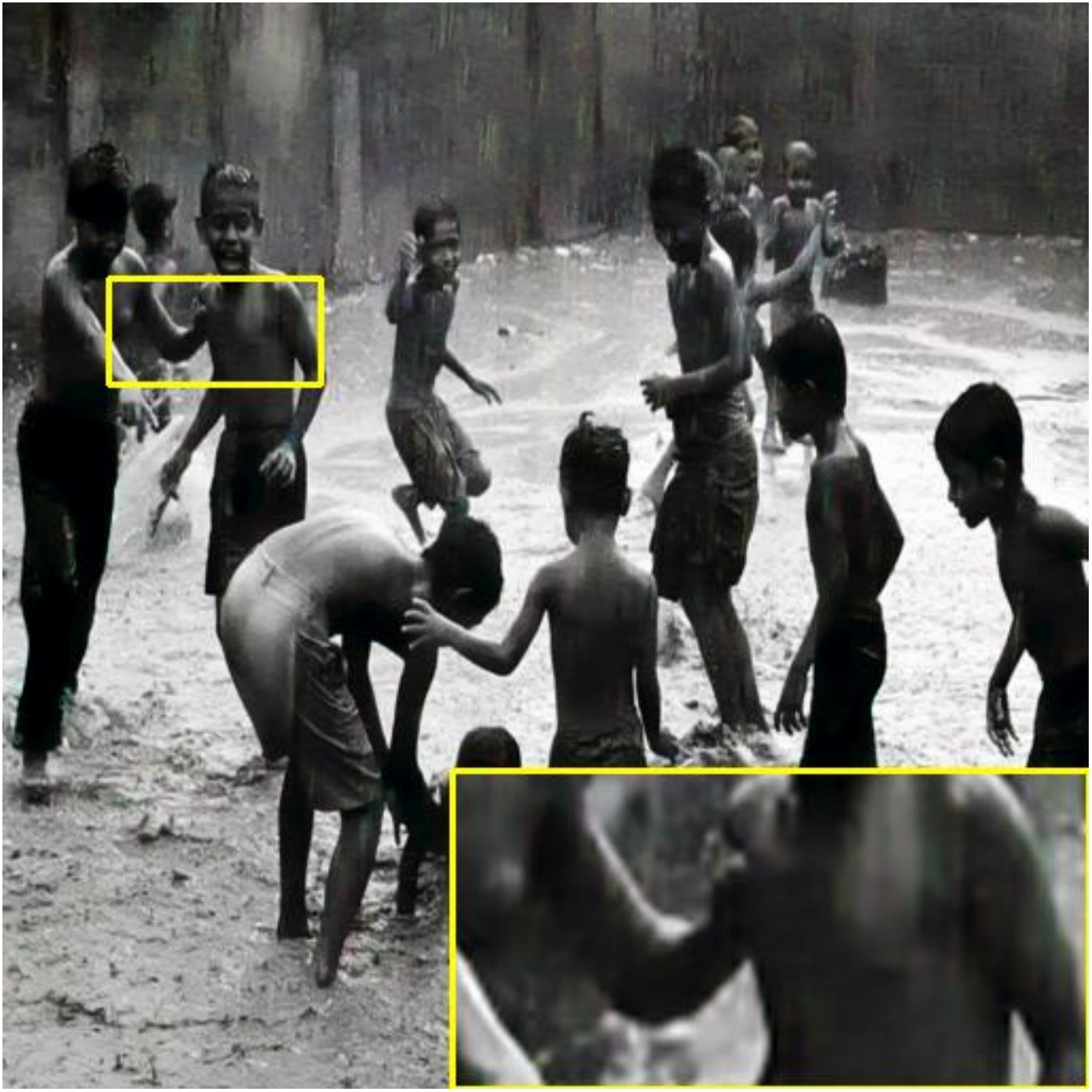}
	\end{subfigure}	
	\begin{subfigure}[t]{0.16\textwidth}
		\includegraphics[width=1\textwidth]{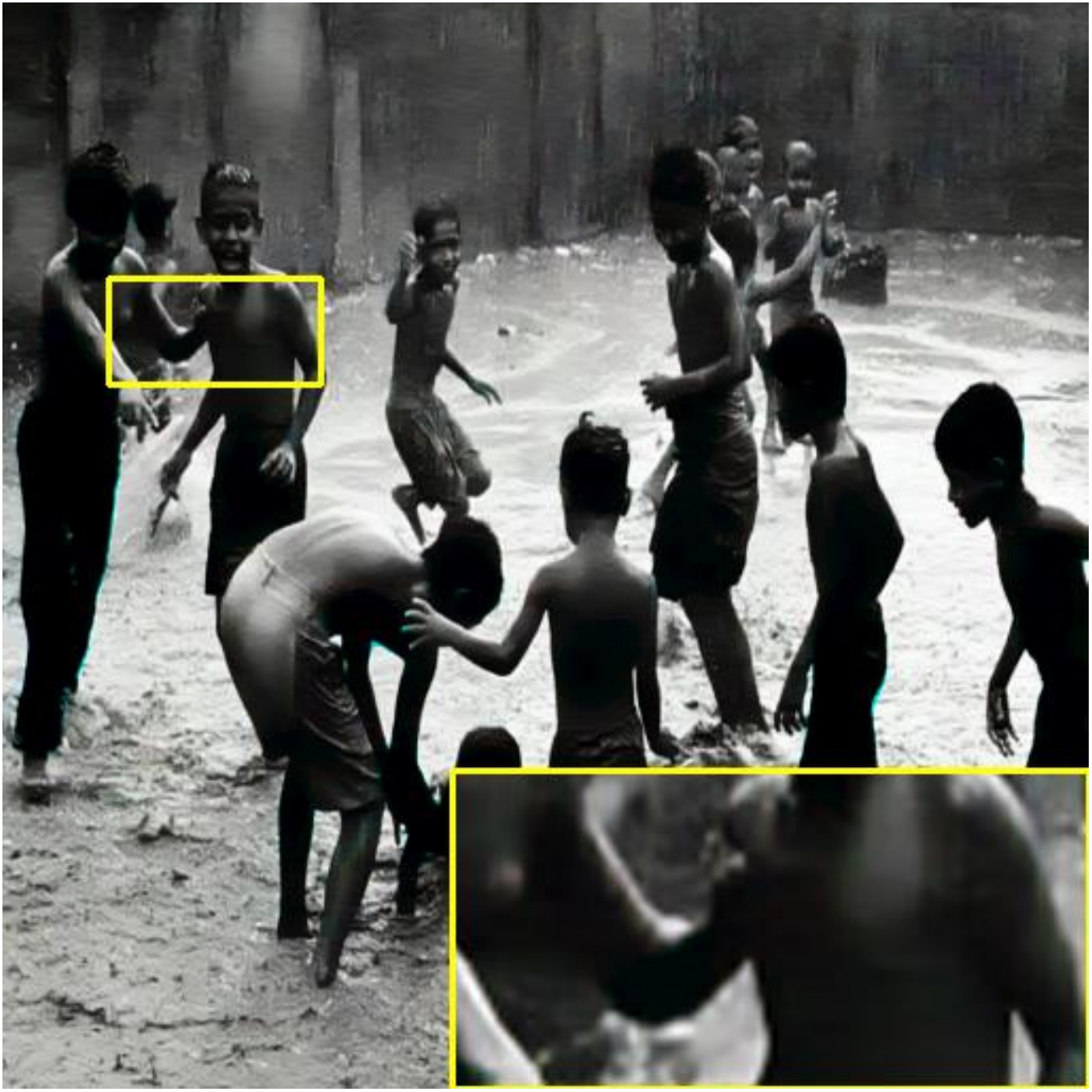}
	\end{subfigure}	
    \begin{subfigure}[t]{0.16\textwidth}
		\includegraphics[width=1\textwidth]{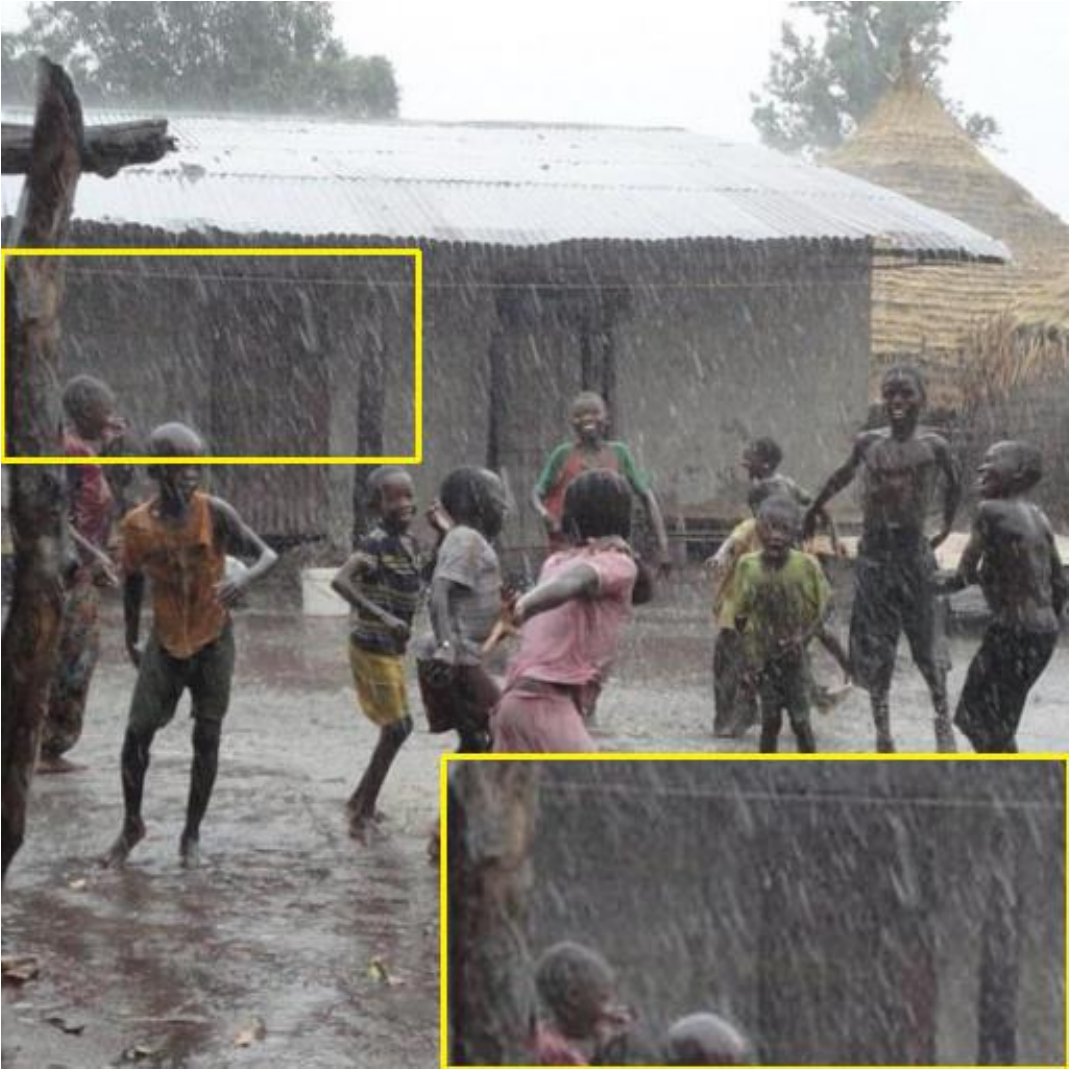}
	\end{subfigure}	
	\begin{subfigure}[t]{0.16\textwidth}
		\includegraphics[width=1\textwidth]{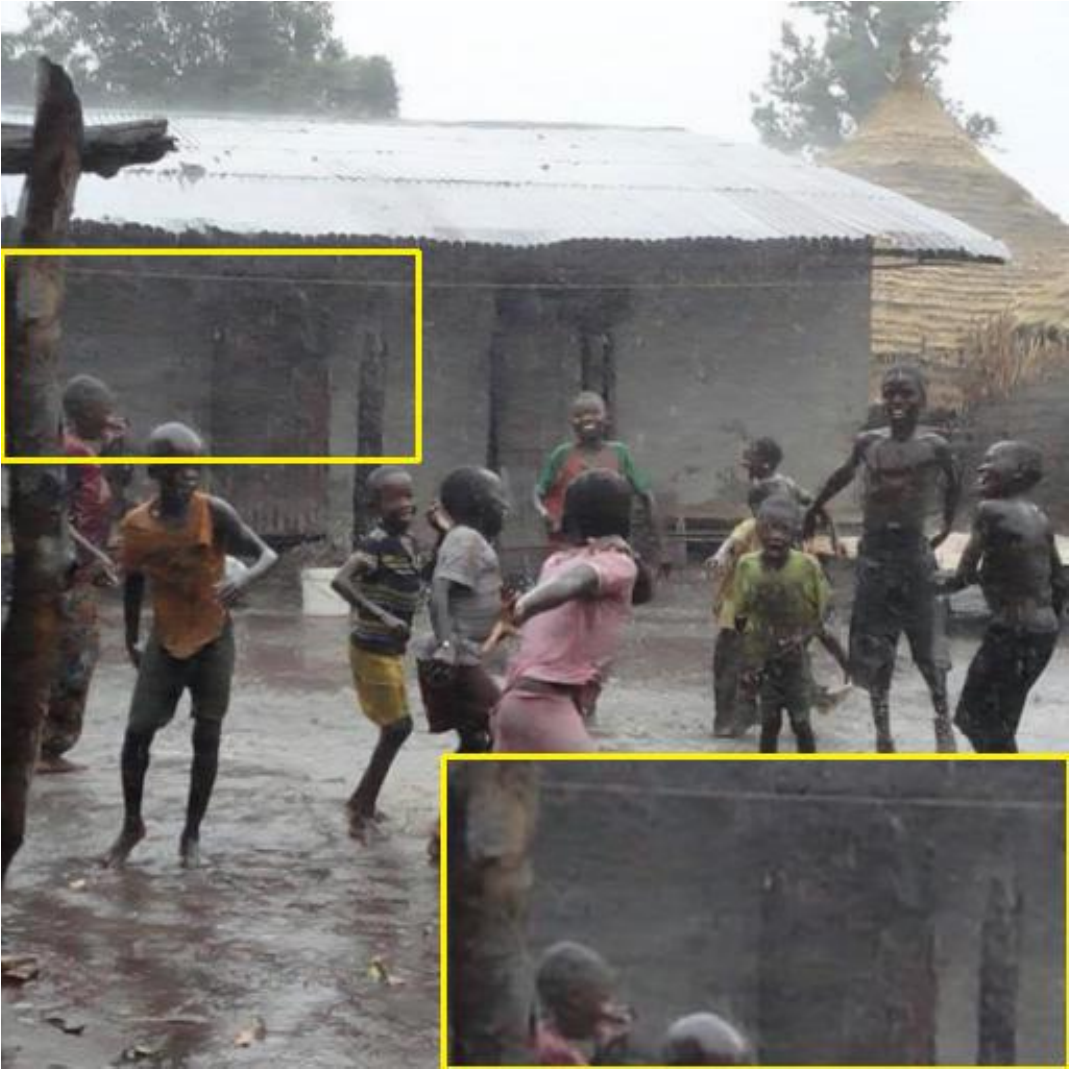}
	\end{subfigure}	
	\begin{subfigure}[t]{0.16\textwidth}
		\includegraphics[width=1\textwidth]{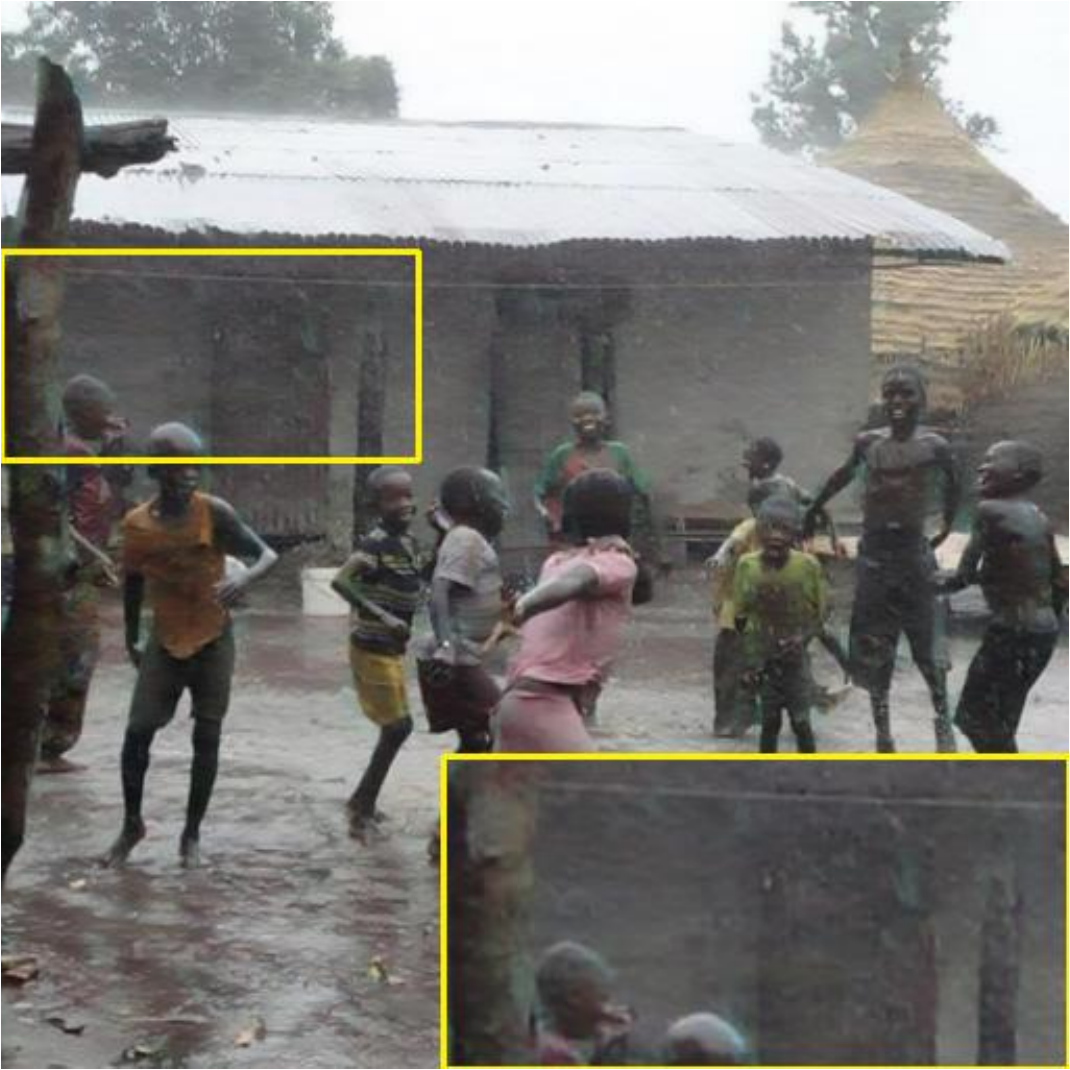}
	\end{subfigure}	
	\begin{subfigure}[t]{0.16\textwidth}
		\includegraphics[width=1\textwidth]{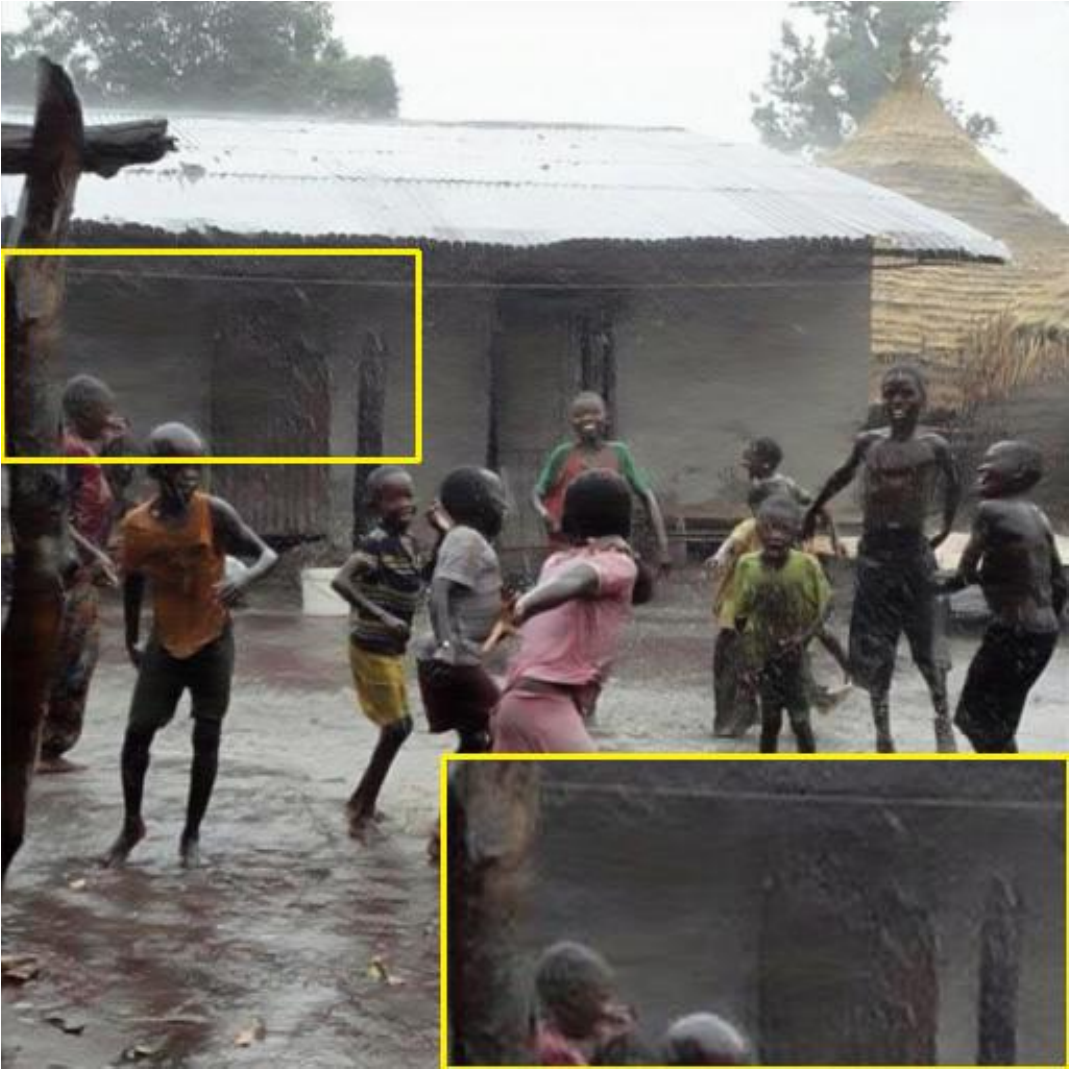}
	\end{subfigure}
    \begin{subfigure}[t]{0.16\textwidth}
		\includegraphics[width=1\textwidth]{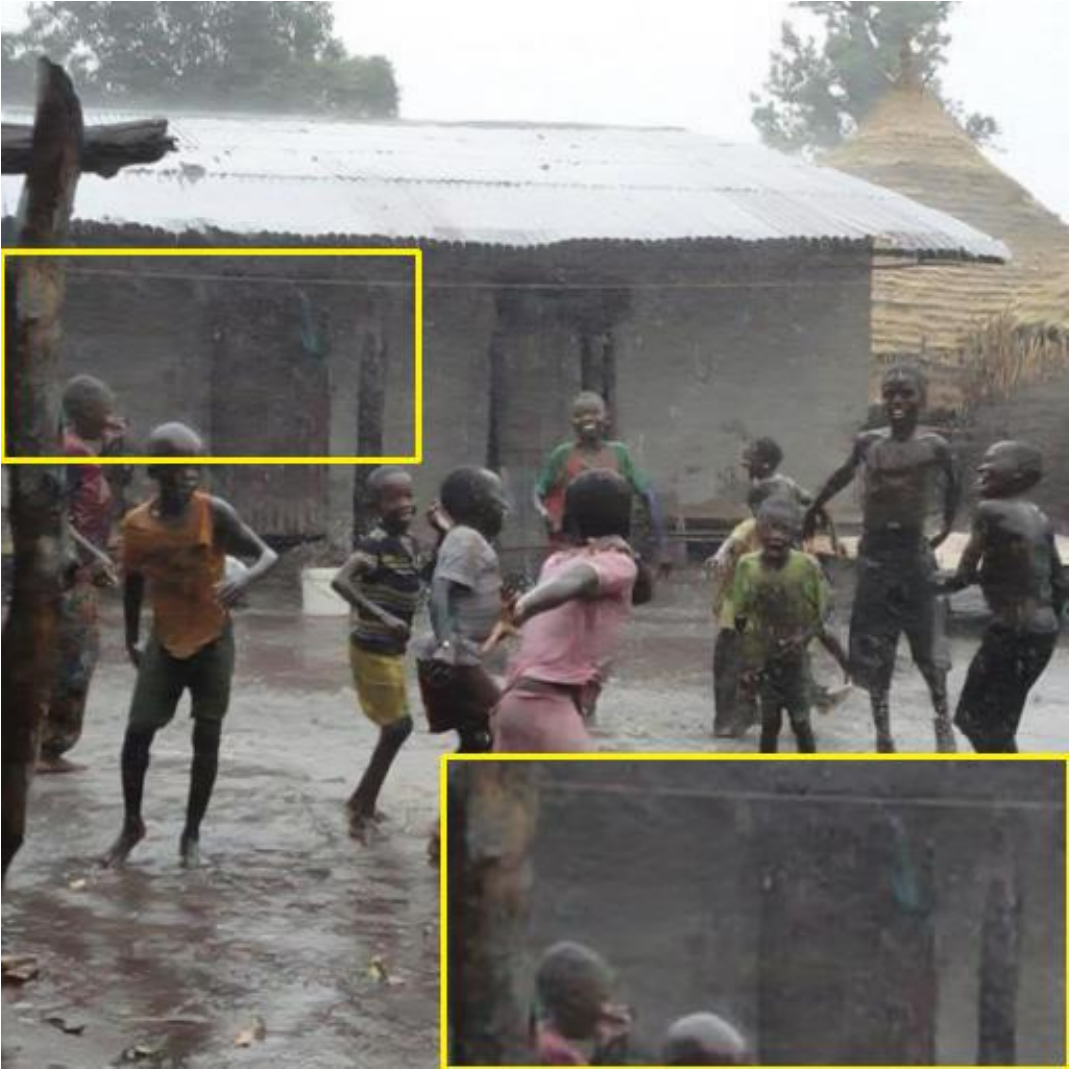}
	\end{subfigure}	
	\begin{subfigure}[t]{0.16\textwidth}
		\includegraphics[width=1\textwidth]{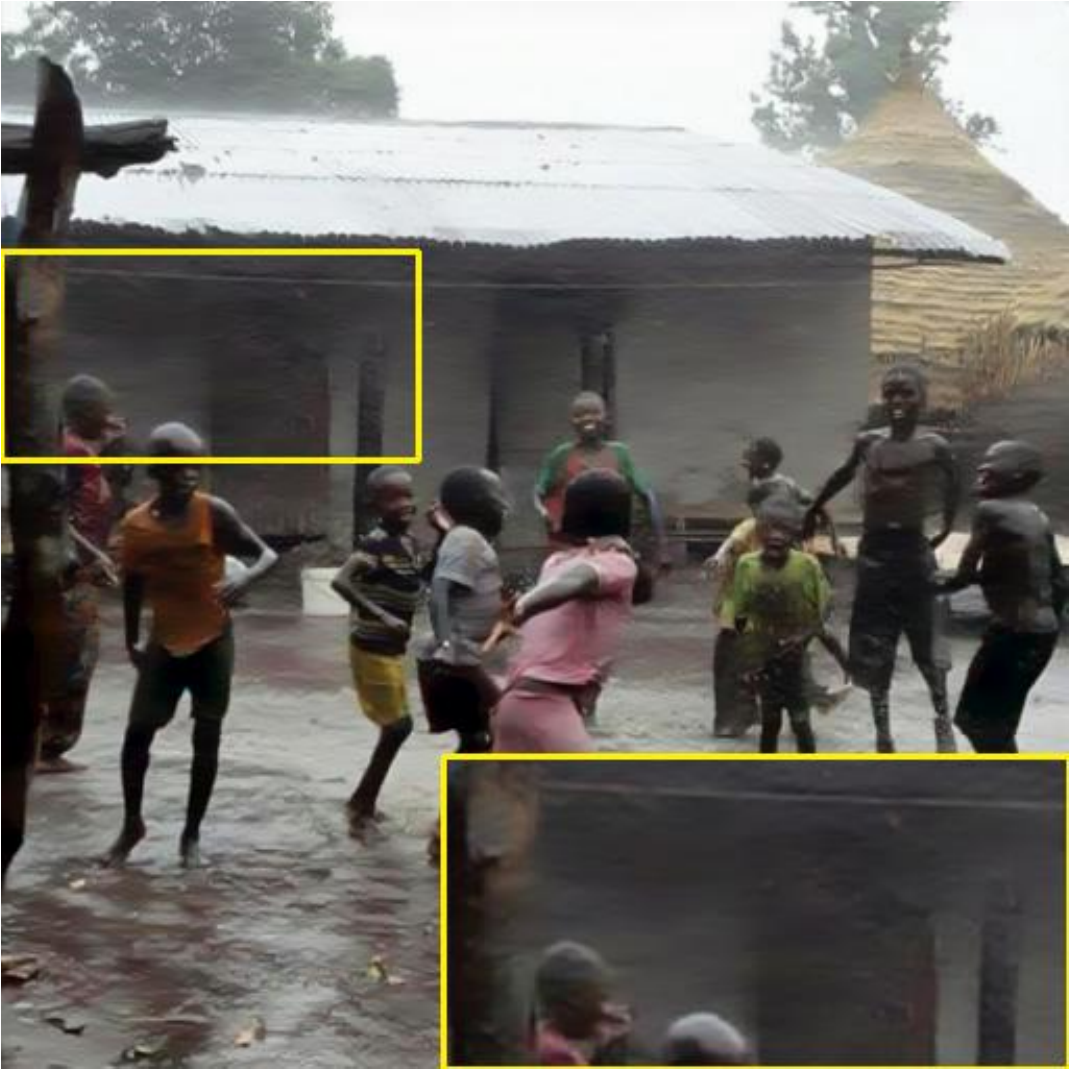}
	\end{subfigure}	
	\begin{subfigure}[t]{0.16\textwidth}
		\includegraphics[width=1\textwidth]{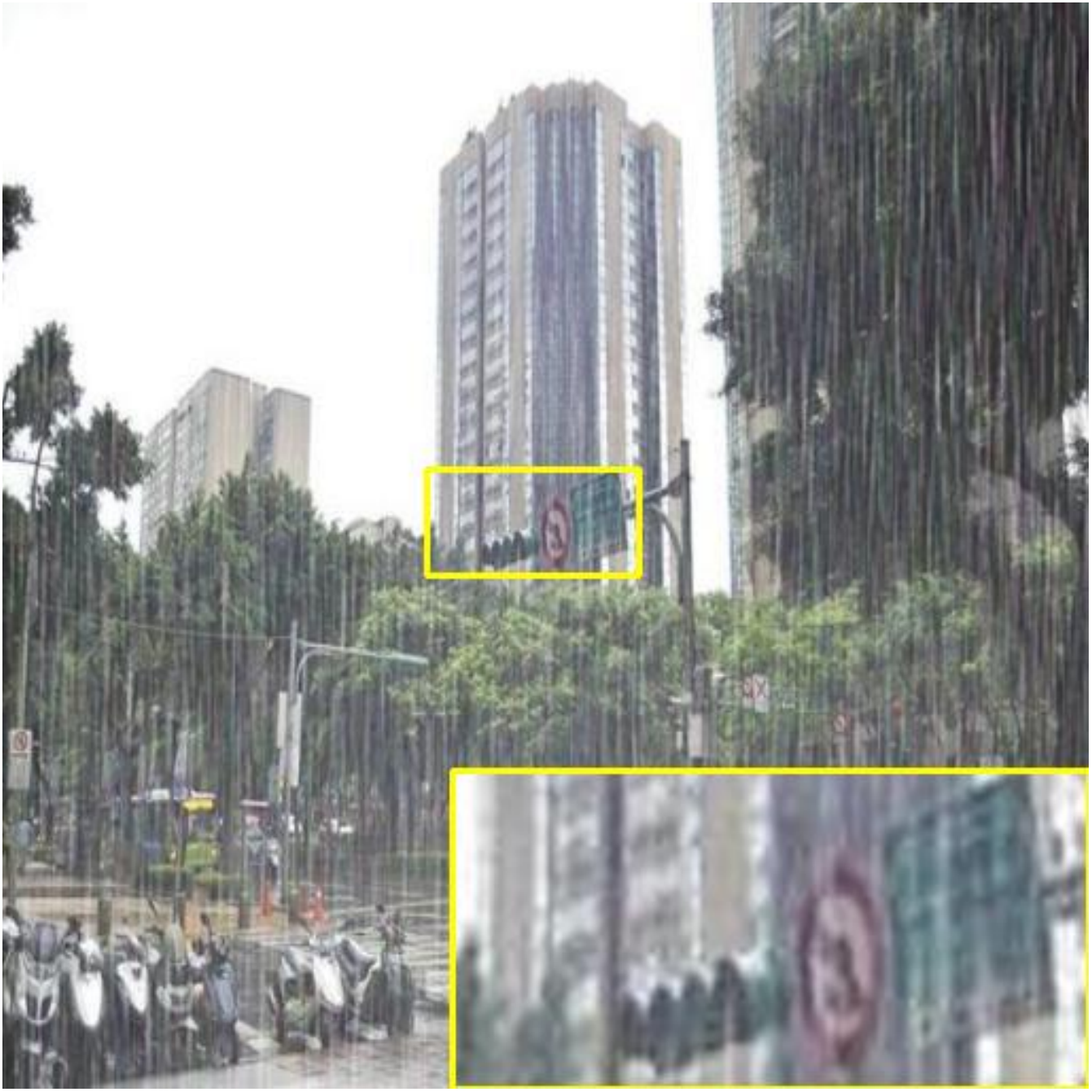}
	\end{subfigure}	
	\begin{subfigure}[t]{0.16\textwidth}
		\includegraphics[width=1\textwidth]{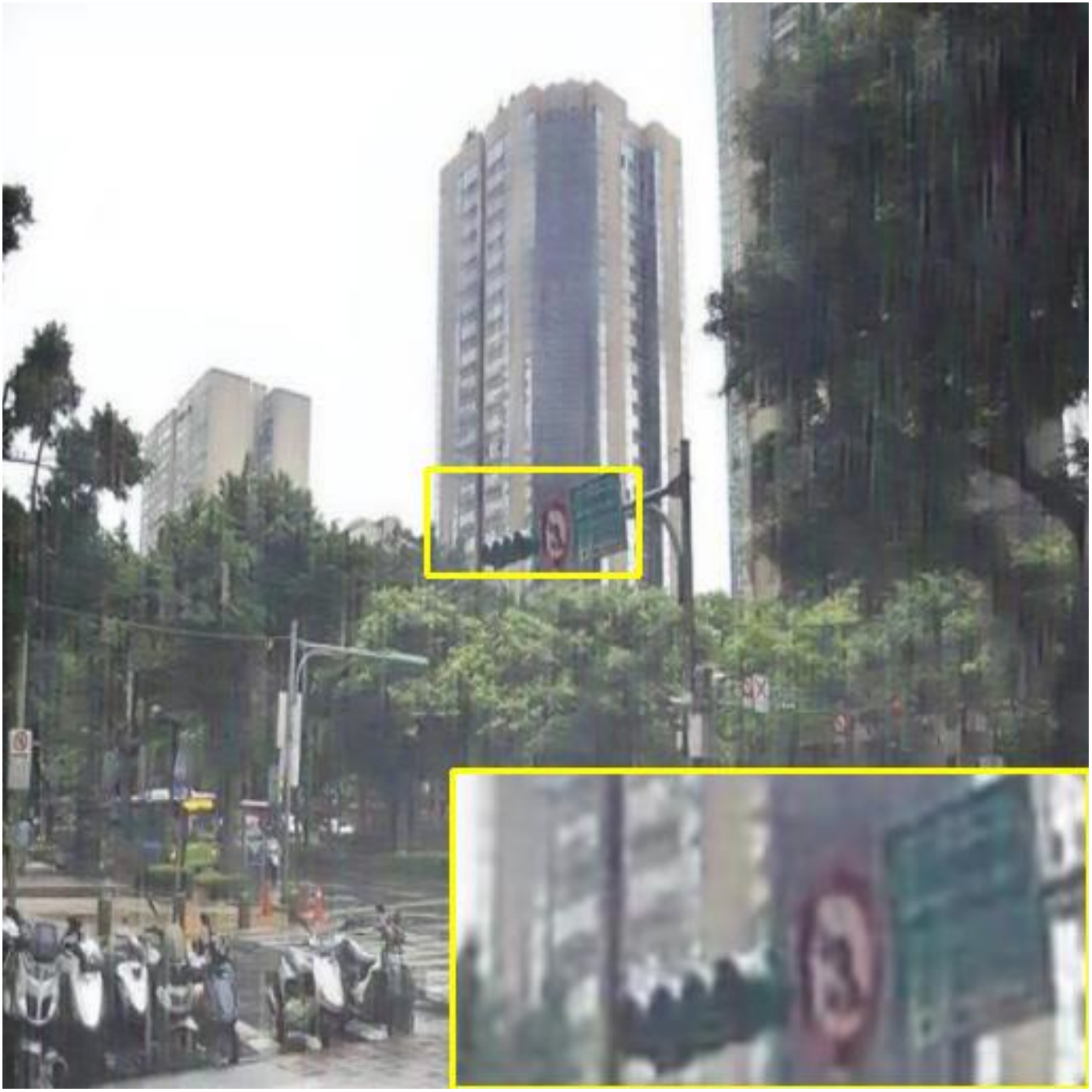}
	\end{subfigure}	
	\begin{subfigure}[t]{0.16\textwidth}
		\includegraphics[width=1\textwidth]{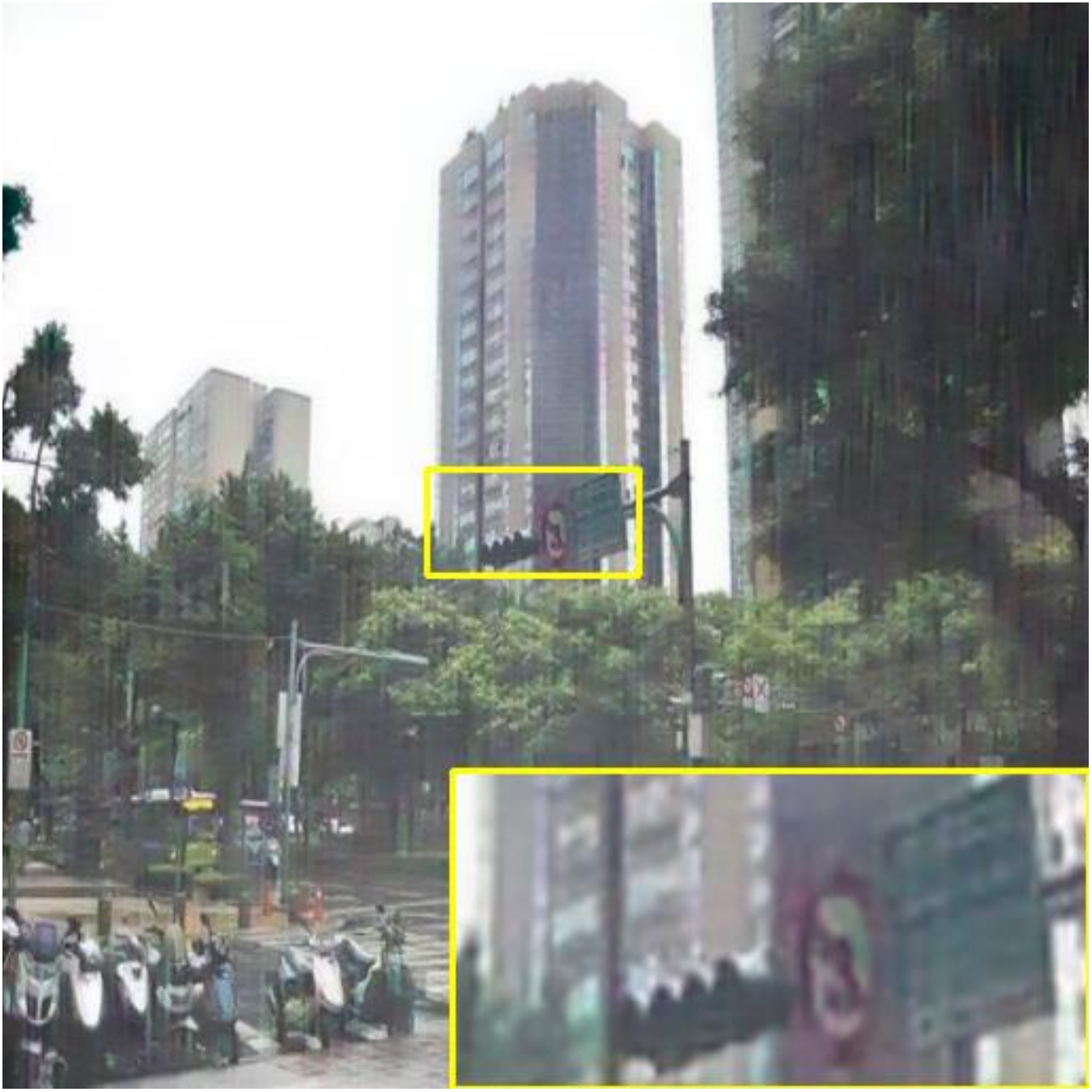}
	\end{subfigure}	
	\begin{subfigure}[t]{0.16\textwidth}
		\includegraphics[width=1\textwidth]{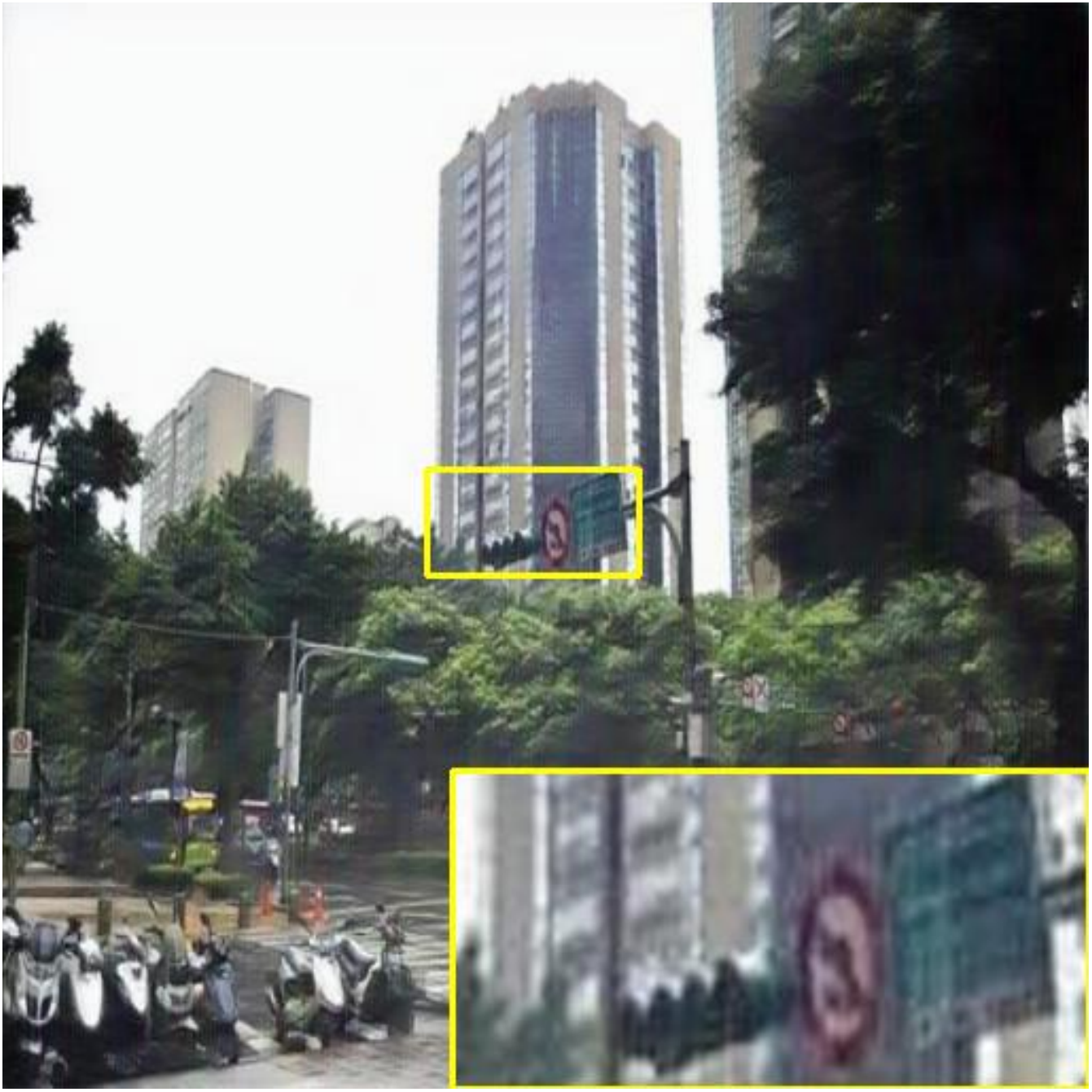}
	\end{subfigure}
    \begin{subfigure}[t]{0.16\textwidth}
		\includegraphics[width=1\textwidth]{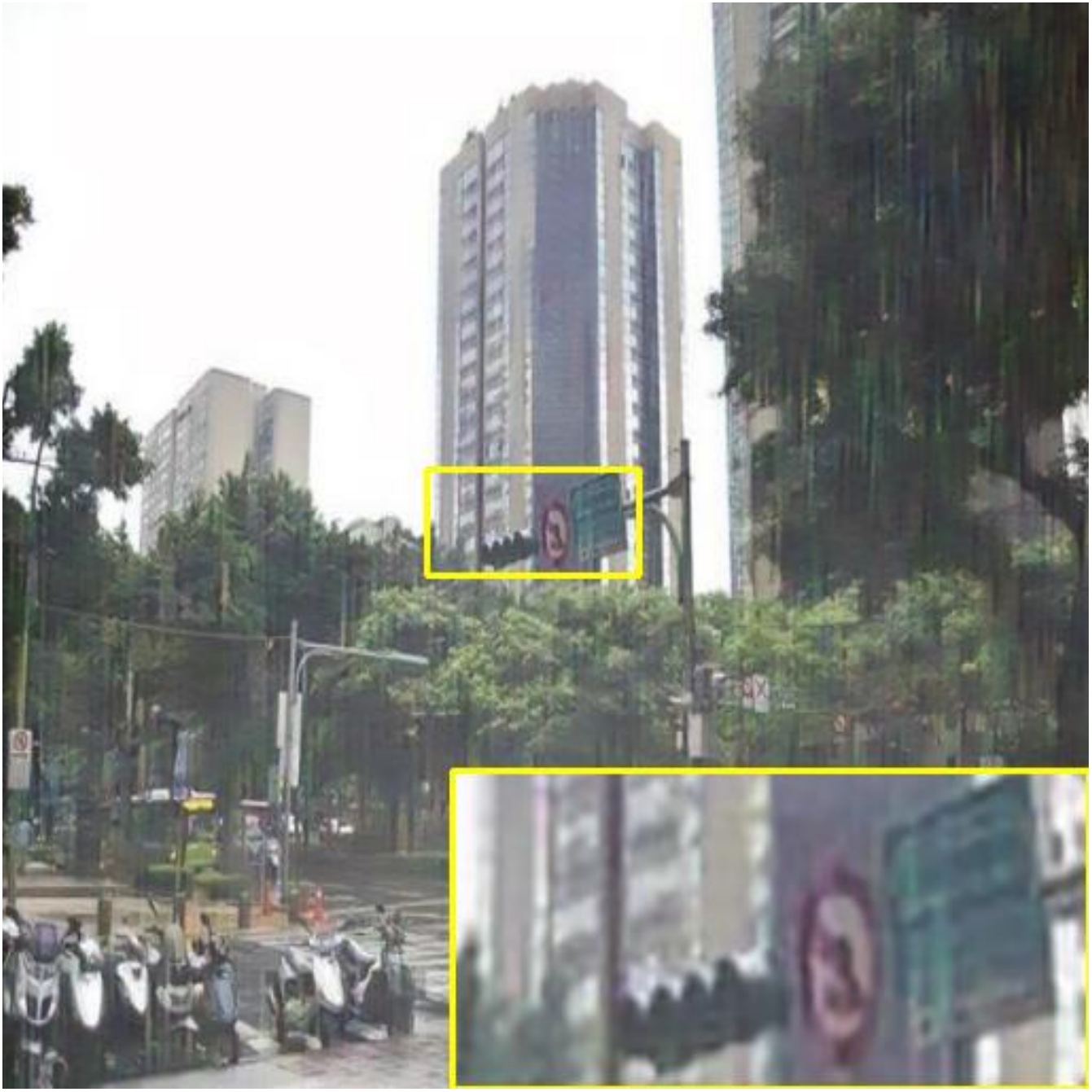}
	\end{subfigure}	
	\begin{subfigure}[t]{0.16\textwidth}
		\includegraphics[width=1\textwidth]{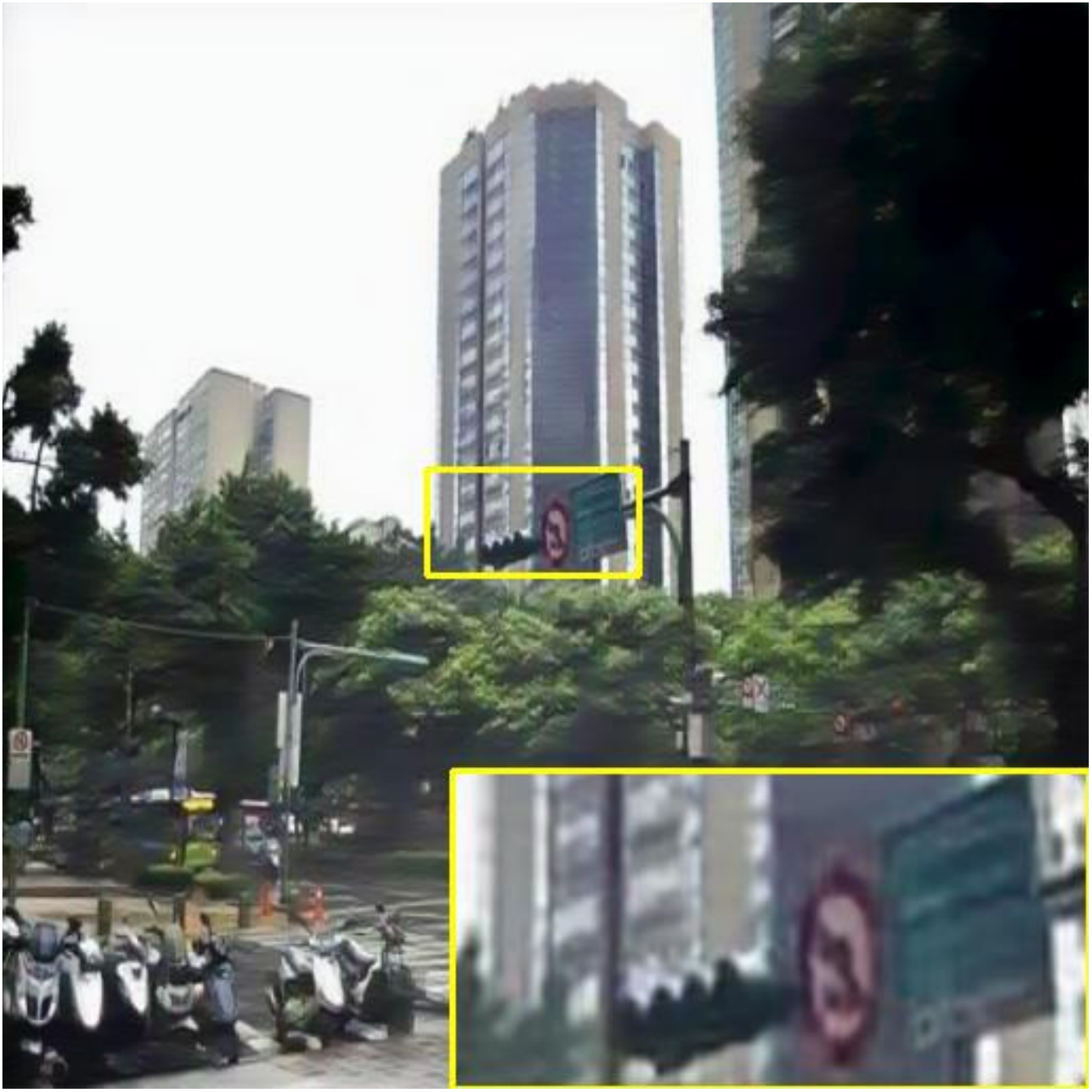}
	\end{subfigure}	
	\begin{subfigure}[t]{0.16\textwidth}
		\includegraphics[width=1\textwidth]{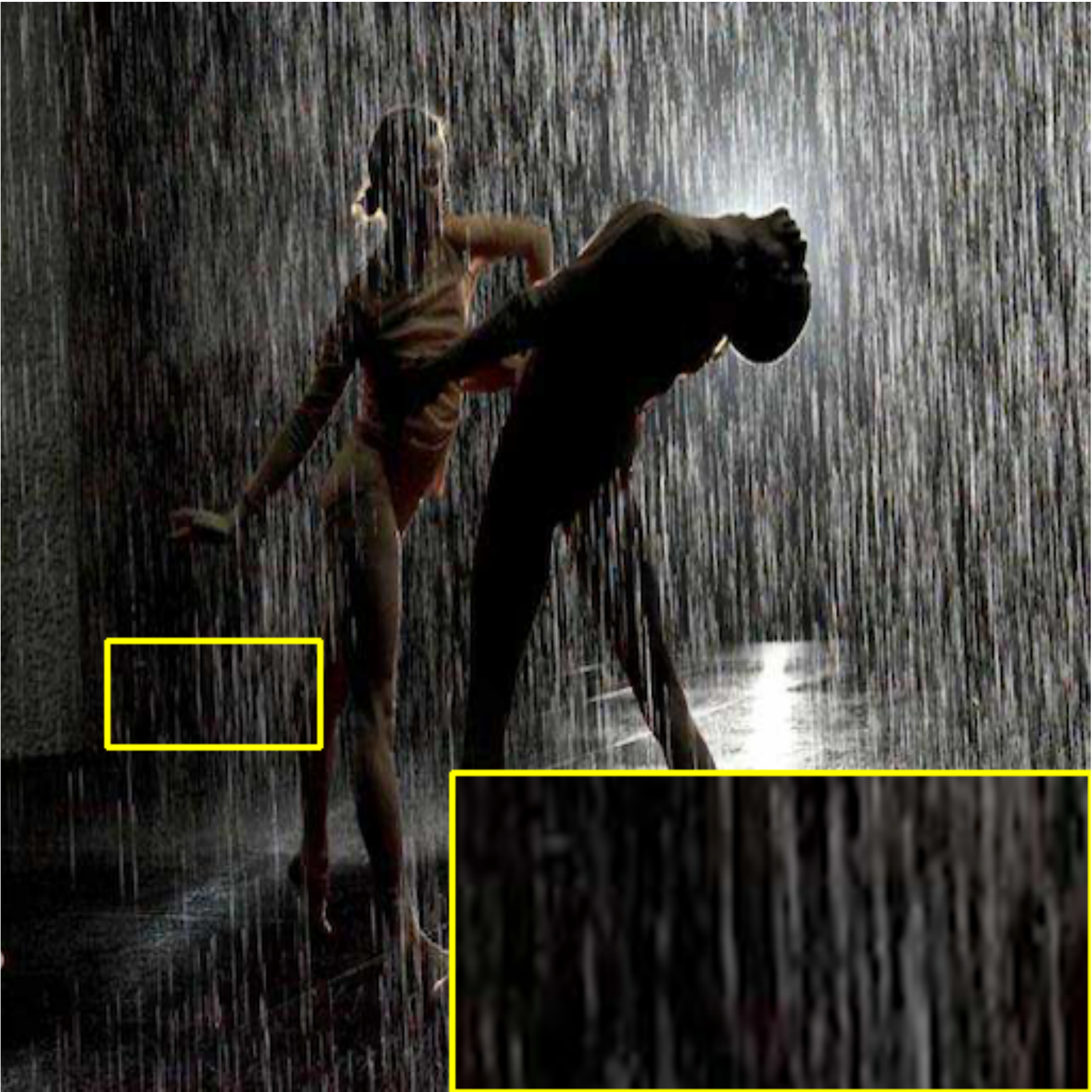}
	\end{subfigure}	
	\begin{subfigure}[t]{0.16\textwidth}
		\includegraphics[width=1\textwidth]{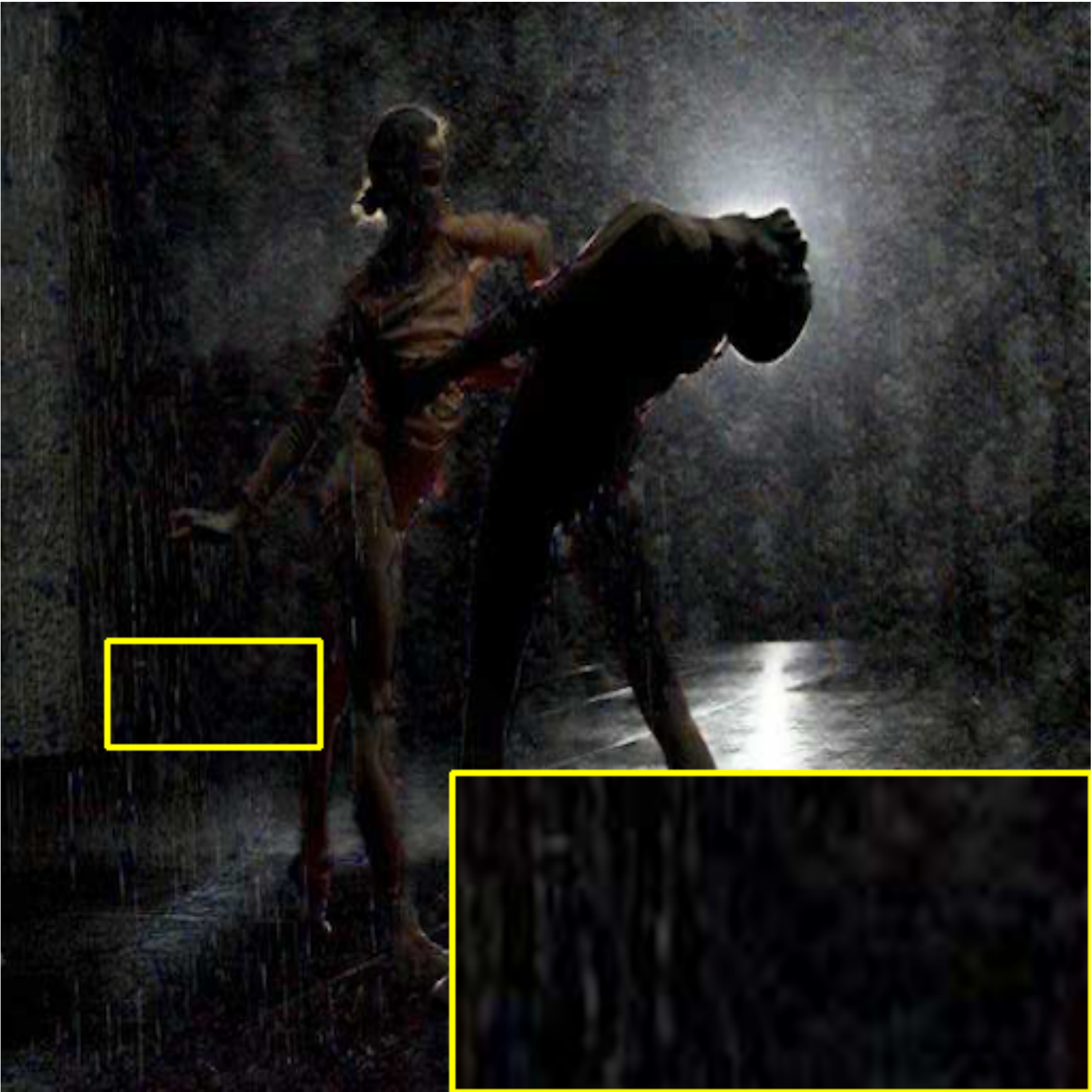}
	\end{subfigure}	
	\begin{subfigure}[t]{0.16\textwidth}
		\includegraphics[width=1\textwidth]{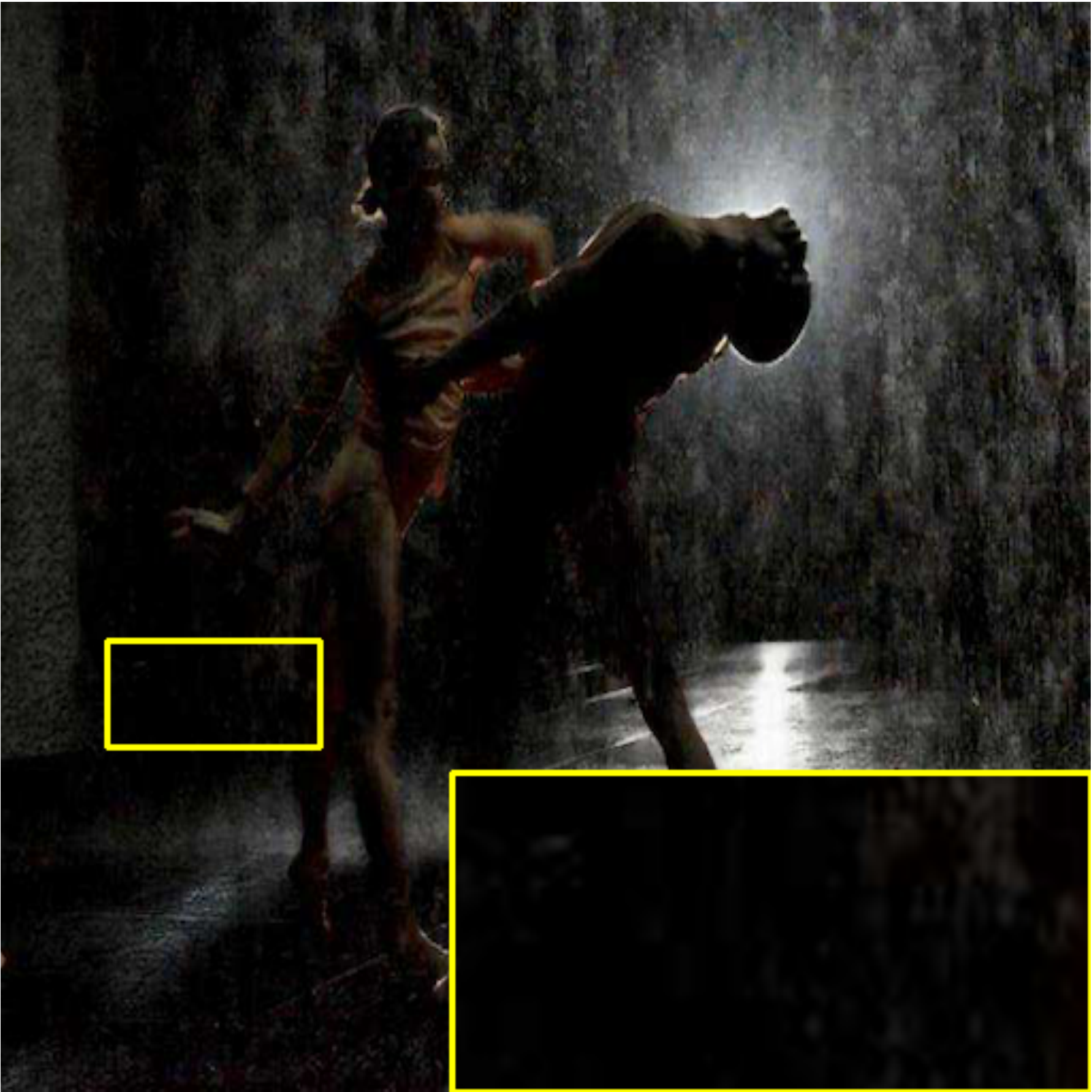}
	\end{subfigure}	
	\begin{subfigure}[t]{0.16\textwidth}
		\includegraphics[width=1\textwidth]{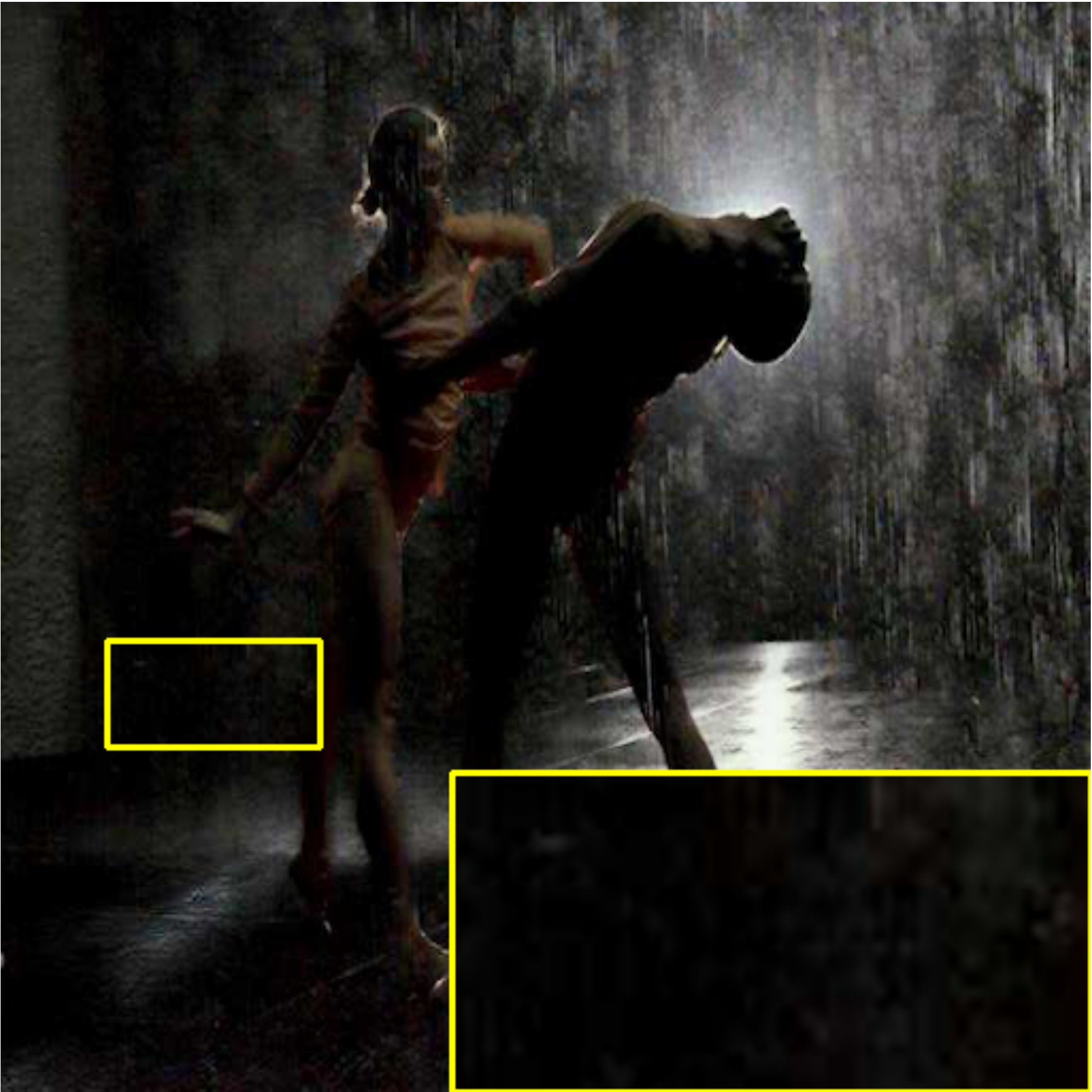}
	\end{subfigure}
    \begin{subfigure}[t]{0.16\textwidth}
		\includegraphics[width=1\textwidth]{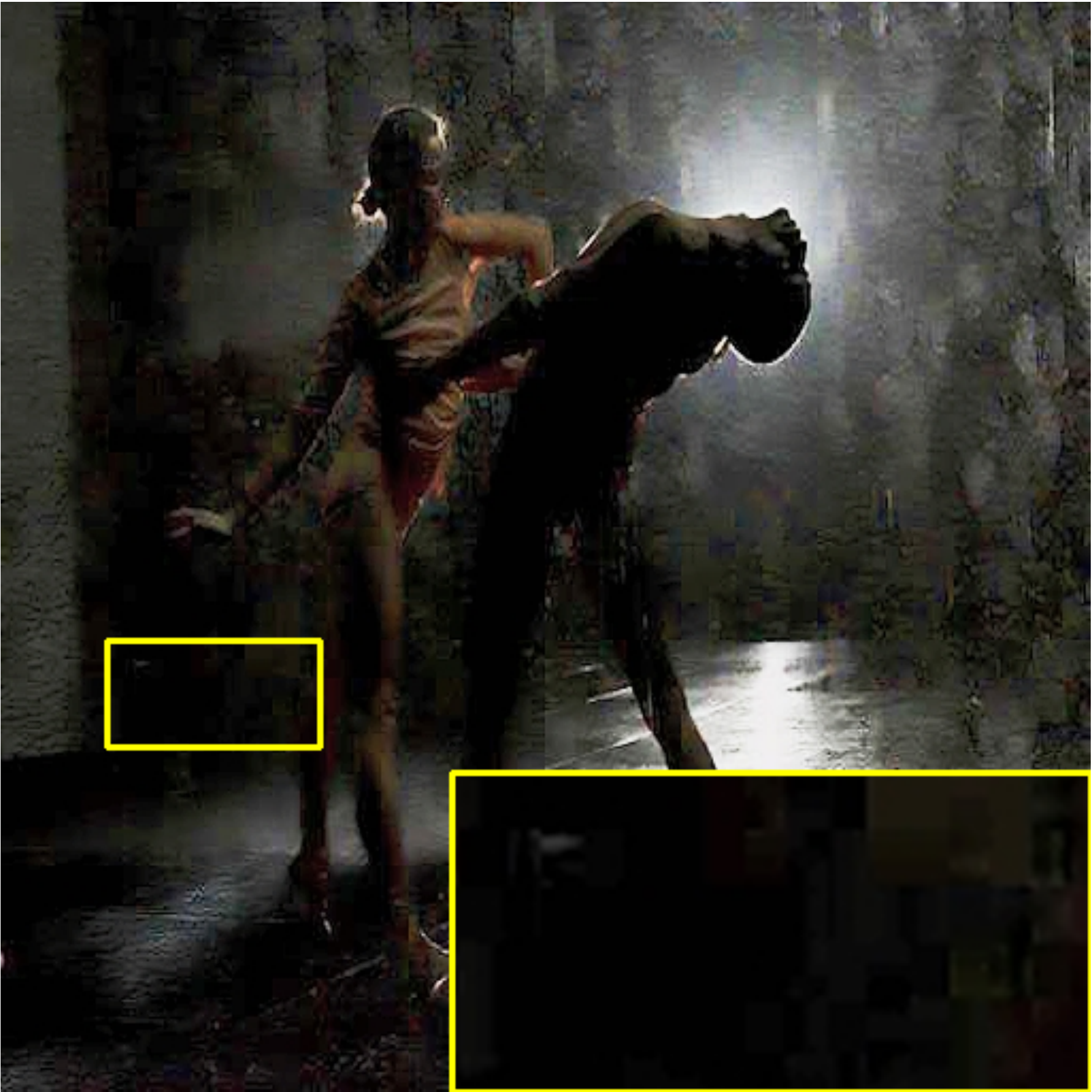}
	\end{subfigure}	
	\begin{subfigure}[t]{0.16\textwidth}
		\includegraphics[width=1\textwidth]{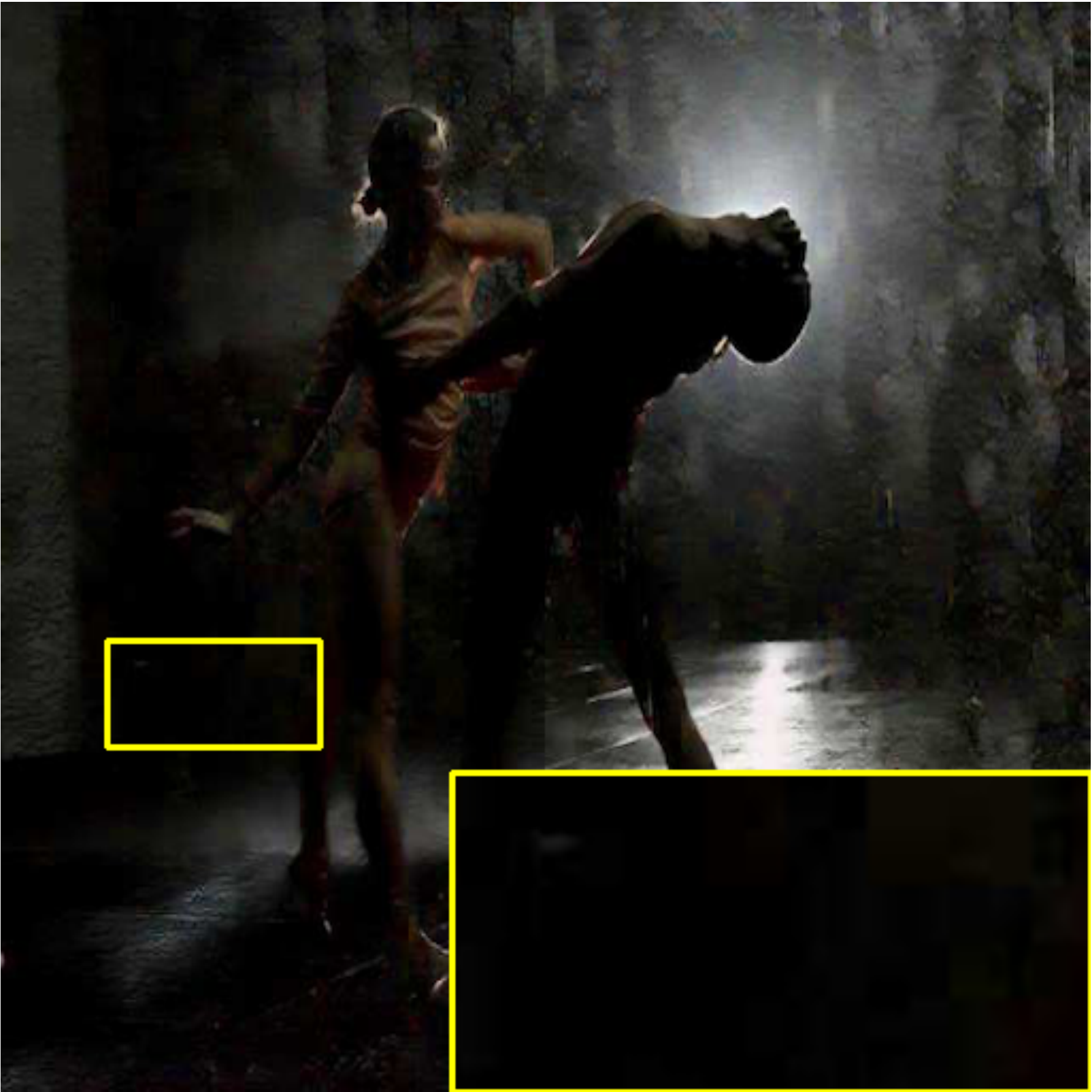}
	\end{subfigure}	
		\begin{subfigure}[t]{0.16\textwidth}
		\includegraphics[width=1\textwidth]{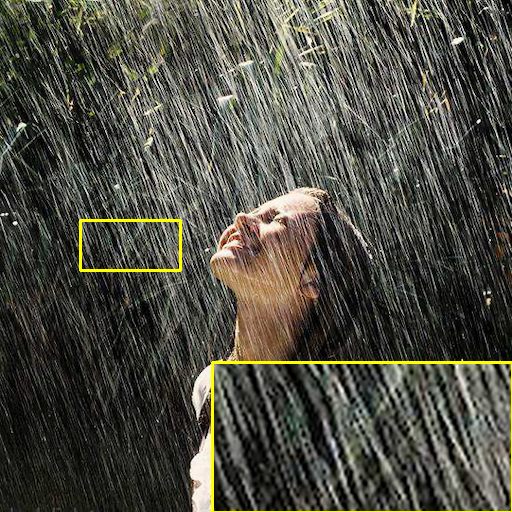}
	\end{subfigure}	
	\begin{subfigure}[t]{0.16\textwidth}
		\includegraphics[width=1\textwidth]{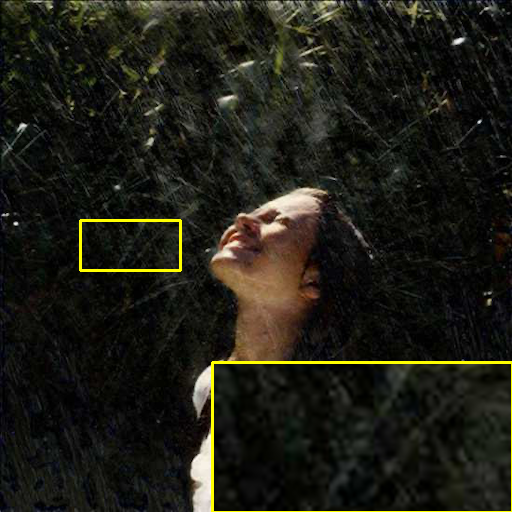}
	\end{subfigure}	
	\begin{subfigure}[t]{0.16\textwidth}
		\includegraphics[width=1\textwidth]{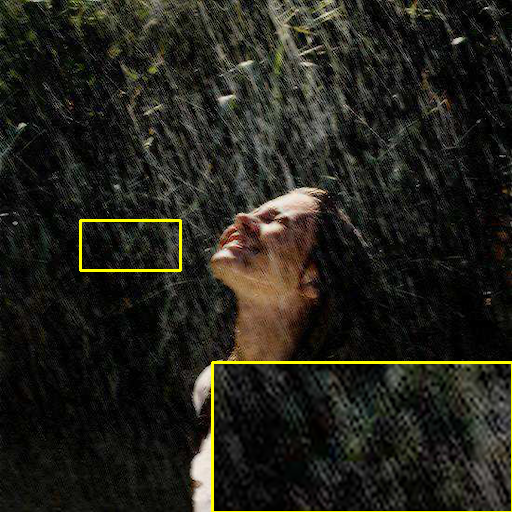}
	\end{subfigure}	
	\begin{subfigure}[t]{0.16\textwidth}
		\includegraphics[width=1\textwidth]{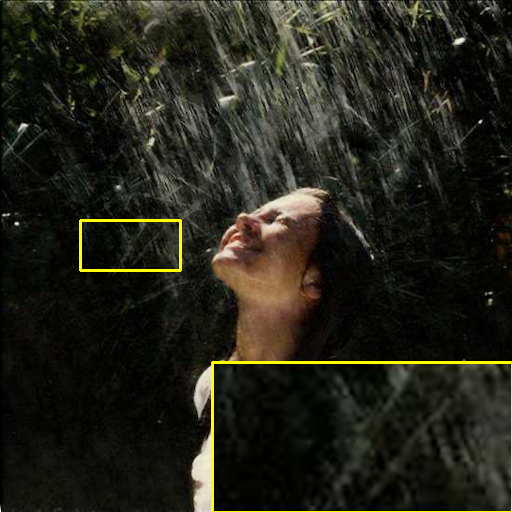}
	\end{subfigure}
    \begin{subfigure}[t]{0.16\textwidth}
		\includegraphics[width=1\textwidth]{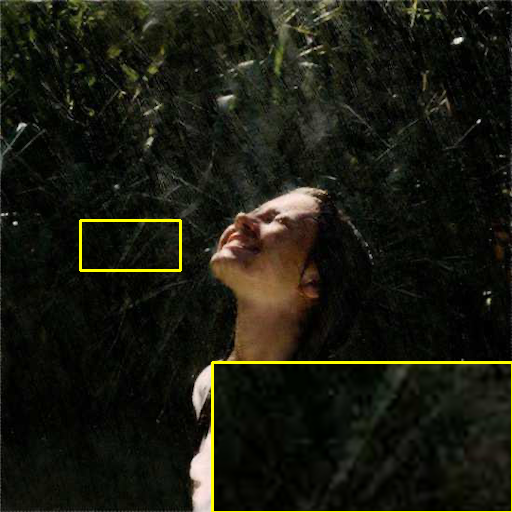}
	\end{subfigure}	
	\begin{subfigure}[t]{0.16\textwidth}
		\includegraphics[width=1\textwidth]{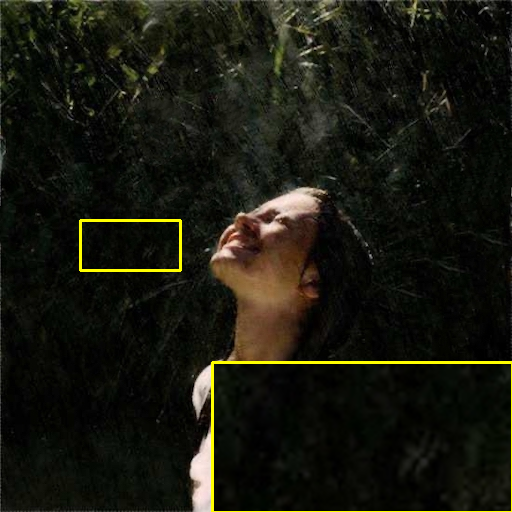}
	\end{subfigure}	
	\begin{subfigure}[t]{0.16\textwidth}
		\includegraphics[width=1\textwidth]{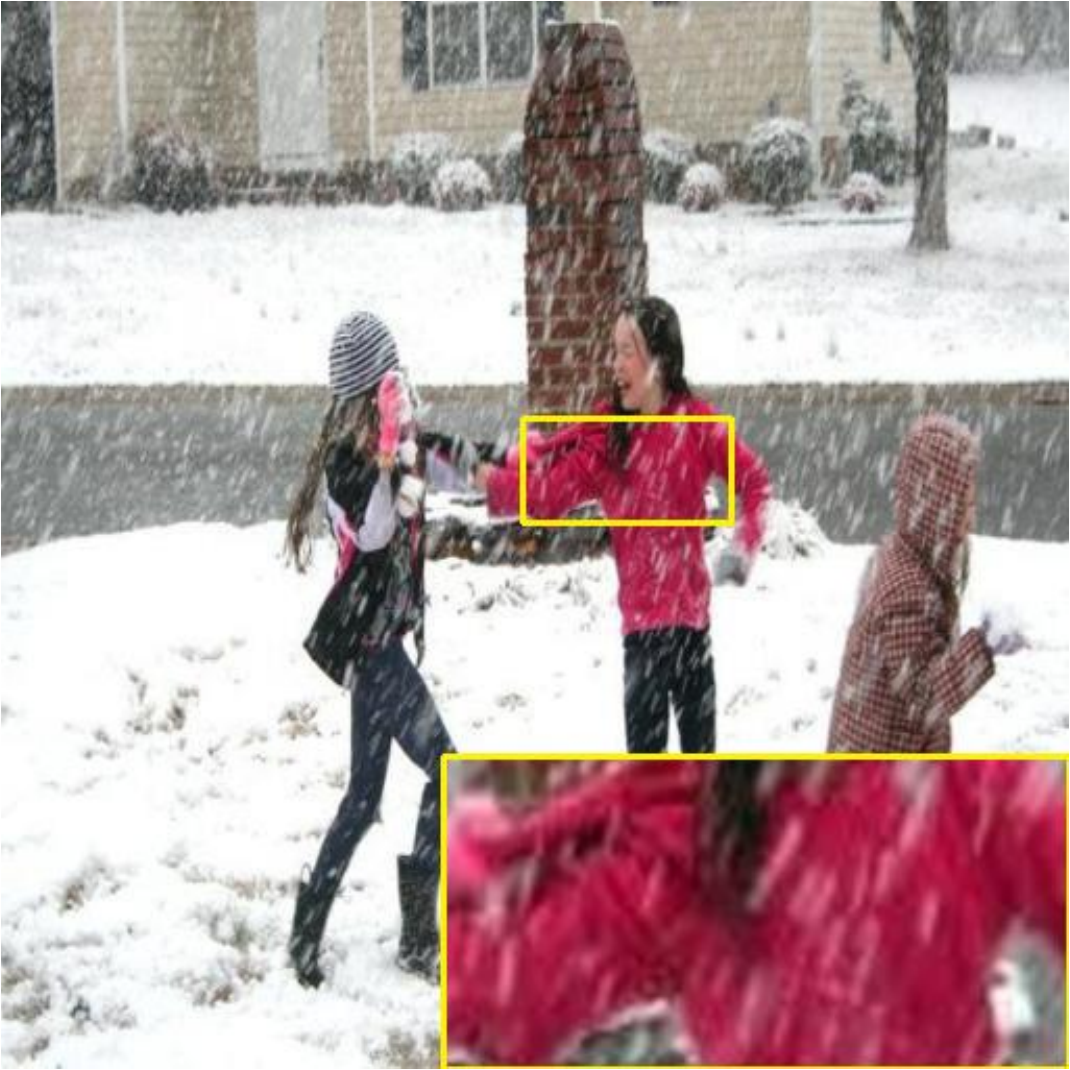}
		\vspace{-6mm}
		\subcaption*{(a) Rainy Image}
	\end{subfigure}	
	\begin{subfigure}[t]{0.16\textwidth}
		\includegraphics[width=1\textwidth]{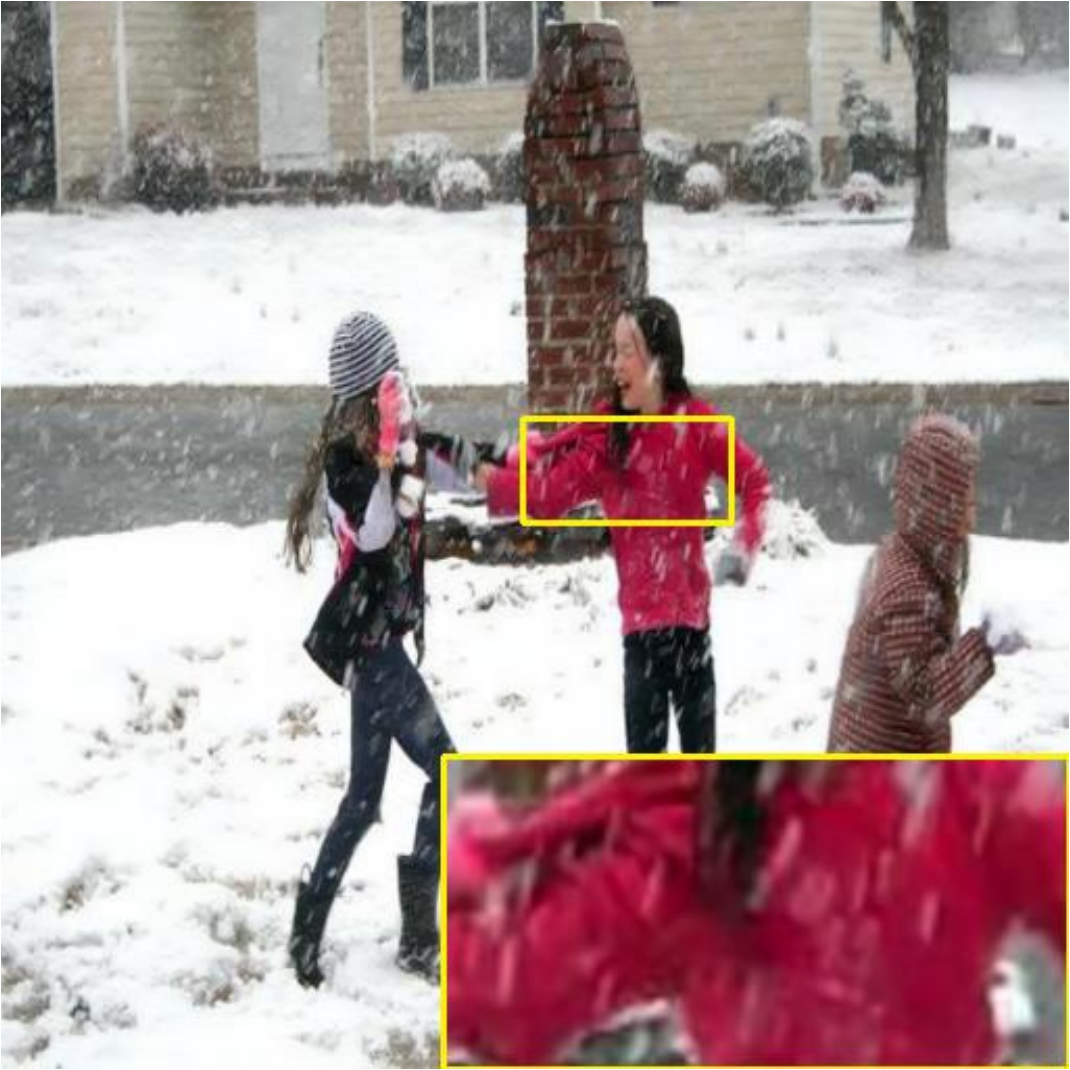}
		\vspace{-6mm}
		\subcaption*{(b) DDN~\cite{fu2017removing}}
	\end{subfigure}	
	\begin{subfigure}[t]{0.16\textwidth}
		\includegraphics[width=1\textwidth]{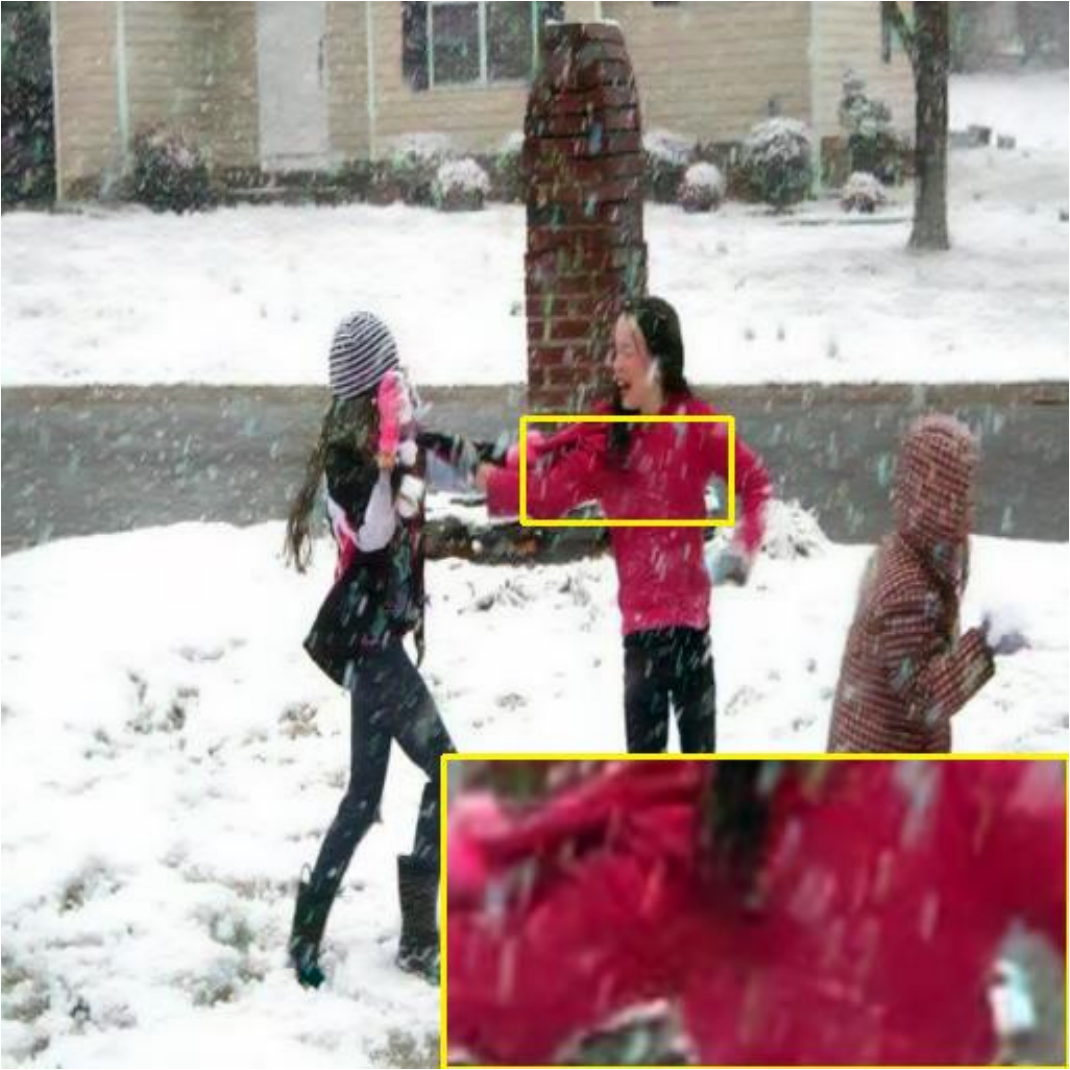}
		\vspace{-6mm}
		\subcaption*{(c) JORDER~\cite{yang2017deeppami}}
	\end{subfigure}	
	\begin{subfigure}[t]{0.16\textwidth}
		\includegraphics[width=1\textwidth]{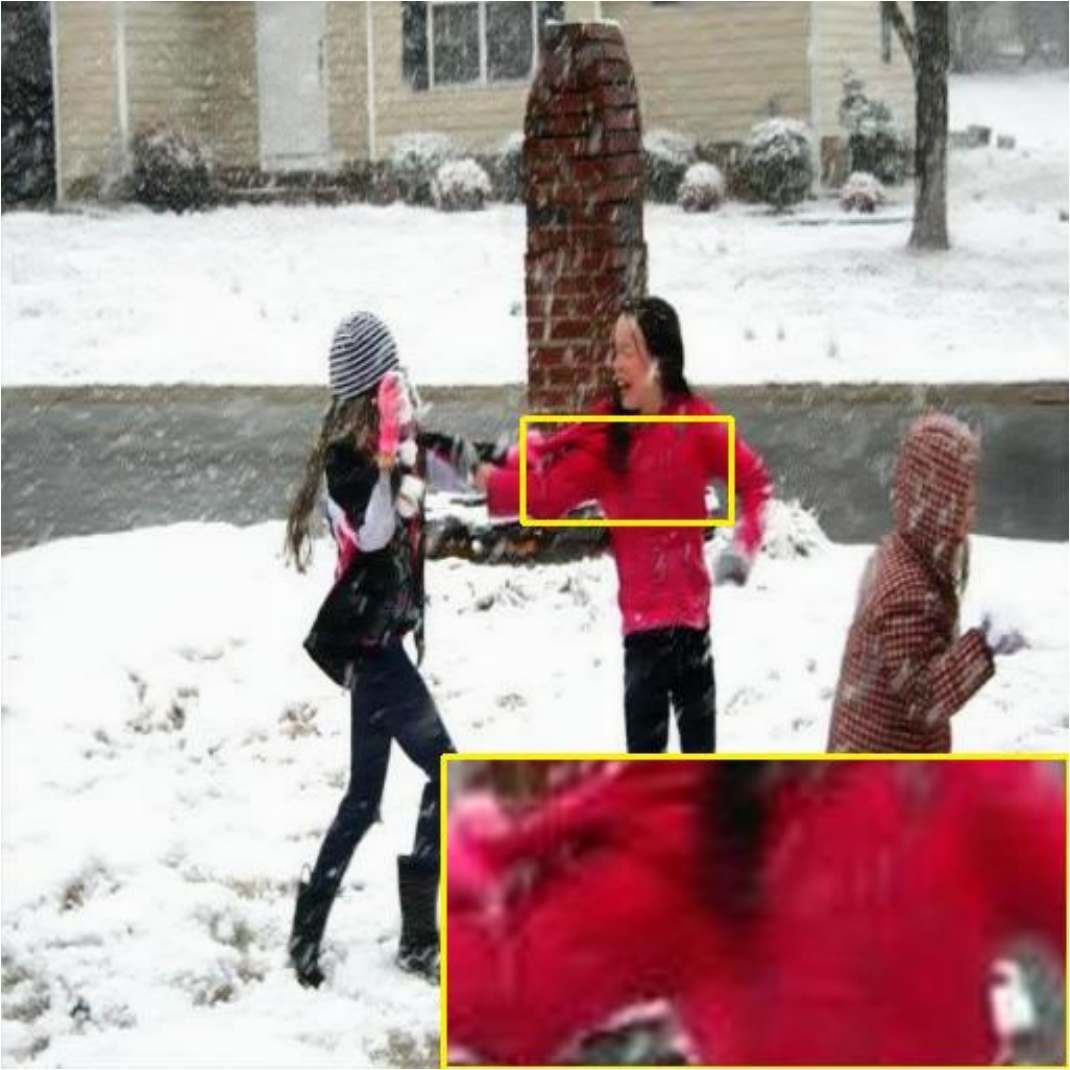}
		\vspace{-6mm}
		\subcaption*{(d) DID-MDN~\cite{zhang2018density}}
	\end{subfigure}
    \begin{subfigure}[t]{0.16\textwidth}
		\includegraphics[width=1\textwidth]{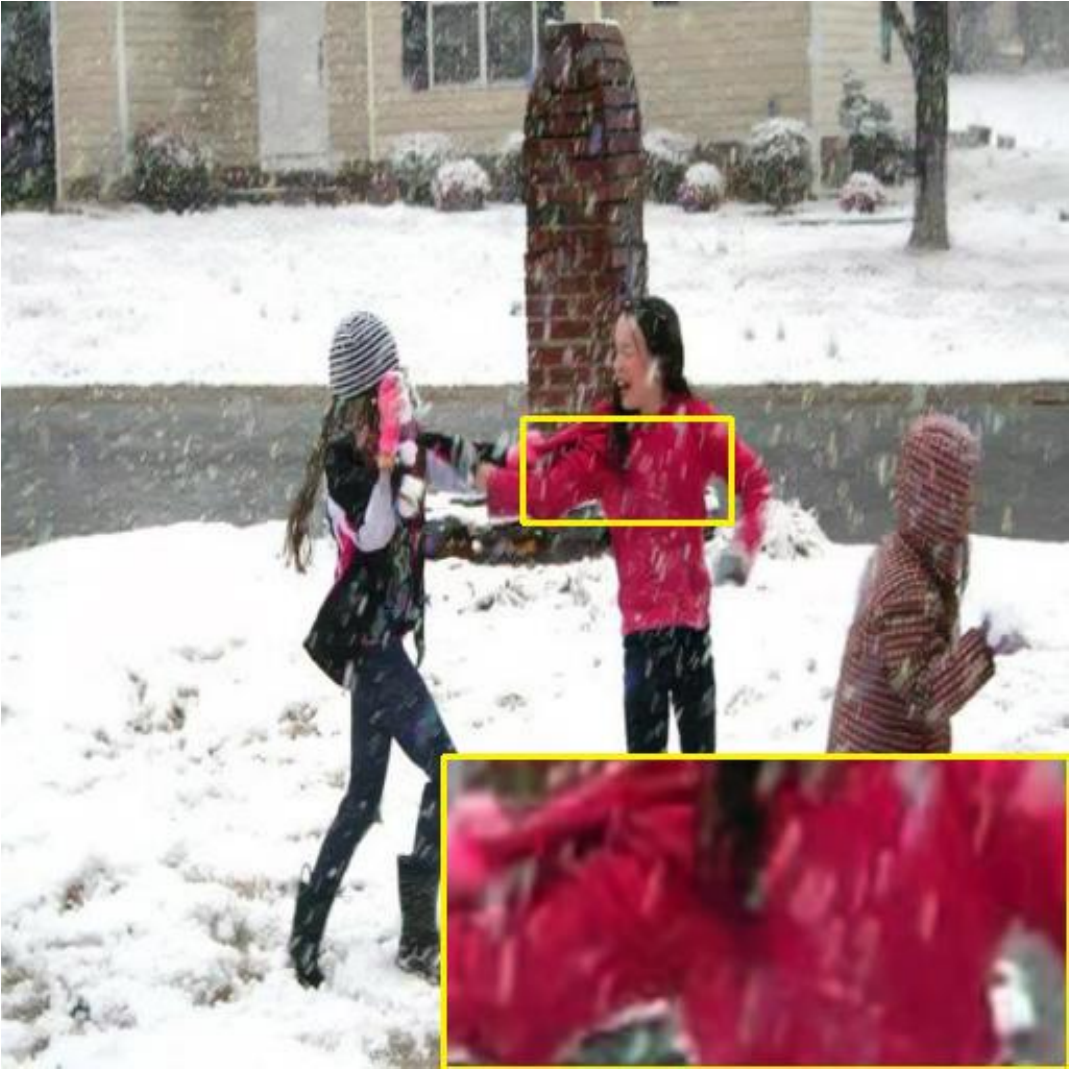}
		\vspace{-6mm}
		\subcaption*{(e) RESCAN~\cite{li2018recurrent}}
	\end{subfigure}	
	\begin{subfigure}[t]{0.16\textwidth}
		\includegraphics[width=1\textwidth]{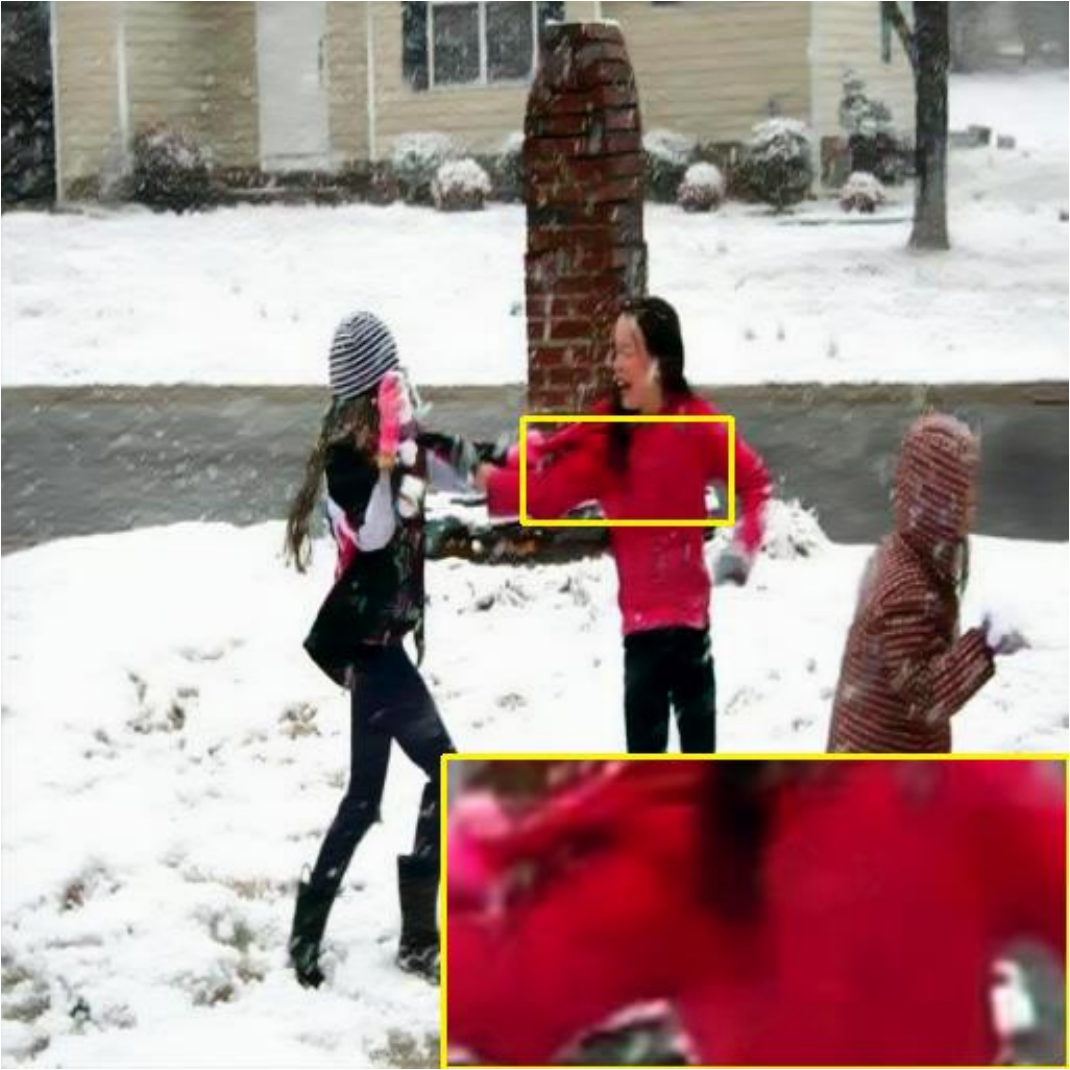}
		\vspace{-6mm}
		\subcaption*{(f) \textbf{CVID (Ours)}}
	\end{subfigure}	
 \end{center}
	\vspace{-4mm}
	\caption{\textbf{Derained images by different methods on five representative real-world rainy scenarios from datasets}~\cite{zhang2017convolutional}: light rain, medium rain, heavy rain, nighttime, dark background, and snow, respectively (from top to bottom).}
	\vspace{-1mm}
	\label{fig:real-rain}
\end{figure*}

\begin{table*}[t]
\normalsize
\begin{center}
\centering
\begin{tabular}{c||ccc|ccccc}
\Xhline{1pt}
\rowcolor[rgb]{ .85,  .9,  0.95}
Image Size & DDN~\cite{fu2017removing}  & JORDER~\cite{yang2017deep}   & RESCAN~\cite{li2018recurrent}  & CVID$_{n=1}$  & CVID$_{n=10}$ & CVID$_{n=100}$ \\
\hline
$500\times 500$ & 0.41 & 0.18 & 0.45 & 0.12 & 0.35 & 0.82 \\
$1024\times 1024$ & 0.76 & 0.82 & 1.81 & 0.42 & 0.68 & 1.38 \\
\hline
\end{tabular}
\end{center}
\vspace{-3mm}
\caption{\textbf{Running time (in seconds) of different methods} on rainy images with different sizes. }
\vspace{-1mm}
\label{table:time}
\end{table*}

\begin{figure*}[htbp]
\begin{center}
	\begin{subfigure}[t]{0.32\textwidth}
		\includegraphics[width=1\textwidth]{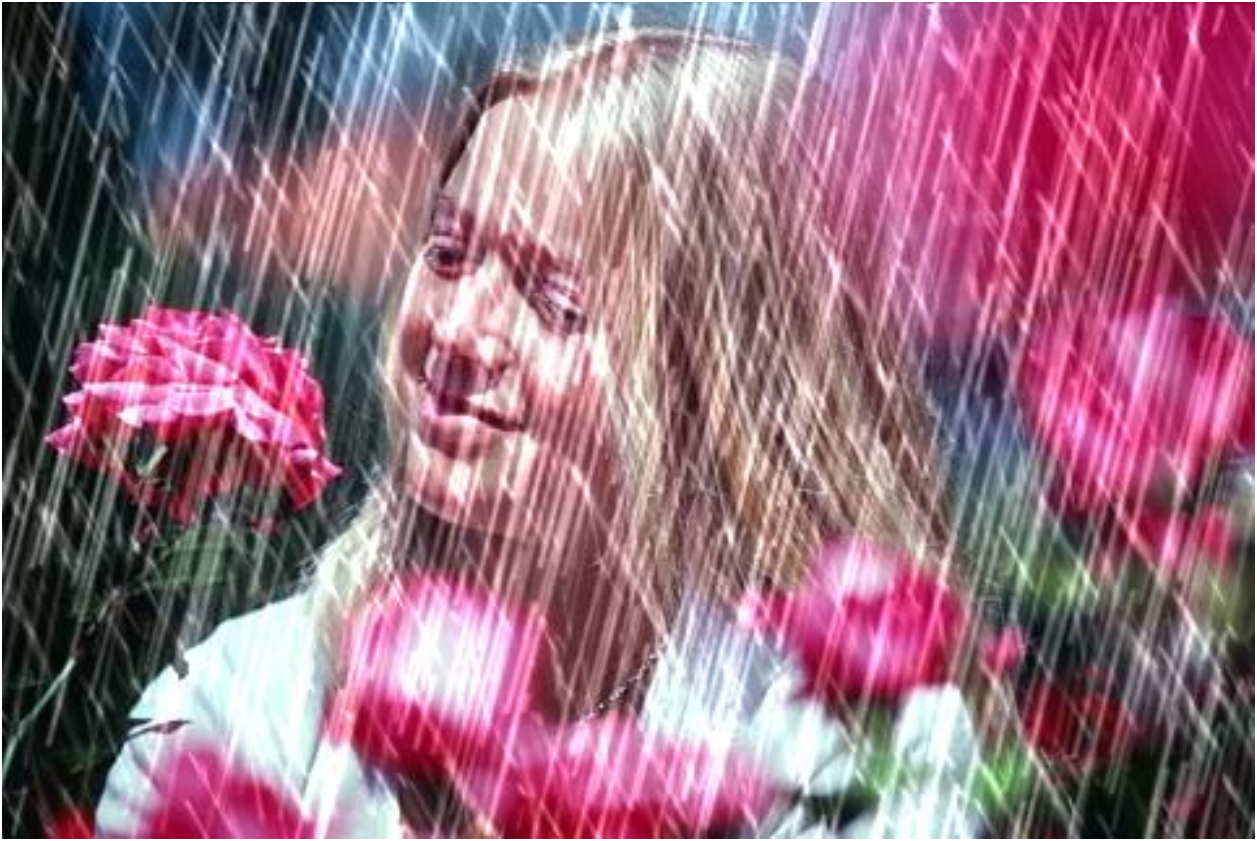}
		\vspace{-6mm}
		\subcaption*{\normalsize (a) Rainy Image: 26.83 dB/0.4525}
	\end{subfigure}	
	\begin{subfigure}[t]{0.32\textwidth}
		\includegraphics[width=1\textwidth]{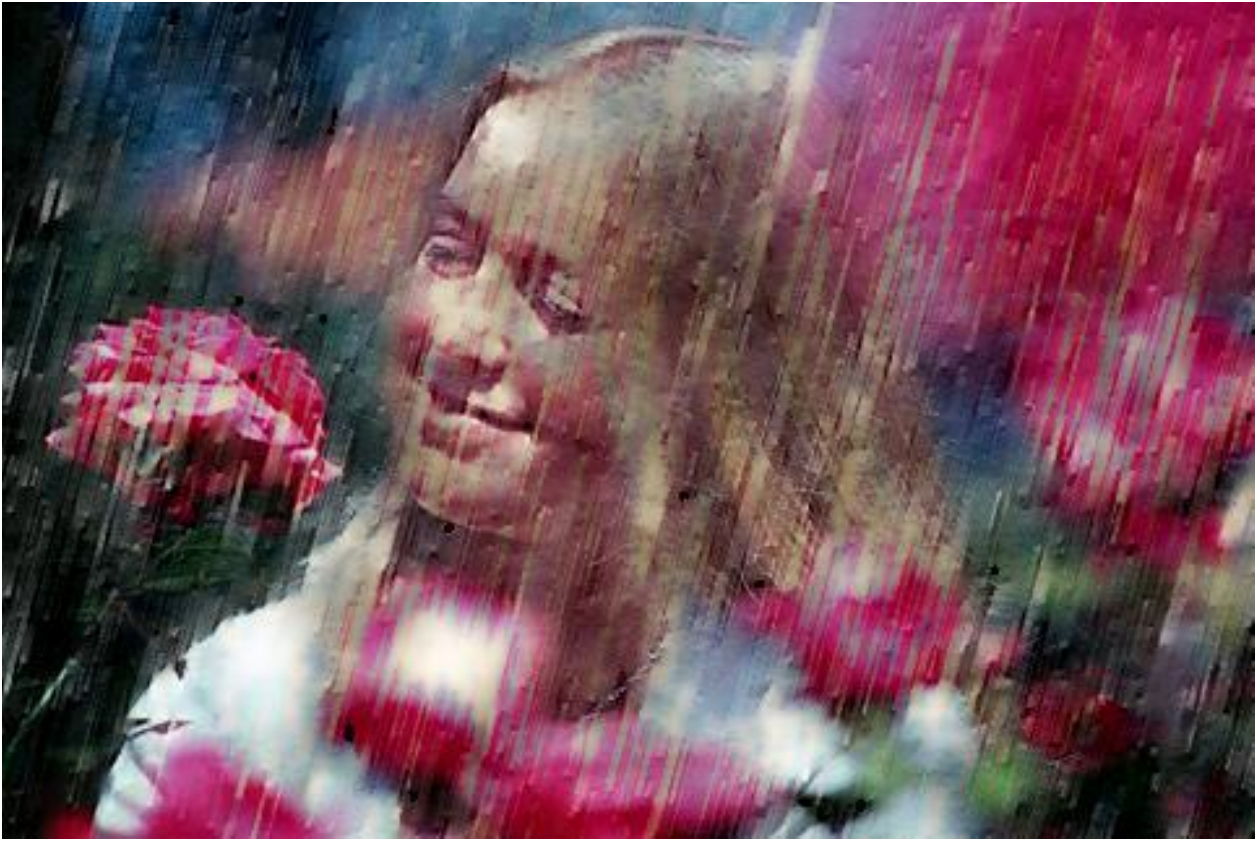}
		\vspace{-6mm}
		\subcaption*{\normalsize (b) DDN~\cite{fu2017removing}: 28.65 dB/0.4877}
	\end{subfigure}	
	\begin{subfigure}[t]{0.32\textwidth}
		\includegraphics[width=1\textwidth]{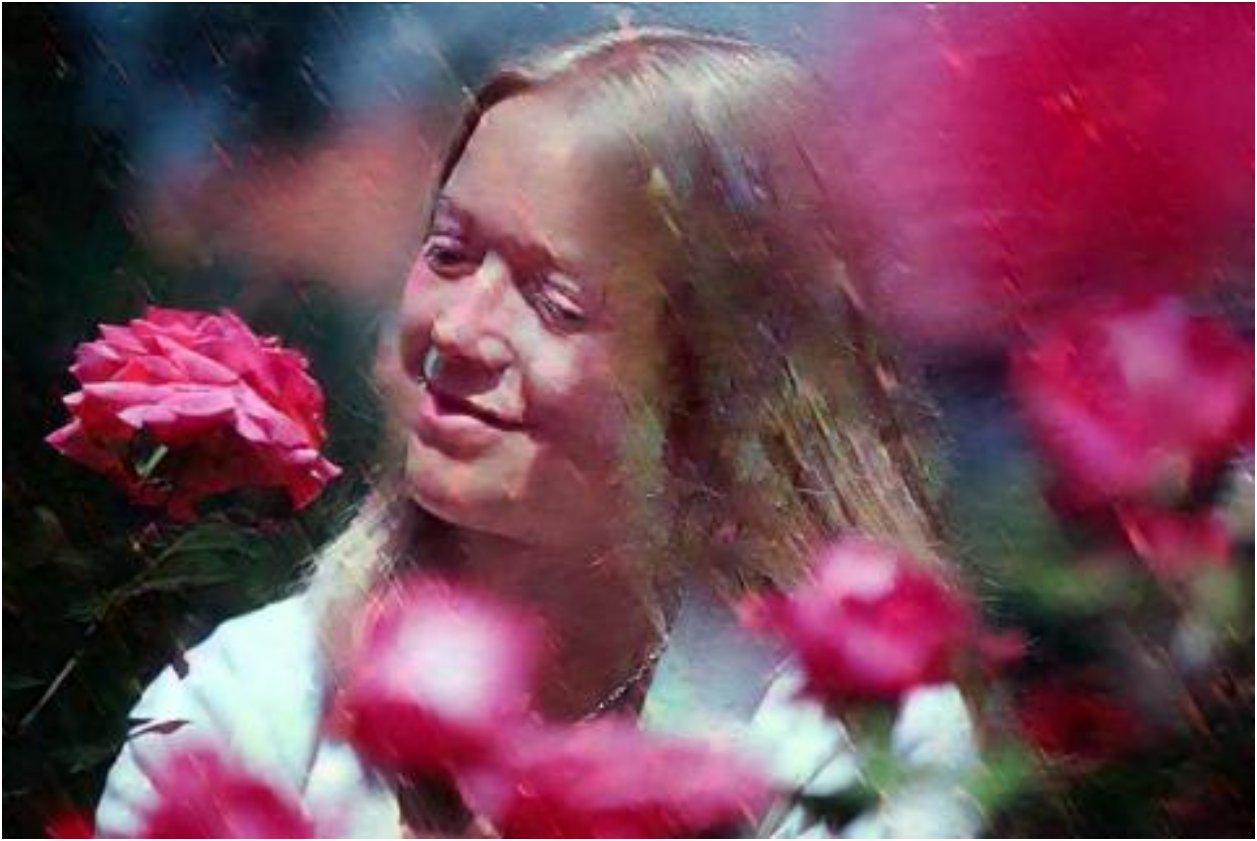}
		\vspace{-6mm}
		\subcaption*{\normalsize (c) CVAE: 30.89 dB/0.7849}
	\end{subfigure}	
	\begin{subfigure}[t]{0.32\textwidth}
		\includegraphics[width=1\textwidth]{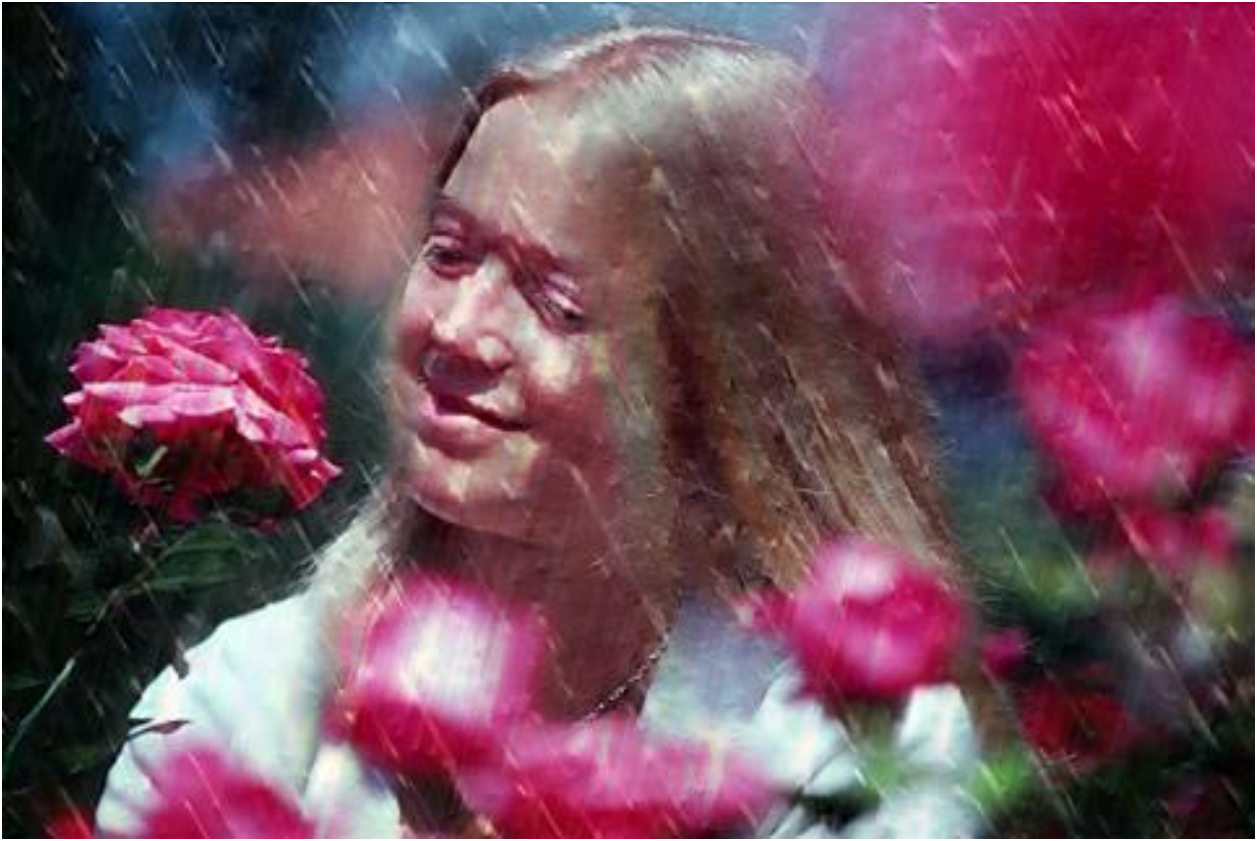}
		\vspace{-6mm}
		\subcaption*{\normalsize (d) DDN{\footnotesize +CW}: 29.66 dB/0.6798}
	\end{subfigure}
    \begin{subfigure}[t]{0.32\textwidth}
		\includegraphics[width=1\textwidth]{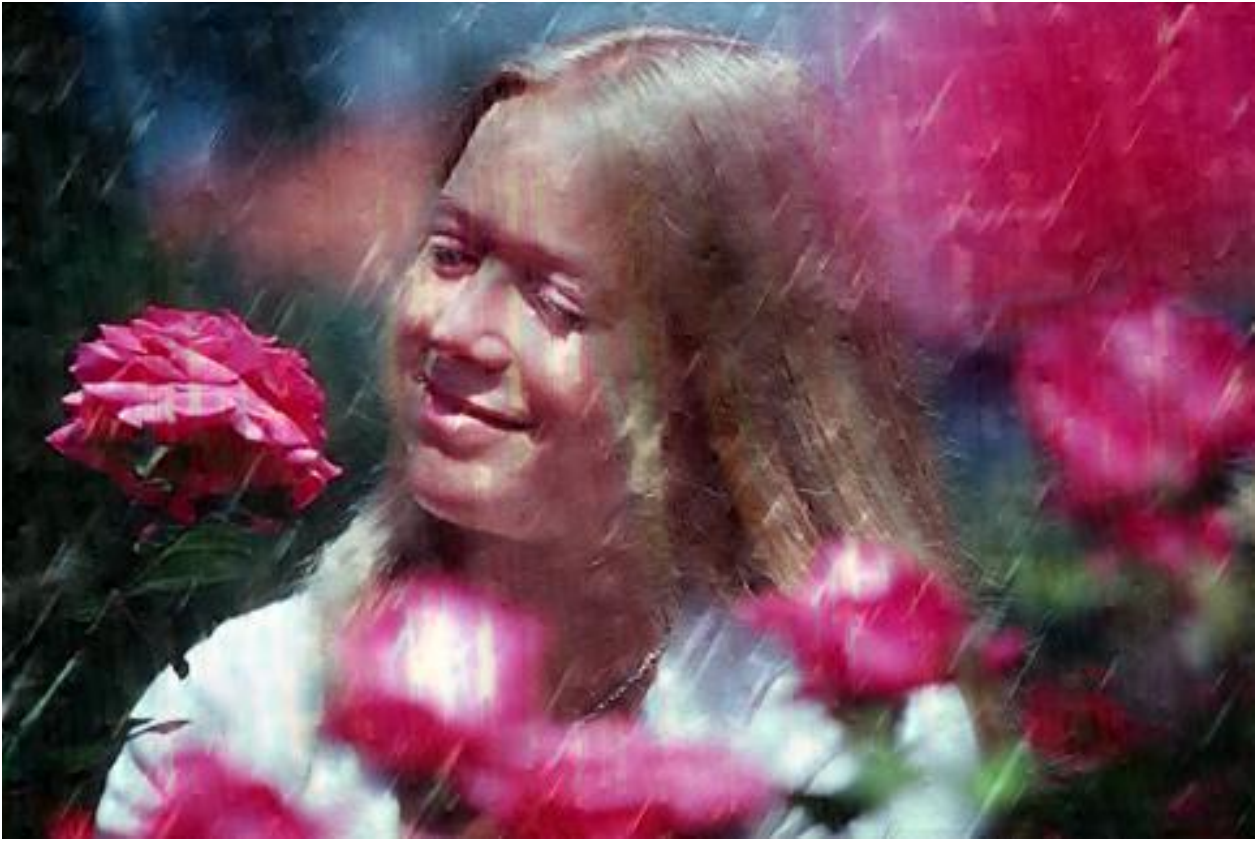}
		\vspace{-6mm}
		\subcaption*{\normalsize (e) DDN{\footnotesize +CW+SDE}: 29.76 dB/0.7008}
	\end{subfigure}	
	\begin{subfigure}[t]{0.32\textwidth}
		\includegraphics[width=1\textwidth]{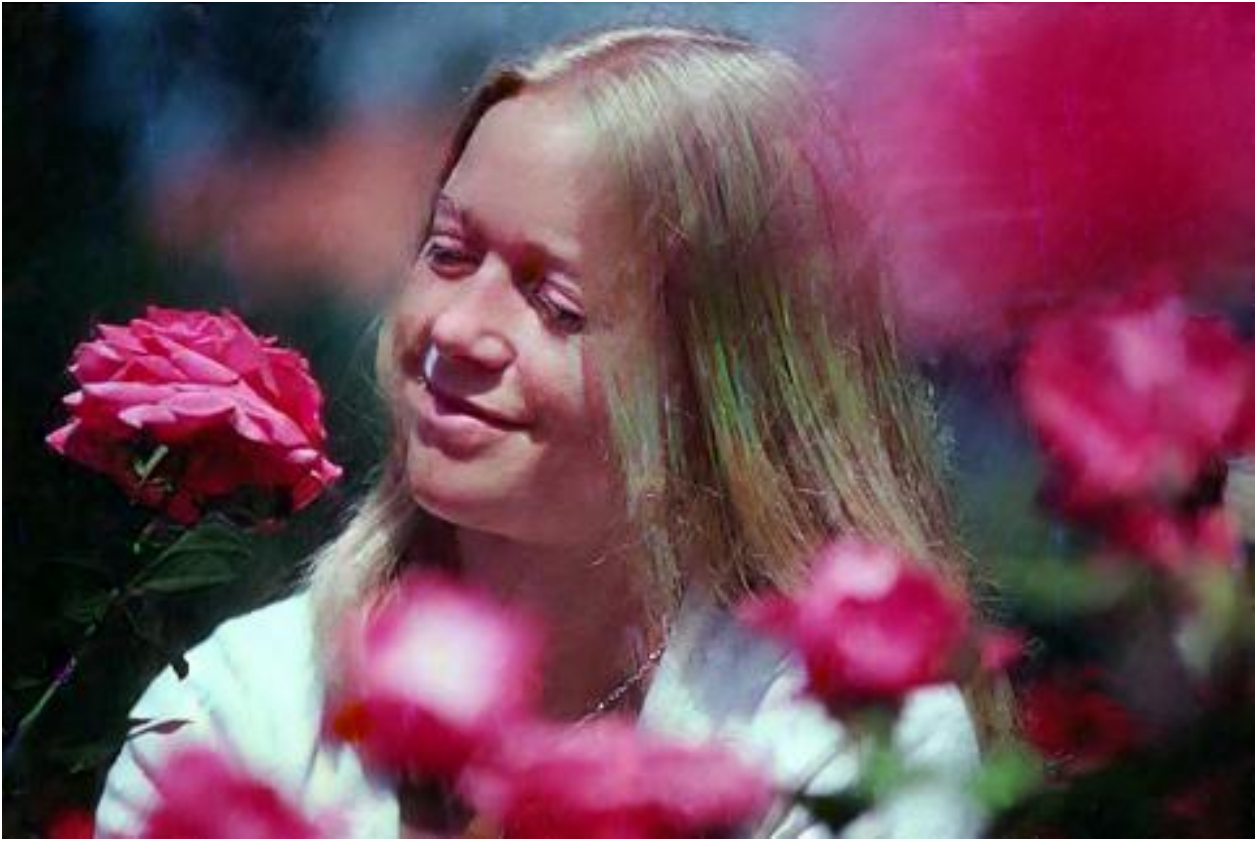}
		\vspace{-6mm}
		\subcaption*{\normalsize (f) CVAE{\footnotesize +CW}: 31.79 dB/0.8522}
	\end{subfigure}	
	\begin{subfigure}[t]{0.32\textwidth}
		\includegraphics[width=1\textwidth]{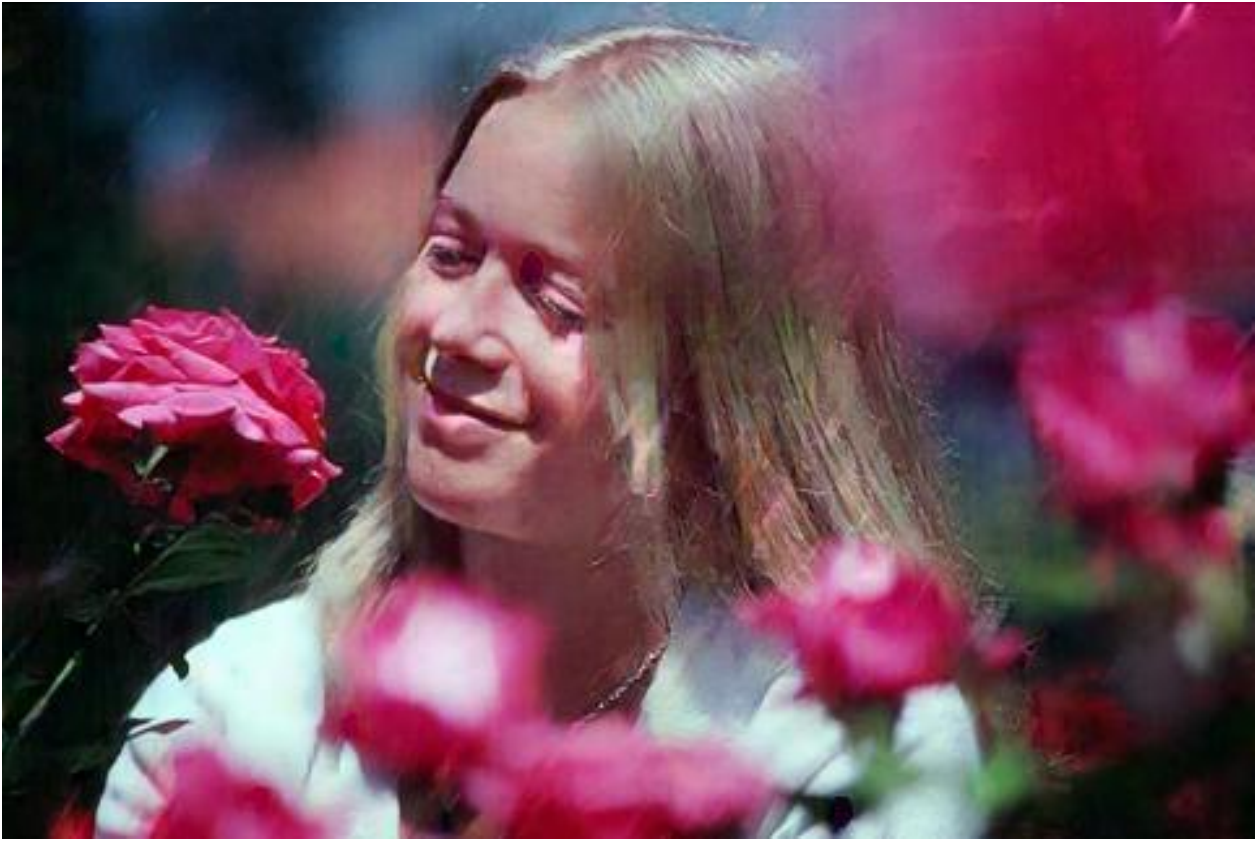}
		\vspace{-6mm}
		\subcaption*{\normalsize (g) CVAE{\footnotesize +SDE}: 32.25 dB/0.8758}
	\end{subfigure}	
	\begin{subfigure}[t]{0.32\textwidth}
		\includegraphics[width=1\textwidth]{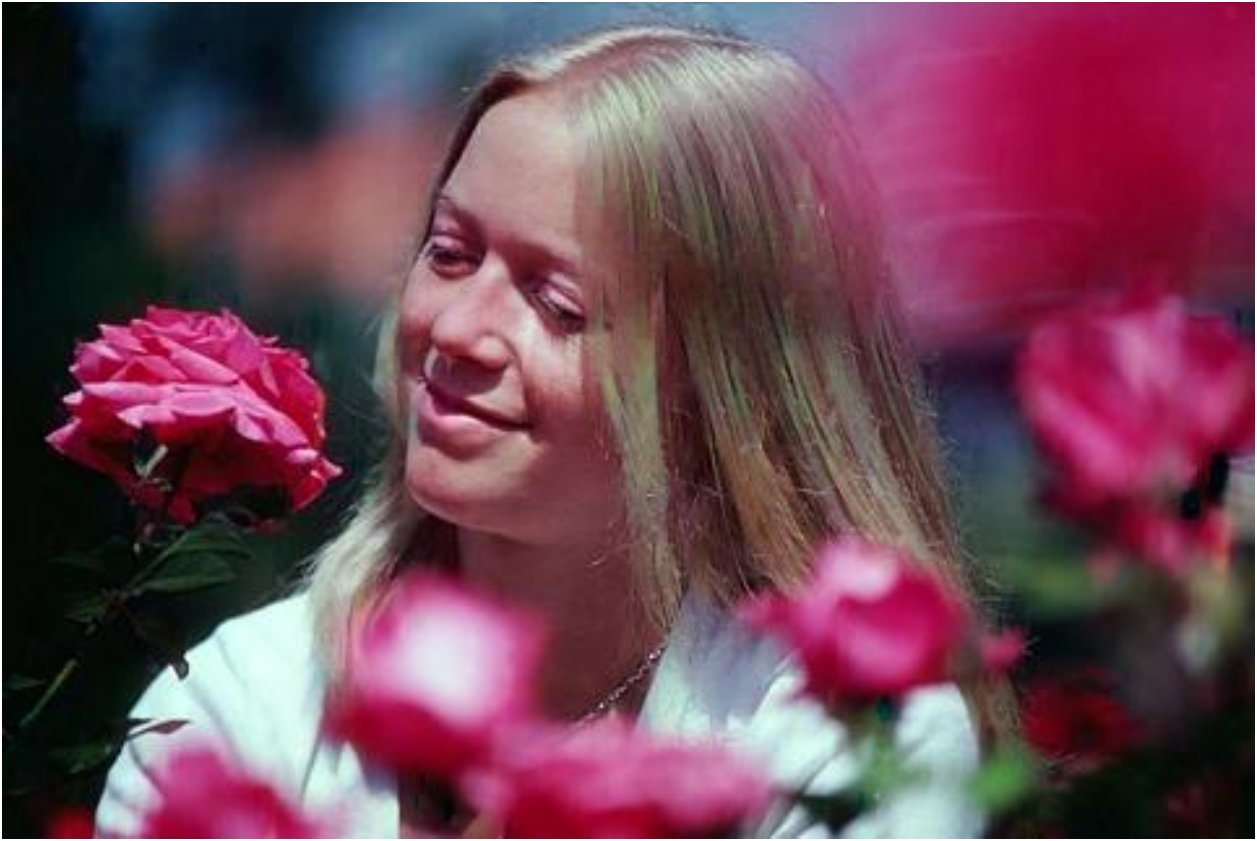}
		\vspace{-6mm}
		\subcaption*{\normalsize (h) \textbf{CVID}: \textbf{33.27} dB/\textbf{0.9073}}
	\end{subfigure}	
	\begin{subfigure}[t]{0.32\textwidth}
		\includegraphics[width=1\textwidth]{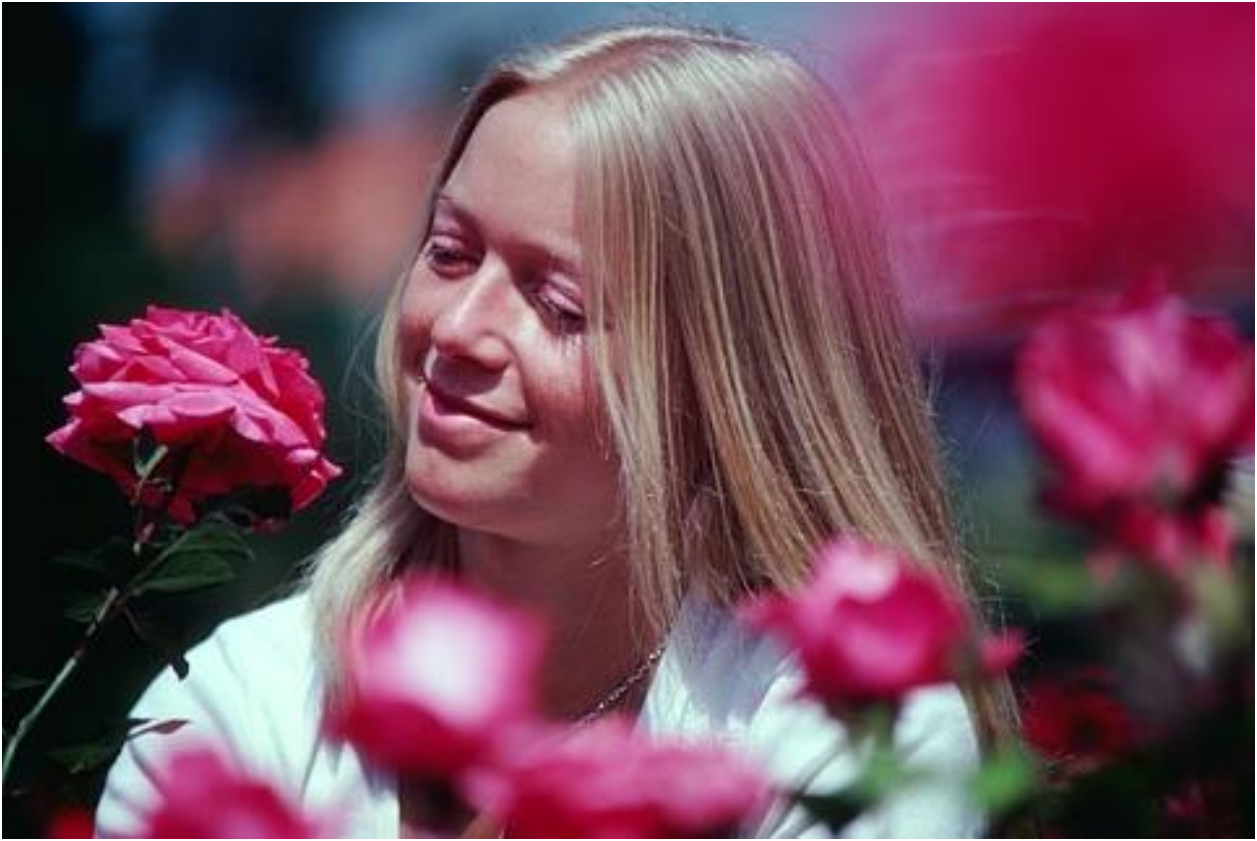}
		\vspace{-6mm}
		\subcaption*{\normalsize (i) Ground Truth}
	\end{subfigure}	
 \end{center}
	\vspace{-4mm}
	\caption{\textbf{Derained images and PSNR/SSIM results by different variants of DDN~\cite{fu2017removing} and the proposed CVAE backbone} on a synthetic rainy image from \textsl{D2} \textsl{Rain100H}~\cite{yang2017deeppami}.\, Our CVID network (CVAE{\footnotesize +SDE+CW}) is an integration of the CVAE backbone, the proposed spatial density estimation (SDE) module and channel-wise (CW) deraining scheme.}
	\label{fig:ablation}
	\vspace{-3mm}
\end{figure*}

\section{Experiments}
\label{sec:exp}
In this section, we conduct extensive experiments to demonstrate the effectiveness of the proposed Conditional Variational Image Deraining (CVID) network for image deraining.\, Comprehensive ablation studies are also performed to validate the effectiveness of different components.\, More results are provided in the \textsl{Supplementary File}.

\subsection{Implementation Details}
In our CVID network, we set $\beta=0.1$, $\lambda=1, n=100$.\, For network training, we randomly generate 2,000 pairs of image patches of size 64$\times$64 from each training set.\,
We use Adam optimizer~\cite{kingma2014adam} with default parameters, at a weight decay of $10^{-10}$ and a mini-batch size of $32$.\, The learning rate is initialized as $0.01$ and divided by $10$ at each epoch.\, The number of epochs is $4$.\,

\subsection{Experimental Protocol}
\noindent
\textbf{Datasets}.\, We perform experiments on 3 synthetic datasets and 1 real-world dataset.\, The first synthetic dataset is provided in~\cite{fu2017removing} and contains 14,000 synthesized clean/rainy image pairs.\, Following the settings in~\cite{zhang2018density}, 13,000 images are used for learning, and the remaining 1,000 images are used for testing (denoted as \textit{D1}).\, The second synthesized dataset is provided in~\cite{yang2017deeppami} and consists of 1,800 pairs of heavy rain images and 200 pairs of light rain images for learning.\, The two sets (\textit{Rain100L} and \textit{Rain100H}) are used for testing (denoted as \textit{D2}).\, The third synthetic dataset~\cite{zhang2018density} contains 12,000 synthesized clean/rainy image pairs, which includes 4,000 heavy rainy images, 4,000 medium rainy images, 4,000 light rainy images.\, The 1,200 pairs of clean/rainy images for testing are denoted as \textit{D3}.\, As far as we know, this is the first work that conducts experimental evaluation on all these three datasets.\, The real rainy images we tested are from the real-world dataset in~\cite{zhang2017convolutional}, which are downloaded from the internet by the authors.
\begin{table*}[ht]
\vspace{-1mm}
\begin{center}
\normalsize
	\begin{tabular}{c||c|c|c|c|c|c|c}
        \Xhline{1pt}
        \rowcolor[rgb]{ .85,  .9,  0.95}
        & DDN~\cite{fu2017removing} & CVAE & DDN{\footnotesize +CW} & DDN{\footnotesize +CW+SDE} & CVAE{\footnotesize +CW} &  CVAE{\footnotesize +SDE} & CVID \\
        \hline
        CVAE &\xmark &\cmark & \xmark  & \xmark & \cmark &\cmark & \cmark \\
        SDE &\xmark  &\xmark &\xmark   &\cmark  & \xmark &\cmark & \cmark \\
        CW & \xmark &\xmark & \cmark  & \cmark & \cmark &\xmark & \cmark \\
        \hline
         \rowcolor[rgb]{ .9,  .9,  .9}
        \textsl{D1}& 25.63/0.8851 & 26.57/0.8994 & 25.96/0.8901 & 26.08/0.8932& 27.38/0.9138 & 27.69/0.9193 & \textbf{28.96}/\textbf{0.9375}\\
        \rowcolor[rgb]{ .9,  .9,  .9}
        \textit{Rain100L}& 33.75/0.9213 & 35.38/0.9574 & 34.83/0.9493 & 35.78/0.9596& 36.03/0.9627 & 36.79/0.9783 & \textbf{37.83}/\textbf{0.9882}\\
        \rowcolor[rgb]{ .9,  .9,  .9}
        \textit{Rain100H}& 22.26/0.6928 & 25.25/0.7738  & 24.39/0.7637 & 26.11/0.8157 & 26.75/0.8332 & 27.04/0.8485 & \textbf{27.89}/\textbf{0.8721} \\
         \rowcolor[rgb]{ .9,  .9,  .9}
        \textsl{D3}& 27.33/0.8978 & 27.91/0.9073 & 27.55/0.9006 & 27.87/0.9065& 29.19/0.9198 & 29.38/0.9207 & \textbf{30.97}/\textbf{0.9374}\\
        \hline
        \end{tabular}
\end{center}
\vspace{-3mm}
\caption{\textbf{Quantitative comparisons of different variants} in terms of PSNR (dB) and SSIM~\cite{wang2004image} on dataset \textsl{D1}~\cite{fu2017removing}, \textsl{D2}~\cite{yang2017deeppami} and \textsl{D2}~\cite{yang2017deeppami}.\, CW means channel-wise scheme, SDE means spatial density estimation module.\, Our CVID is also ``CVAE{\footnotesize +CW+SDE}''.}
\label{ablation study}
\vspace{-3mm}
\end{table*}

\noindent
\textbf{Evaluation Metrics}.\,
We adopt three commonly-used metrics, i.e., peak signal to noise ratio (PSNR), structure similarity index (SSIM)~\cite{wang2004image}, and a perception-based metric NIQE~\cite{niqe}, to evaluate the performance of deraining on synthesized datasets.\, Since the real-world rainy images have no ``ground truth'' images, we only compare the visual quality of derained images by the competing methods.

\subsection{Comparison to the State-of-the-art}
\noindent
\textbf{Comparison Methods}.\, We compare the proposed CVID network with 6 state-of-the-art image deraining methods, including Gaussian Mixture Models (GMM)~\cite{li2016rain}, Deep Detail Network (DDN)~\cite{fu2017removing}, Joint Rain Detection and Removal (JORDER)~\cite{yang2017deeppami}, Image Deraining using conditional Generative Adversarial Network (ID-GAN)~\cite{zhang2017image}, Density-aware Deraining (DID-MDN)~\cite{zhang2018density}, and Recurrent Squeeze-and-excitation Context Aggregation Network (RESCAN)~\cite{li2018recurrent}.

\noindent
\textbf{Results on synthetic rain removal}.
The quantitative comparisons are reported in Table~\ref{ssim_1}.
Our CVID network substantially exceeds previous methods on all three datasets.
In particularly, on \textit{D1}, our method outperforms the second best method, i.e., ID-GAN~\cite{zhang2017image} by $2.42$ dB, $0.0233$, and $0.44$ in terms of PSNR, SSIM, and NIQE, respectively.
The superior performance demonstrates the great effectiveness of our method for single image deraining.
In Figs.~\ref{fig:syne1} and~\ref{fig:syne2}, we compare the derained images as well as PSNR/SSIM results by different methods.
We observe that our CVID removes rain streaks more clearly, while preserving image details better than previous methods.
%Other methods exhibit under deraining, leaving many rain streaks on certain regions of the image.

\noindent
\textbf{Results on realistic rain removal}.\,
It is a common challenge that the deraining methods learned on synthetic rainy images will suffer huge performance drop when processing real-world rainy images.\, However, this problem is largely alleviated by the exclusive generative property of the introduced CVAE framework, which can output multiple candidate solutions for one rainy input.\, To this end, we apply the proposed CVID network on removing the rain streaks in real rainy photographs.\, The proposed CVAE network is learned on the training set used in DID-MDN~\cite{zhang2018density}.\, We use the rainy images in~\cite{zhang2017convolutional}, including 4 different representative scenarios (shown in Figs.~\ref{fig:real-rain} (a)): light rain, medium rain, heavy rain, and snow (from top to bottom).\, As shown in Figs.~\ref{fig:real-rain} (b)-(f), our CVID outperforms previous competitors on diverse real scenarios.\, More results are provided in the \textsl{Supplementary File}.

\noindent
\textbf{Speed}.\, The comparison results on speed are listed in Table~\ref{table:time}.\, We observe that our CVID with $n=1$ generated sample is faster than other methods, while CVID with $n=100$ is slower than other methods on images of size $500\times500$, but still faster than RESCAN~\cite{li2018recurrent} on images of size $1024\times1024$.\,
\begin{table*}[t]
\begin{center}
\normalsize
%\footnotesize
\begin{tabular}{c||c|c|c|c|c|c|c}
\Xhline{1pt}
\rowcolor[rgb]{ .85,  .9,  0.95}
& CVID$_{n=1}$ &CVID$_{n=5}$ & CVID$_{n=10}$ &CVID$_{n=100}$ & CVID$_{n=200}$  & CVID$_{n=300}$  & CVID$_{n=500}$\\
\hline
\textit{Rain100L}&36.91  &37.19  &37.75  &37.83& \textbf{37.87} & 37.84 & 37.81 \\
\textit{Rain100H}&25.73  &26.18  &27.38 &27.89& \textbf{27.95} & 27.91 & 27.88  \\
\hline
\end{tabular}
\end{center}
\vspace{-3mm}
\caption{ \textbf{PSNR (dB) results our CVID network} on \textsl{D2}~\cite{yang2017deeppami} with sample number $n=1, 5, 10,100, 200, 300, 500$.
}
\label{table:samples}
\vspace{-0mm}
\end{table*}

\noindent
\subsection{Validation of the Proposed CVID Network}
\label{Ablation}

To further validate the working mechanism of our CVID network, we conduct deeper analysis on rainy image dataset \textsl{D2}~\cite{yang2017deeppami}.
Specifically, we assess 1) the importance of the employed CVAE framework;
2) the effect of the proposed channel-wise deraining scheme;
3) the influence of the proposed SDE module to CVID network;
4) how does the sample number $n$ influence the performance of our CVID network;
5) the influence of the hyper-parameters $\beta$ and $\lambda$;
6) the impact of these components in our CVID network on cumulative error distribution (CED);
and 7) the choice of ReLU or PReLU in our CVID. 
In all experiments, DDN~\cite{fu2017removing} is employed as the baseline due to its simple network architecture.

\noindent
\textbf{1)\ How important is the employed CVAE framework for image deraining?}\ To evaluate the importance of our employed CVAE backbone network on image deraining, we compare the baseline DDN network~\cite{fu2017removing} and our CVAE backbone.\, As shown in Table~\ref{ablation study} (the 2nd and 3rd columns), the results of CVAE are 35.38 dB/0.9574 on PSNR/SSIM, much higher than those (33.75 dB/0.9213) of the baseline DDN~\cite{fu2017removing}.\, What's more, when comparing the channel-wise variants of the DDN and CVAE, we observed that the channel-wise CVAE achieves PSNR/SSIM results of 36.03 dB/0.9627, still much better than those of channel-wise DDN (34.83 dB/0.9493).\, On visual quality, from Figs.~\ref{fig:ablation} (b) and (c), we observe that our CVAE backbone achieves more clear results than DDN~\cite{fu2017removing}.\, All these results demonstrate the importance of the employed CVAE backbone for image deraining over the ResNet backbone~\cite{he2016deep}.

\noindent
\textbf{2)\ How is the effect of the proposed channel-wise (CW) deraining scheme?}\ To study the effect of the proposed CW scheme, we embed it into both the DDN network~\cite{fu2017removing} and our CVAE backbone, we call them DDN{\footnotesize +CW} and CVAE{\footnotesize +CW}, respectively.\, From Table~\ref{ablation study} (the 4th and 6th columns), we observed that the results of baseline DDN on PSNR/SSIM are dramatically improved from 33.75 dB/0.9213 to 34.83 dB/0.9493 (DDN{\footnotesize +CW}) by our proposed CW deraining scheme.\, Similarly, with CW, the results of our CVAE backbone on PSNR/SSIM are also improved from 35.38 dB/0.9574 to 36.03 dB/0.9627 (CVAE{\footnotesize +CW}).\, Comparing Figs.~\ref{fig:ablation} (b) and (d), (c) and (f), we observe that the DDN{\footnotesize +CW} and CVAE{\footnotesize +CW} achieve much better visual quality than the baseline DDN~\cite{fu2017removing} and our CVAE backbone, respectively.\, All these results clearly demonstrate the effectiveness of our CW deraining scheme.

\noindent
\textbf{3)\ How does the proposed SDE module influence our CVID network?}\ To answer this question, we embed the proposed SDE module into the variant DDN{\footnotesize +CW}, the proposed CVAE backbone, and the variant CVAE{\footnotesize +CW}, the resulting variants are called DDN{\footnotesize +CW+SDE}, CVAE{\footnotesize +SDE}, and CVAE{\footnotesize +CW+SDE} (CVID).\, The corresponding results of these variants on dataset \textsl{D2} (\textsl{Rain100L} and \textsl{Rain100H})~\cite{yang2017deeppami} are listed in Table~\ref{ablation study} (the 5th, 7th, and 8th columns).\, It can be seen that all these variants armed with our SDE module performs significantly better than the corresponding baselines.\, For example, DDN{\footnotesize +CW+SDE} achieves 35.78 dB/0.9596 on PSNR/SSIM, 0.95 dB/0.0103 higher than those of DDN{\footnotesize +CW}.\, On visual quality, from Figs.~\ref{fig:ablation} (c) and (g), (d) and (e), (f) and (h), we observe that our our SDE module consistently improves the performance of several variants.\, 
These improvements validate that the proposed SDE module indeed boost our baseline to a great extent.

\begin{figure*}
    \centering
		\begin{subfigure}[t]{0.48\textwidth}
 \includegraphics[width=1\textwidth]{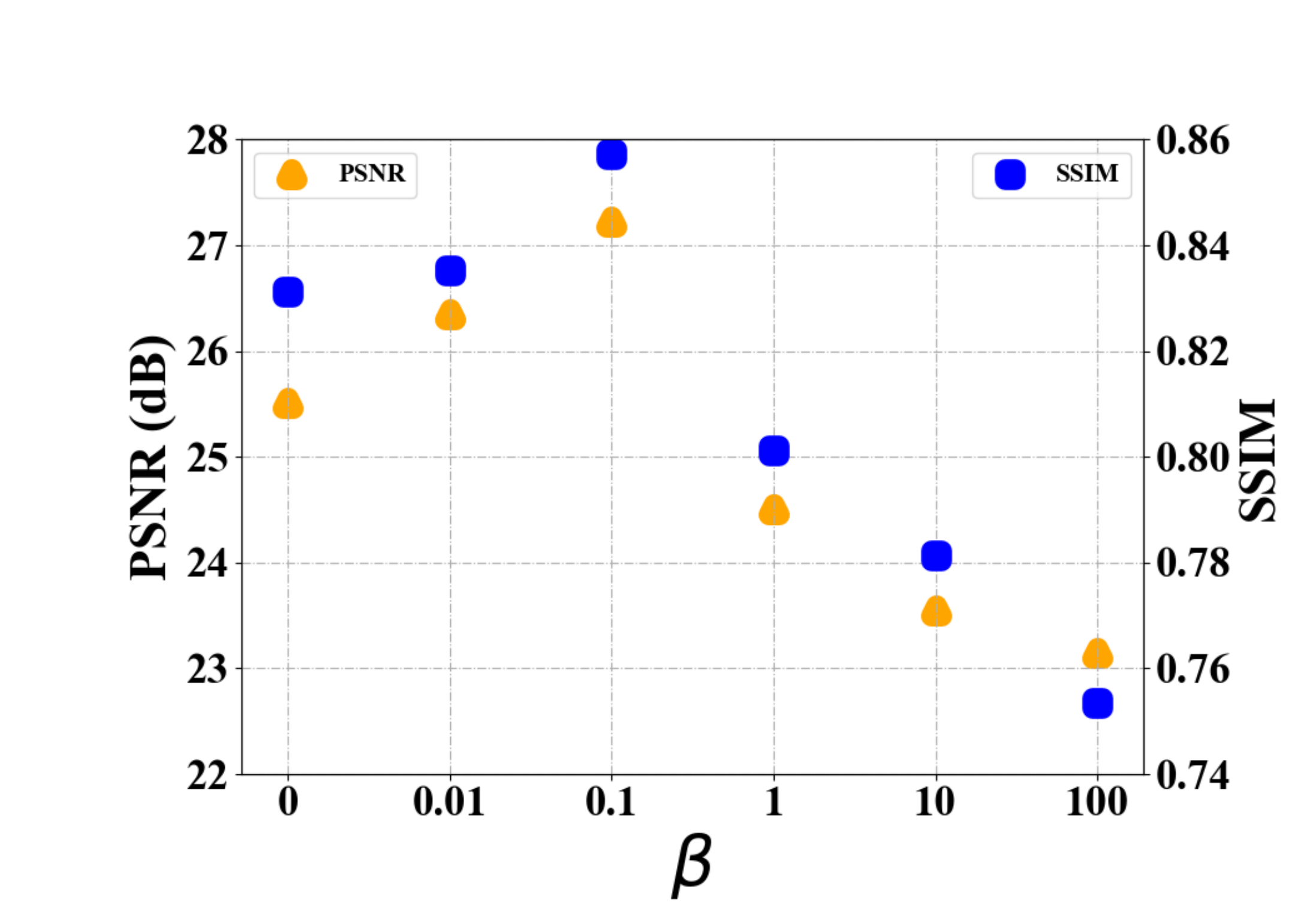}
	\end{subfigure}	
		\begin{subfigure}[t]{0.48\textwidth}
		    \includegraphics[width=1\textwidth]{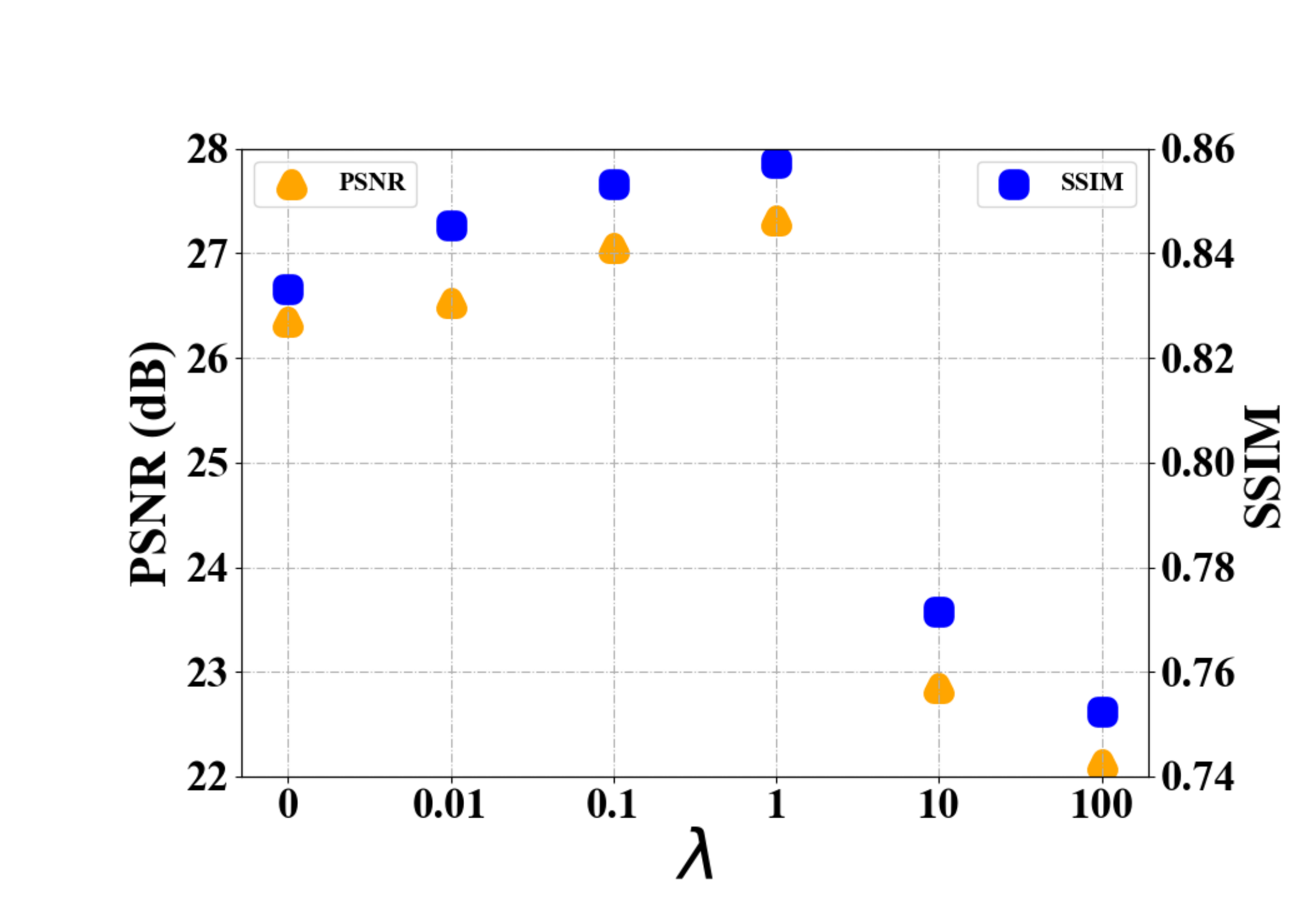}
	\end{subfigure}
	\vspace{-2mm}
    \caption{\textbf{PSNR (dB) and SSIM results of our CVID network} with different $\beta$ (left) and $\lambda$ (right) on \textsl{D2} \textsl{Rain100H}~\cite{yang2017deeppami}.}
    \label{fig:parameter}
\end{figure*}

\begin{table}[t]
\begin{center}
%\normalsize
\footnotesize
\vspace{-1mm}
	\begin{tabular}{c|| c | c }
        \Xhline{1pt}
        \rowcolor[rgb]{ .85,  .9,  0.95}
        & CVID with ReLU & CVID with Leaky ReLU \\
        \hline
        \textit{Rain100L} & 37.13/0.9817 & \textbf{37.83}/\textbf{0.9882} \\
        \textit{Rain100H} & 26.97/0.8597 & \textbf{27.89}/\textbf{0.8721} \\
        \hline
    \end{tabular}
\end{center}
\vspace{-3mm}
\caption{\textbf{PSNR (dB) and SSIM~\cite{wang2004image} results of our CVID with different activation functions} on \textsl{D2} dataset~\cite{yang2017deeppami}.}
\label{table:relu}
\vspace{-3mm}
\end{table}

\begin{figure*}[t]
\begin{center}
  \begin{subfigure}[t]{0.325\textwidth}
		\includegraphics[width=1\textwidth]{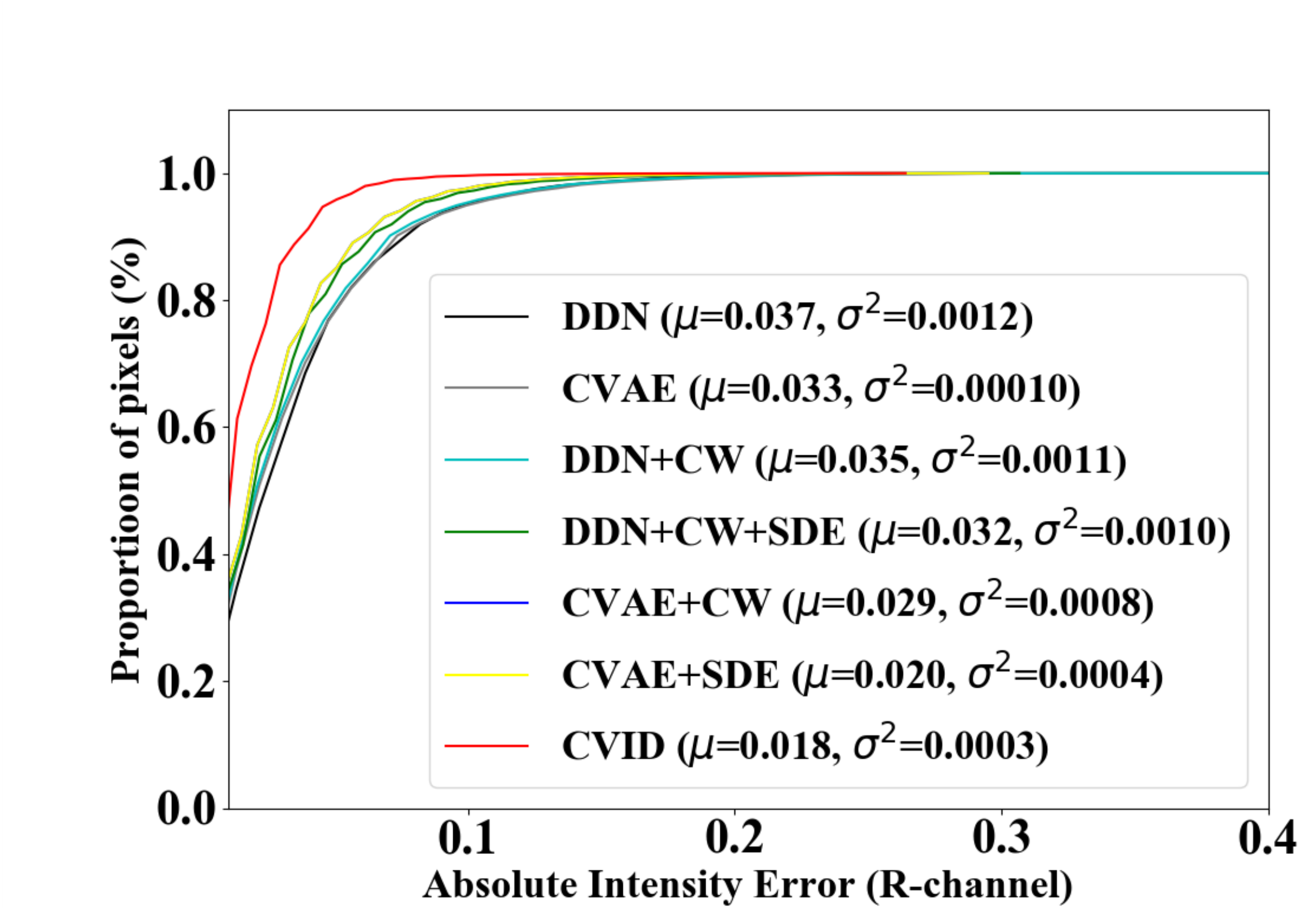}
	\end{subfigure}	
	\begin{subfigure}[t]{0.325\textwidth}
		\includegraphics[width=1\textwidth]{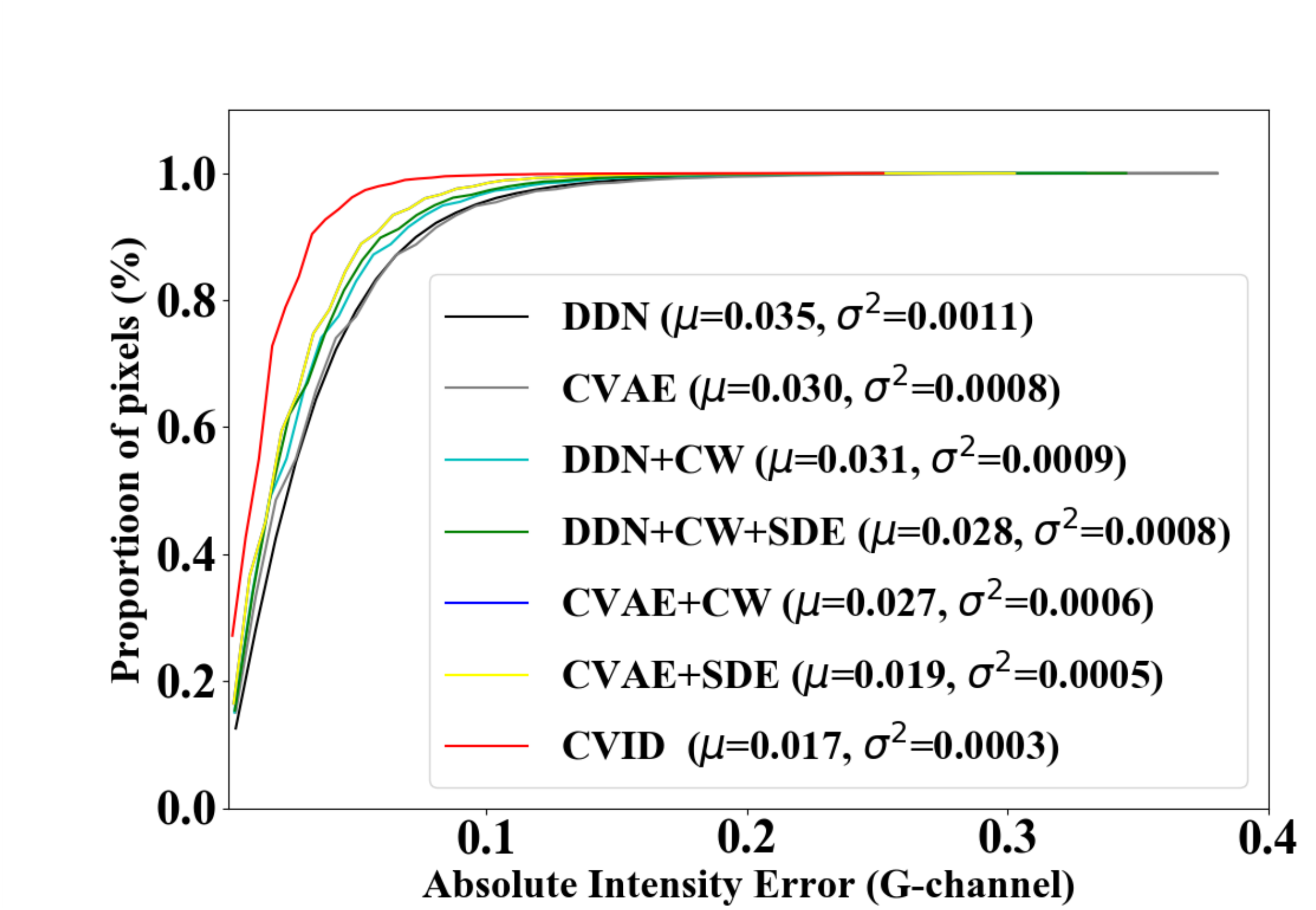}
	\end{subfigure}	
	\begin{subfigure}[t]{0.325\textwidth}
		\includegraphics[width=1\textwidth]{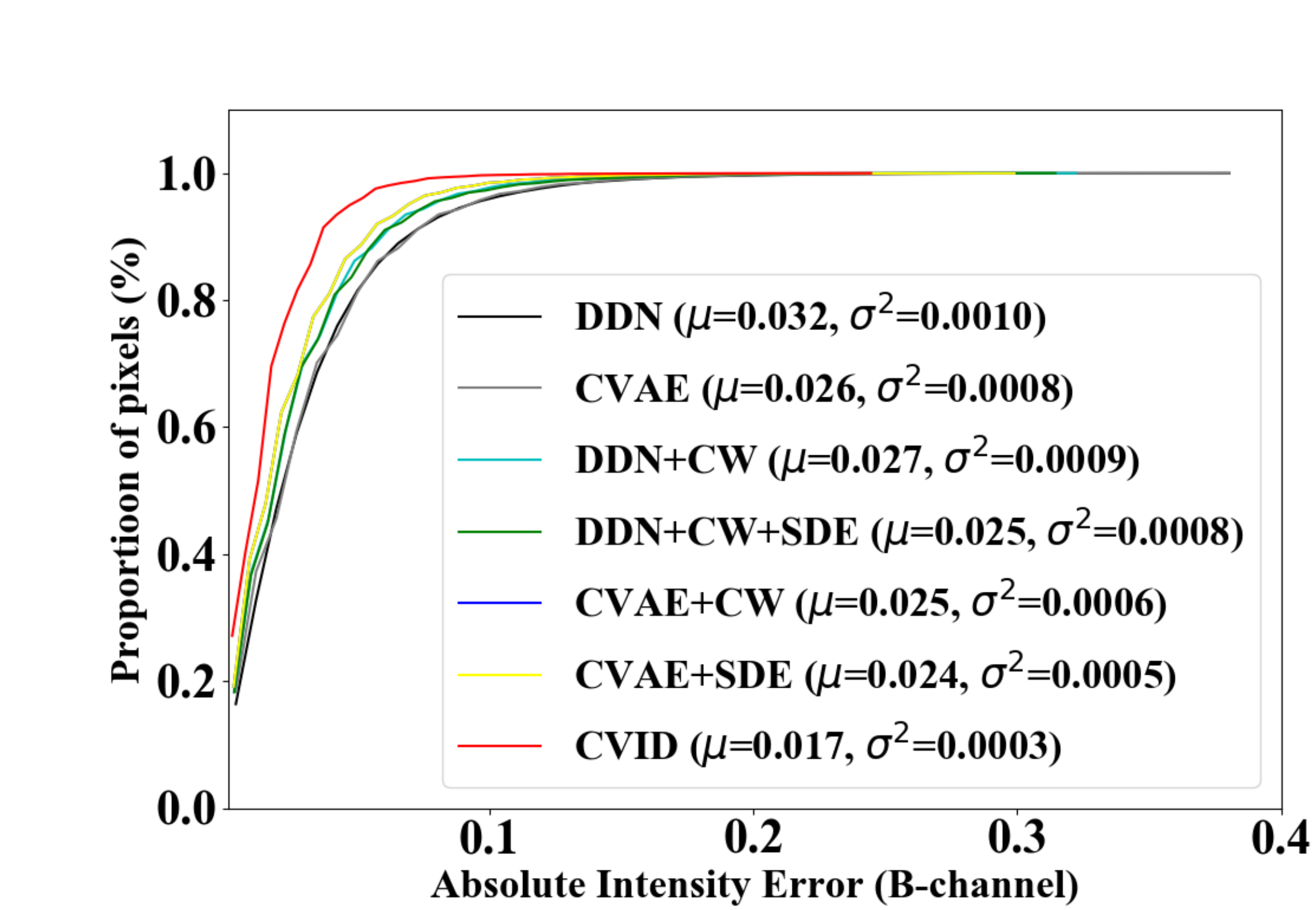}
	\end{subfigure}	
	\caption{\textbf{Illustration of cumulative error distribution (CED) curves} in terms of absolute intensity errors, along with their mean ($\mu$) and variance ($\sigma^2$) on R (left), G (middle), and B (right) channels, respectively.}
	\label{fig:CED}
\end{center}
\vspace{-5mm}
\end{figure*}

\noindent
\textbf{4) How does our CVID network perform with larger sample number $n$}?
To study this point, we run our CVID network with different sample number $n$ on the \textsl{Rain100L} and \textsl{Rain100H} datasets in \textsl{D2}~\cite{yang2017deeppami}.
The results on PSNR (dB) are listed in Table~\ref{table:samples}.
We observe that our CVID: 1) achieves inferior results to RESCAN~\cite{li2018recurrent} with $n=1$, which achieves 37.27 dB/0.9813 and 26.45 dB/0.8458 on the \textsl{Rain100L} and \textsl{Rain100H} datasets, respectively;
2) performs consistently better with $n=5, 10, 100, 200$ generated images, but converges at $n=200$ and becomes worse when $n=300, 500$;
3) is faster than RESCAN with $n=1$, but requires more running time with more sample number, as shown in Table~\ref{table:time}.
This again demonstrates that our CVAE framework is more flexible on the accuracy-speed trade-off over the deterministic framework employed by the comparison methods.

\noindent
\textbf{5)\ How does the parameters $\beta$ and $\lambda$ influence the performance of our CVID network?}\ Our CVID has two hyper-parameters $\beta$ and $\lambda$ balancing the importance of the KL divergence loss in Eqn.\, (\ref{cvae_loss}) and SDE loss in Eqn.\, (\ref{sde_loss}), respectively.\, In our CVID network, we set $\beta=0.1,\lambda=1$.\, To study their influence to our CVID, we perform experiments on \textsl{D2} \textsl{Rain100H} dataset~\cite{yang2017deeppami}.\, We change one parameter at a time, while fixing the other.\, The results are plotted in Fig.\, \ref{fig:parameter}.\, We observe that our CVID performs better when $\beta$ is increased from 0 to 0.1, but worse when $\beta=1,10,100$.\, Similar trends can be found on the influence of parameter $\lambda$ to our CVID.\,

\noindent
\textbf{6)\ Performance of our CVID network on cumulative error distribution (CED)}.\, To further validate the effectiveness of our CVID network, we employ CED as a supplementary way of PSNR and SSIM for performance measurement.\, Here, CED calculates the cumulative histogram curve of the absolute intensity errors between two images.\, Earlier saturating curve indicates more accurate approximation of the two images.\, In Fig.~\ref{fig:CED}, we plot the CED curves in terms of absolute pixel intensity errors between derained images and the ground truth, by different variants of the baseline DDN~\cite{fu2017removing} and our CVAE backbone.\, We observe that the curve of our CVID network saturates much earlier than those variants of DDN~\cite{fu2017removing} and our CVAE backbone.\, In addition, the curve of our CVID network achieves the lowest mean and variance of the errors among all competing variants, again demonstrating the effectiveness of our CVID network for single image deraining.

\noindent
\textbf{7) The choice of ReLU or Leaky ReLU in our CVID}.
We employ Leaky ReLU~\cite{xu2015empirical} in our CVID network, since Leaky ReLu is usually better than vanilla ReLU~\cite{maas2013rectifier} on nonlinear activation. 
The comparison of our CVID with ReLU and Leaky ReLU operation is listed in Table~\ref{table:relu}. 
We observe that the results also support the choice of Leaky ReLU in our CVID network for the image deraining task.

\section{Conclusion}
\label{sec:con}
In this paper, we proposed a Conditional Variational Image Deraining (CVID) network to tackle the image deraining problem.\, CVID leverages the powerful generative ability of Conditional Variational Auto-Encoder (CVAE) framework on modeling the latent distributions of clean image priors, from which multiple derained images are generated for image deraining.\, Moreover, we proposed a spatial density estimation module and a channel-wise deraining scheme to achieve more adaptive image deraining in different color channels.\, A spatial density estimation module is developed to achieve spatially adaptive deraining performance on uneven rainy images.\, Experiments on both synthetic and real-world datasets show that our CVID network achieves consistently better performance than previous state-of-the-art image deraining methods.

%\vspace{1mm}
%\noindent
%\textbf{Acknowledgements}.\,
% \noindent
% \textbf{Acknowledgements}.\,
% This work was supported by Natural Science Foundation of China (No.\, 61976060, 61871016, 61572264, 61620106008, 61802324, 61772443), the National Youth Talent Support Program, and Tianjin Natural Science Foundation (17JCJQJC43700, 18ZXZNGX00110).

\section{Appendix}
\label{appendix}

\noindent
\textbf{Proof of Proposition 1}:\
Denote $\bar{B}$ ($B$) and $\bar{R}$ ($R$) as derained image and removed rain streak layer without (with) distinguishing different color channels, respectively.\, We have
\begin{equation}
O_c = B_c + R_c
,\ 
O_c = \bar{B}_c + \bar{R}_c,
\label{all}
\end{equation}
where $O$, $B$, and $R$ denote the rainy image, the clean background image and the rain streak layer, and $c\in \{R, G, B\}$ is the color channel index.\, From (\ref{all}), we have:
\begin{equation}
\label{o_b}
\begin{aligned}
\bar{B}_c - B_c= R_c - \bar{R}_c
\end{aligned}
\end{equation}

In previous deraining models that do not distinguish color channels, the gray rain streaks $R$ are simply added onto the clean RGB image, which results in the same density distribution of rain streaks for the three color channels.\, However in our model, we treat the three color channels separately by specifying the rain streaks $R_c$ for each color channel, since rain streaks are distinctively distributed in three color channels.\, Therefore, it holds that $R_c(y) \leq \bar{R}_c(y)$.\, Then we can get $B_c \geq \bar{B}_c$, and naturally have $\underset{y \in \Omega (x)} \max (B_c(y))\geq\underset{y \in \Omega (x)} \max (\bar{B}_c(y))$, which gives rise to
\begin{equation}\label{sub_result}
  \underset{y \in \Omega (x)}{\max}(\underset{c \in \{R, G, B\}}{\max} B_{c}(y)) \geq \underset{y \in \Omega (x)}{\max}(\underset{c \in \{R, G, B\}}{\max} \bar{B}_{c}(y))
\end{equation}
\begin{equation}
\label{J_B}
J^{bright}(B_c)(x) \geq J^{bright}(\bar{B}_c)(x),
\end{equation}
this indicates that the bright channel for an image derained without channel distinction will have a lower intensity than the one that is channel-wisely derained.

The conclusion is, the number of brightest pixels in an image derained channel-wisely is larger than that of an image without distinguished color channels, i.e.,
\begin{equation}
\|1- B_c\|\ \|1- \bar{B}_c\|_0
\end{equation}
and
\begin{equation} \label{end_corollary}
\|1- B\|_0 \leq \|1- \bar{B}\|_0,
\end{equation}
which ends the proof.

{
\bibliographystyle{ieee}
\bibliography{cvid}

\begin{thebibliography}{10}\itemsep=-1pt

\bibitem{bao2017cvae}
J.~Bao, D.~Chen, F.~Wen, H.~Li, and G.~Hua.
\newblock {CVAE-GAN}: fine-grained image generation through asymmetric
  training.
\newblock In {\em Proceedings of the IEEE International Conference on Computer
  Vision (ICCV)}, pages 2745--2754, 2017.

\bibitem{barnum2010analysis}
P.~C. Barnum, S.~Narasimhan, and T.~Kanade.
\newblock Analysis of rain and snow in frequency space.
\newblock {\em International Journal of Computer Vision}, 86(2):256, Jan 2010.

\bibitem{chang2017transformed}
Y.~Chang, L.~Yan, and S.~Zhong.
\newblock Transformed low-rank model for line pattern noise removal.
\newblock In {\em IEEE International Conference on Computer Vision (ICCV)},
  pages 1726--1734, 2017.

\bibitem{chen2013generalized}
Y.-L. Chen and C.-T. Hsu.
\newblock A generalized low-rank appearance model for spatio-temporally
  correlated rain streaks.
\newblock In {\em IEEE International Conference on Computer Vision (ICCV)},
  pages 1968--1975, 2013.

\bibitem{deshpande2017learning}
A.~Deshpande, J.~Lu, M.-C. Yeh, J.~M. Chong, and D.~Forsyth.
\newblock Learning diverse image colorization.
\newblock In {\em Proceedings of the IEEE Conference on Computer Vision and
  Pattern Recognition (CVPR)}, pages 6837--6845, 2017.

\bibitem{vid_wacv}
Y.~Du, J.~Xu, Q.~Qiu, X.~Zhen, and L.~Zhang.
\newblock Variational image deraining.
\newblock In {\em The IEEE Winter Conference on Applications of Computer Vision
  (WACV)}, March 2020.

\bibitem{esser2018variational}
P.~Esser, E.~Sutter, and B.~Ommer.
\newblock A variational u-net for conditional appearance and shape generation.
\newblock In {\em Proceedings of the IEEE Conference on Computer Vision and
  Pattern Recognition (CVPR)}, pages 8857--8866, 2018.

\bibitem{fu2017clearing}
X.~{Fu}, J.~{Huang}, X.~{Ding}, Y.~{Liao}, and J.~{Paisley}.
\newblock Clearing the skies: A deep network architecture for single-image rain
  removal.
\newblock {\em IEEE Transactions on Image Processing}, 26(6):2944--2956, 2017.

\bibitem{fu2017removing}
X.~Fu, J.~Huang, D.~Zeng, Y.~Huang, X.~Ding, and J.~Paisley.
\newblock Removing rain from single images via a deep detail network.
\newblock In {\em Proceedings of the IEEE Conference on Computer Vision and
  Pattern Recognition (CVPR)}, pages 3855--3863, 2017.

\bibitem{fu2013variational}
X.~Fu, D.~Zeng, Y.~Huang, X.~Ding, and X.-P. Zhang.
\newblock A variational framework for single low light image enhancement using
  bright channel prior.
\newblock In {\em IEEE Global Conference on Signal and Information Processing},
  pages 1085--1088, 2013.

\bibitem{gan}
I.~Goodfellow, J.~Pouget-Abadie, M.~Mirza, B.~Xu, D.~Warde-Farley, S.~Ozair,
  A.~Courville, and Y.~Bengio.
\newblock Generative adversarial nets.
\newblock In {\em In Advances in Neural Information Processing Systems
  (NeurIPS)}, pages 2672--2680, 2014.

\bibitem{Ham2018}
C.~Ham, A.~Raj, V.~Cartillier, and I.~Essa.
\newblock Variational image inpainting.
\newblock In {\em NeurIPS workshop on Bayesian Deep Learning}, 2018.

\bibitem{maskrcnn}
K.~{He}, G.~{Gkioxari}, P.~{Dollár}, and R.~{Girshick}.
\newblock Mask r-cnn.
\newblock In {\em IEEE International Conference on Computer Vision (ICCV)},
  pages 2980--2988, 2017.

\bibitem{he2010single}
K.~He, J.~Sun, and X.~Tang.
\newblock Single image haze removal using dark channel prior.
\newblock {\em IEEE Transactions on Pattern Analysis and Machine Intelligence},
  33(12):2341--2353, 2010.

\bibitem{he2016deep}
K.~He, X.~Zhang, S.~Ren, and J.~Sun.
\newblock Deep residual learning for image recognition.
\newblock In {\em Proceedings of the IEEE Conference on Computer Vision and
  Pattern Recognition (CVPR)}, pages 770--778, 2016.

\bibitem{hoffman2013stochastic}
M.~D. Hoffman, D.~M. Blei, C.~Wang, and J.~Paisley.
\newblock Stochastic variational inference.
\newblock {\em The Journal of Machine Learning Research}, 14(1):1303--1347,
  2013.

\bibitem{huang2017densely}
G.~Huang, Z.~Liu, L.~Van Der~Maaten, and K.~Q. Weinberger.
\newblock Densely connected convolutional networks.
\newblock In {\em Proceedings of the IEEE Conference on Computer Vision and
  Pattern Recognition (CVPR)}, pages 4700--4708, 2017.

\bibitem{ioffe2015batch}
S.~Ioffe and C.~Szegedy.
\newblock Batch normalization: Accelerating deep network training by reducing
  internal covariate shift.
\newblock In {\em International Conference on Machine Learning (ICML)}, pages
  448--456, 2015.

\bibitem{JIN2020107143}
X.~Jin, Z.~Chen, and W.~Li.
\newblock Ai-gan: Asynchronous interactive generative adversarial network for
  single image rain removal.
\newblock {\em Pattern Recognition}, 100:107143, 2020.

\bibitem{kang2012automatic}
L.-W. Kang, C.-W. Lin, and Y.-H. Fu.
\newblock Automatic single-image-based rain streaks removal via image
  decomposition.
\newblock {\em IEEE Transactions on Image Processing}, 21(4):1742--1755, 2012.

\bibitem{kingma2014adam}
D.~P. Kingma and J.~Ba.
\newblock Adam: A method for stochastic optimization.
\newblock {\em arXiv preprint arXiv:1412.6980}, 2014.

\bibitem{kingma2013auto}
D.~P. Kingma and M.~Welling.
\newblock Auto-encoding variational bayes.
\newblock {\em arXiv preprint arXiv:1312.6114}, 2013.

\bibitem{kohl2018probabilistic}
S.~Kohl, B.~Romera-Paredes, C.~Meyer, J.~De~Fauw, J.~R. Ledsam, K.~Maier-Hein,
  S.~A. Eslami, D.~J. Rezende, and O.~Ronneberger.
\newblock A probabilistic u-net for segmentation of ambiguous images.
\newblock In {\em Advances in Neural Information Processing Systems}, pages
  6965--6975, 2018.

\bibitem{li2018recurrent}
X.~Li, J.~Wu, Z.~Lin, H.~Liu, and H.~Zha.
\newblock Recurrent squeeze-and-excitation context aggregation net for single
  image deraining.
\newblock In {\em European Conference on Computer Vision (ECCV)}, pages
  254--269, 2018.

\bibitem{li2016rain}
Y.~Li, R.~T. Tan, X.~Guo, J.~Lu, and M.~S. Brown.
\newblock Rain streak removal using layer priors.
\newblock In {\em Proceedings of the IEEE Conference on Computer Vision and
  Pattern Recognition (CVPR)}, pages 2736--2744, 2016.

\bibitem{luo2015removing}
Y.~Luo, Y.~Xu, and H.~Ji.
\newblock Removing rain from a single image via discriminative sparse coding.
\newblock In {\em IEEE International Conference on Computer Vision (ICCV)},
  pages 3397--3405, 2015.

\bibitem{maas2013rectifier}
A.~L. Maas, A.~Y. Hannun, and A.~Y. Ng.
\newblock Rectifier nonlinearities improve neural network acoustic models.
\newblock In {\em International Conference on Machine Learning (ICML)}, 2013.

\bibitem{mirza2014conditional}
M.~Mirza and S.~Osindero.
\newblock Conditional generative adversarial nets.
\newblock {\em arXiv preprint arXiv:1411.1784}, 2014.

\bibitem{niqe}
A.~Mittal, R.~Soundararajan, and A.~C. Bovik.
\newblock Making a completely blind image quality analyzer.
\newblock {\em IEEE Signal Processing Letters}, 22(3):209–--212, 2013.

\bibitem{qian2018attentive}
R.~Qian, R.~T. Tan, W.~Yang, J.~Su, and J.~Liu.
\newblock Attentive generative adversarial network for raindrop removal from a
  single image.
\newblock In {\em Proceedings of the IEEE Conference on Computer Vision and
  Pattern Recognition (CVPR)}, pages 2482--2491, 2018.

\bibitem{Ren_2019_CVPR}
D.~Ren, W.~Zuo, Q.~Hu, P.~Zhu, and D.~Meng.
\newblock Progressive image deraining networks: A better and simpler baseline.
\newblock In {\em Proceedings of the IEEE Conference on Computer Vision and
  Pattern Recognition (CVPR)}, 2019.

\bibitem{ren2019pami}
D.~Ren, W.~Zuo, D.~Zhang, L.~Zhang, and M.-H. Yang.
\newblock Simultaneous fidelity and regularization learning for image
  restoration.
\newblock {\em IEEE Transactions on Pattern Analysis and Machine Intelligence},
  2019.

\bibitem{rezende2014stochastic}
D.~J. Rezende, S.~Mohamed, and D.~Wierstra.
\newblock Stochastic backpropagation and approximate inference in deep
  generative models.
\newblock {\em arXiv preprint arXiv:1401.4082}, 2014.

\bibitem{ronneberger2015u}
O.~Ronneberger, P.~Fischer, and T.~Brox.
\newblock U-net: Convolutional networks for biomedical image segmentation.
\newblock In {\em International Conference on Medical image computing and
  computer-assisted intervention}, pages 234--241. Springer, 2015.

\bibitem{2007smcm.book}
R.~Y. Rubinstein.
\newblock {\em Simulation and the Monte Carlo Method}.
\newblock John Wiley \& Sons, Inc., New York, NY, USA, 1st edition, 1981.

\bibitem{sohn2015learning}
K.~Sohn, H.~Lee, and X.~Yan.
\newblock Learning structured output representation using deep conditional
  generative models.
\newblock In {\em In Advances in Neural Information Processing Systems
  (NeurIPS)}, pages 3483--3491, 2015.

\bibitem{sultani2018real}
W.~Sultani, C.~Chen, and M.~Shah.
\newblock Real-world anomaly detection in surveillance videos.
\newblock In {\em Proceedings of the IEEE Conference on Computer Vision and
  Pattern Recognition (CVPR)}, pages 6479--6488, 2018.

\bibitem{sun2020learning}
H.~Sun, Y.~Du, J.~Xu, Y.~Yin, X.~Zhen, and L.~Shao.
\newblock Learning to learn kernels with variational random features, 2020.

\bibitem{walker2016uncertain}
J.~Walker, C.~Doersch, A.~Gupta, and M.~Hebert.
\newblock An uncertain future: Forecasting from static images using variational
  autoencoders.
\newblock In {\em European Conference on Computer Vision (ECCV)}, pages
  835--851. Springer, 2016.

\bibitem{wang2004image}
Z.~Wang, A.~C. Bovik, H.~R. Sheikh, and E.~P. Simoncelli.
\newblock Image quality assessment: from error visibility to structural
  similarity.
\newblock {\em IEEE Transactions on Image Processing}, 13(4):600--612, 2004.

\bibitem{RANet2019}
Z.~Wang, J.~Xu, L.~Liu, F.~Zhu, and L.~Shao.
\newblock Ranet: Ranking attention network for fast video object segmentation.
\newblock In {\em The IEEE International Conference on Computer Vision (ICCV)},
  Oct 2019.

\bibitem{Wei_2019_CVPR}
W.~Wei, D.~Meng, Q.~Zhao, Z.~Xu, and Y.~Wu.
\newblock Semi-supervised transfer learning for image rain removal.
\newblock In {\em Proceedings of the IEEE Conference on Computer Vision and
  Pattern Recognition (CVPR)}, pages 3877--3886, 2019.

\bibitem{xu2015empirical}
B.~Xu, N.~Wang, T.~Chen, and M.~Li.
\newblock Empirical evaluation of rectified activations in convolutional
  network.
\newblock {\em arXiv preprint arXiv:1505.00853}, 2015.

\bibitem{STAR2020}
J.~Xu, Y.~Hou, D.~Ren, L.~Liu, F.~Zhu, M.~Yu, H.~Wang, and L.~Shao.
\newblock Star: A structure and texture aware retinex model.
\newblock {\em IEEE Transactions on Image Processing}, 29:5022--5037, 2020.

\bibitem{gid2018}
J.~Xu, L.~Zhang, and D.~Zhang.
\newblock External prior guided internal prior learning for real-world noisy
  image denoising.
\newblock {\em IEEE Transactions on Image Processing}, 27(6):2996--3010, June
  2018.

\bibitem{twsc}
J.~Xu, L.~Zhang, and D.~Zhang.
\newblock A trilateral weighted sparse coding scheme for real-world image
  denoising.
\newblock In {\em European Conference on Computer Vision (ECCV)}, September
  2018.

\bibitem{mcwnnm}
J.~Xu, L.~Zhang, D.~Zhang, and X.~Feng.
\newblock Multi-channel weighted nuclear norm minimization for real color image
  denoising.
\newblock In {\em IEEE International Conference on Computer Vision (ICCV)}, Oct
  2017.

\bibitem{pgpd}
J.~Xu, L.~Zhang, W.~Zuo, D.~Zhang, and X.~Feng.
\newblock Patch group based nonlocal self-similarity prior learning for image
  denoising.
\newblock In {\em IEEE International Conference on Computer Vision (ICCV)},
  pages 244--252, 2015.

\bibitem{yan2018mt}
X.~Yan, A.~Rastogi, R.~Villegas, K.~Sunkavalli, E.~Shechtman, S.~Hadap,
  E.~Yumer, and H.~Lee.
\newblock Mt-vae: Learning motion transformations to generate multimodal human
  dynamics.
\newblock In {\em European Conference on Computer Vision (ECCV)}, pages
  265--281, 2018.

\bibitem{yang2019tip}
W.~{Yang}, J.~{Liu}, S.~{Yang}, and Z.~{Guo}.
\newblock Scale-free single image deraining via visibility-enhanced recurrent
  wavelet learning.
\newblock {\em IEEE Transactions on Image Processing}, 28(6):2948--2961, 2019.

\bibitem{yang2017deep}
W.~Yang, R.~T. Tan, J.~Feng, J.~Liu, Z.~Guo, and S.~Yan.
\newblock Deep joint rain detection and removal from a single image.
\newblock In {\em Proceedings of the IEEE Conference on Computer Vision and
  Pattern Recognition (CVPR)}, pages 1357--1366, 2017.

\bibitem{yang2017deeppami}
W.~{Yang}, R.~T. {Tan}, J.~{Feng}, J.~{Liu}, S.~{Yan}, and Z.~{Guo}.
\newblock Joint rain detection and removal from a single image with
  contextualized deep networks.
\newblock {\em IEEE Transactions on Pattern Analysis and Machine Intelligence},
  2019.

\bibitem{zhang2017convolutional}
H.~Zhang and V.~M. Patel.
\newblock Convolutional sparse and low-rank coding-based rain streak removal.
\newblock In {\em WACV}, pages 1259--1267. IEEE, 2017.

\bibitem{zhang2018density}
H.~Zhang and V.~M. Patel.
\newblock Density-aware single image deraining using a multi-stream dense
  network.
\newblock In {\em Proceedings of the IEEE Conference on Computer Vision and
  Pattern Recognition (CVPR)}, pages 695--704, 2018.

\bibitem{zhang2017image}
H.~Zhang, V.~Sindagi, and V.~M. Patel.
\newblock Image deraining using a conditional generative adversarial network.
\newblock {\em IEEE Transactions on Circuits and Systems for Video Technology},
  2019.

\bibitem{zhu2017joint}
L.~{Zhu}, C.~{Fu}, D.~{Lischinski}, and P.~{Heng}.
\newblock Joint bi-layer optimization for single-image rain streak removal.
\newblock In {\em IEEE International Conference on Computer Vision (ICCV)},
  pages 2545--2553, 2017.

\end{thebibliography}
}
\end{document}